\UseRawInputEncoding
\documentclass{article}
\usepackage{enumitem}  

\setlist[itemize]{noitemsep}
\usepackage{microtype}
\usepackage{graphicx}
\usepackage{subcaption}
\usepackage{booktabs} 
\usepackage{booktabs}  

\usepackage{multirow} 
\usepackage{spverbatim}
\usepackage{listings}
\usepackage{xcolor}
\usepackage[most]{tcolorbox}
\tcbuselibrary{listings, breakable, skins}

\usepackage{hyperref}

\usepackage[preprint]{icml2026}

\usepackage{amsmath}
\usepackage{amssymb}
\usepackage{mathtools}
\usepackage{amsthm}

\usepackage[capitalize,noabbrev]{cleveref}

\theoremstyle{plain}

\theoremstyle{definition}

\theoremstyle{remark}

\usepackage[textsize=tiny]{todonotes}

\icmltitlerunning{T2MBench: A Benchmark for Out-of-Distribution Text-to-Motion Generation}

\begin{document}

\twocolumn[
  \icmltitle{T2MBench: A Benchmark for Out-of-Distribution Text-to-Motion Generation}

\icmlsetsymbol{equal}{*}

\begin{icmlauthorlist}
  \icmlauthor{Bin Yang}{equal,hkust}
  \icmlauthor{Rong Ou}{equal,hkust}
  \icmlauthor{Weisheng Xu}{equal,hkust}
  \icmlauthor{Jiaqi Xiong}{oxford}
  \icmlauthor{Xintao Li}{hkust}
  \icmlauthor{Taowen Wang}{hkust}
  \icmlauthor{Luyu Zhu}{hkust}
  \icmlauthor{Xu Jiang}{hkust}
  \icmlauthor{Jing Tan}{hkust}
  \icmlauthor{Renjing Xu}{hkust}
\end{icmlauthorlist}

\icmlaffiliation{hkust}{The Hong Kong University of Science and Technology (Guangzhou), Guangzhou, China}
\icmlaffiliation{oxford}{University of Oxford, Oxford, United Kingdom}

\icmlcorrespondingauthor{Renjing Xu}{renjingxu@hkust-gz.edu.cn}

\icmlkeywords{Machine Learning, Text-to-Motion, Humanoid Robots, ICML}

\vskip 0.3in
]

\printAffiliationsAndNotice{\icmlEqualContribution}




\begin{abstract}
Most existing evaluations of text-to-motion generation focus on in-distribution textual inputs and a limited set of evaluation criteria, which restricts their ability to systematically assess model generalization and motion generation capabilities under complex out-of-distribution (OOD) textual conditions. To address this limitation, we propose a benchmark specifically designed for OOD text-to-motion evaluation, which includes a comprehensive analysis of 14 representative baseline models and the two datasets derived from evaluation results. Specifically, we construct an OOD prompt dataset consisting of 1,025 textual descriptions. Based on this prompt dataset, we introduce a unified evaluation framework that integrates LLM-based Evaluation, Multi-factor Motion evaluation, and Fine-grained Accuracy Evaluation. Our experimental results reveal that while different baseline models demonstrate strengths in areas such as text-to-motion semantic alignment, motion generalizability, and physical quality, most models struggle to achieve strong performance with Fine-grained Accuracy Evaluation. These findings highlight the limitations of existing methods in OOD scenarios and offer practical guidance for the design and evaluation of future production-level text-to-motion models.

\end{abstract}

\section{Introduction}

Text-to-motion (T2M) generation aims to generate temporally coherent and semantically aligned human motions from natural language descriptions. It has broad applications in embodied AI, VR/AR, and human--computer interaction. 
As generation quality improves, existing evaluation protocols are no longer sufficient to assess generalization in complex settings. 

\textbf{Current Limitations in Text-to-Motion Models Evaluation.} Most existing T2M evaluations focus on in-distribution inputs and a limited set of metrics, making it difficult to systematically evaluate robustness and physical plausibility in OOD datasets. Moreover, OOD datasets are still scarce, limiting the ability to rigorously assess model performance under diverse and complex conditions. These limitations in evaluation frameworks hinder the adoption of T2M methods for production-level generation, where robustness, scalability, and versatility are critical for real-world applications.

\textbf{Our Contribution: T2MBench.} T2MBench addresses the limitations of existing evaluation methods by proposing a multi-dimensional evaluation framework and releasing an out-of-distribution (OOD) text prompt dataset to compensate for the scarcity of OOD data. Additionally, we release two datasets, each exhibiting the best performance in semantic alignment and physical quality, aimed at addressing the limitations of current evaluation frameworks.

Our OOD text prompt dataset consists of 1,025 prompts, categorized into four main types: Dynamics, Complexity, Interaction, and Accuracy. T2MBench includes three evaluation dimensions: (1) LLM-based evaluation (evaluate rendered motions using LLM and designed system prompts); (2) Multi-factor Motion Evaluation (calculate the semantic alignment, generalizability, and physical quality of motions); (3) Fine-grained Accuracy Evaluation (Measure the numerical accuracy of generated motions). The two released datasets are based on T2MBench evaluation results, selecting the best semantic and physical attribute motions for each prompt across 14 baselines to form the released semantic and physical attribute datasets. we conducted tracking experiments on the physical quality dataset, achieving a tracking success rate of over 90\%.
Our contributions are summarized as follows:
\vspace{-10pt}
\begin{itemize}
    \item \textit{\textbf{Multi-Dimensional Benchmark}}. We propose \textbf{T2M-Bench}, to our knowledge the first comprehensive benchmark for assessing Text-to-Motion models, measuring the semantic alignment, generalizability, physical plausibility, and accuracy of generated motions.
    \item  \textbf{\textit{High-Quality Dataset Release.}} We released a high quality OOD prompt dataset and two datasets that excel in physical and semantic attributes, respectively.
    \item  \textbf{\textit{Systematic Model Evaluation.}}  We systematically evaluate state-of-the-art Text-to-Motion models across various dimensions, uncovering limitations, especially in fine-grained accuracy of motion generation.
    \item \textbf{\textit{Standardized Research Protocols.}} We provide unified evaluation formats and prompt design protocols to support reproducible research and standardized comparison for the future Text-to-Motion models.
\end{itemize}
\vspace{-10pt}
\textbf{T2MBench} provides a comprehensive foundation for future research on Text-to-Motion models and offers a standardized platform to evaluate model performance across various dimensions.

\begin{figure*}[t] 
  \centering
  \includegraphics[width=0.9\textwidth]{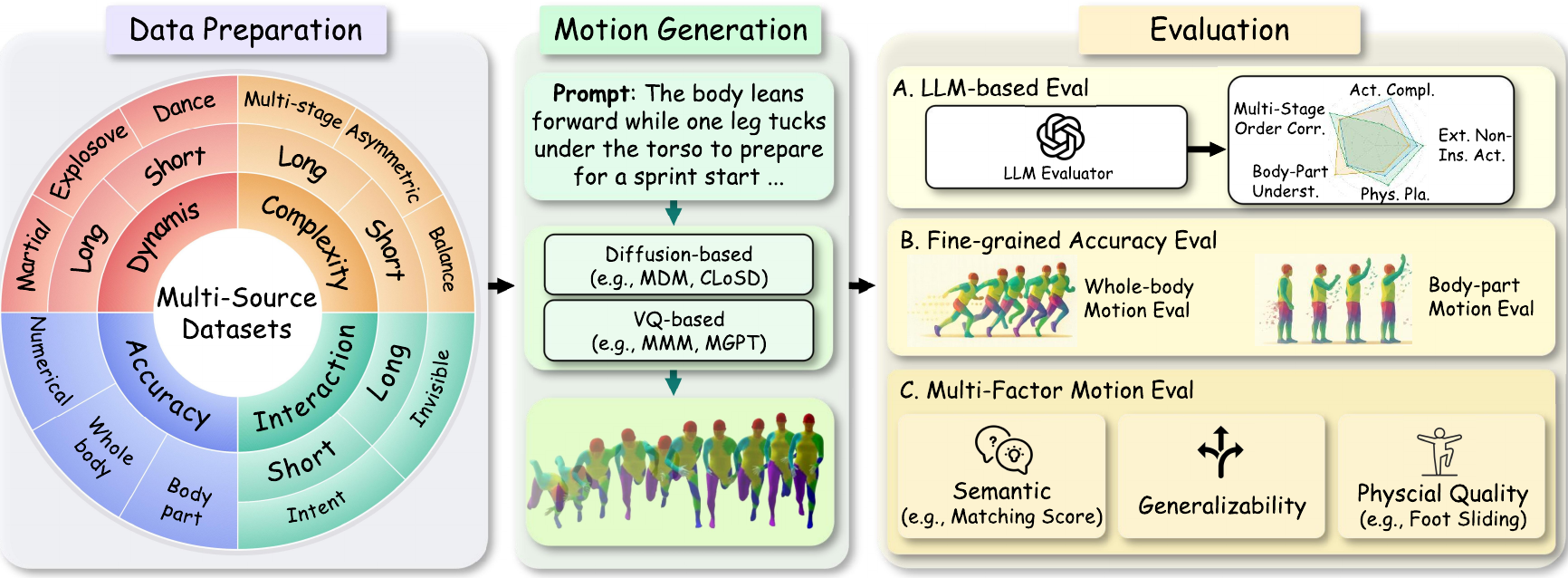} 
  \caption{Overall pipeline of T2MBench.}
  \vspace{-10pt}
  \label{fig:pipeline}
\end{figure*}

\section{Related Works}
\label{sec:related_works}
\subsection{T2M Models and Datasets.} 
Text-to-motion (T2M) generation primarily employs diffusion-based methods \cite{tevet2022human, shafir2023human, yuan2024mogents} or VQ-based paradigms \cite{Guo_2022_CVPR, jiang2023motiongpt, zhou2024avatargpt}. While diffusion models excel in latent consistency and VQ-based models offer strong semantic diversity, both remain fragile under out-of-distribution (OOD) textual instructions. Foundational datasets like KIT-ML\cite{Plappert_2016}, HumanML3D \cite{Guo_2022_CVPR} and Motion-X \cite{lin2024motionxlargescale3dexpressive} have underpinned these models, yet they predominantly focus on In-Distribution (ID) scenarios that mirror training data. Furthermore, reliance on abstract adjectives often results in underspecified kinematic precision. 

\subsection{Evolution of Motion Evaluation.} 
Early Text-to-Motion evaluation lacked standardized protocols and relied heavily on qualitative visualization and small-scale user studies, limiting reproducibility and comparability\cite{liu2024bridginggaphumanmotion}. With the widespread adoption of large paired datasets such as KIT-ML\cite{Plappert_2016} and HumanML3D\cite{Guo_2022_CVPR}, the community converged on automatic, benchmark-style protocols dominated by embedding-space retrieval and distributional similarity: R-Precision measures text--motion alignment via retrieval in a shared latent space, while FID compares feature distributions between generated and real motions; Diversity and Multimodality are often reported to characterize generalizability. This paradigm improves repeatability, yet it is inherently biased toward statistical matching and can under-represent physical feasibility and fine-grained execution constraints.

As diffusion-based models \cite{tevet2022human, zhang2023t2mgptgeneratinghumanmotion, guo2024momask} improved motion fidelity, evaluation limitations became more pronounced: models can score well on statistical metrics while exhibiting contact and constraint failures such as foot sliding, floating, ground and body penetration. To address these artifacts, evaluation has expanded toward kinematics- and physics-inspired checks \cite{tevet2024closd, lin2025vimogen}, including \textit{Foot Sliding}, \textit{Foot Floating}, \textit{Ground Penetration}, and \textit{Jitter Degree}, and further incorporates manifold- or prior-based measures to quantify \textit{Pose Quality} \cite{he2024nrdfneuralriemanniandistance}. These metrics are interpretable and diagnostic for contact-related errors, but they often struggle to capture long-horizon temporal logic (e.g., multi-stage ordering) and the alignment between action intensity and textual intent.

More recently, \emph{LLM/VLM-as-a-judge} has introduced semantic reasoning into T2M evaluation by scoring motions against rubric-style criteria, improving coverage for open-vocabulary and compositional prompts \cite{liu2023gevalnlgevaluationusing, gu2025surveyllmasajudge}. However, such pipelines typically render 3D kinematics into 2D videos/frames, creating a representation gap between 3D geometry and VLM perception \cite{cheng2024spatialrgptgroundedspatialreasoning}; text-caption intermediate pipelines can further amplify information loss \cite{dong2019wordsequalvideospecificinformation}, especially for spatial orientation and fine-grained relative pose constraints. Overall, T2M evaluation has progressed from statistical metrics to kinematic checks and LLM-based judging, yet robust measurements for long-term temporal structure and \emph{intensity--intent} alignment remain insufficiently standardized.

\vspace{-10pt}
\section{T2MBench}

\subsection{OOD Text Prompt Dataset}
\label{sec:benchmark}

\begin{figure}[htbp]
  \centering
  \includegraphics[width=0.9\columnwidth]{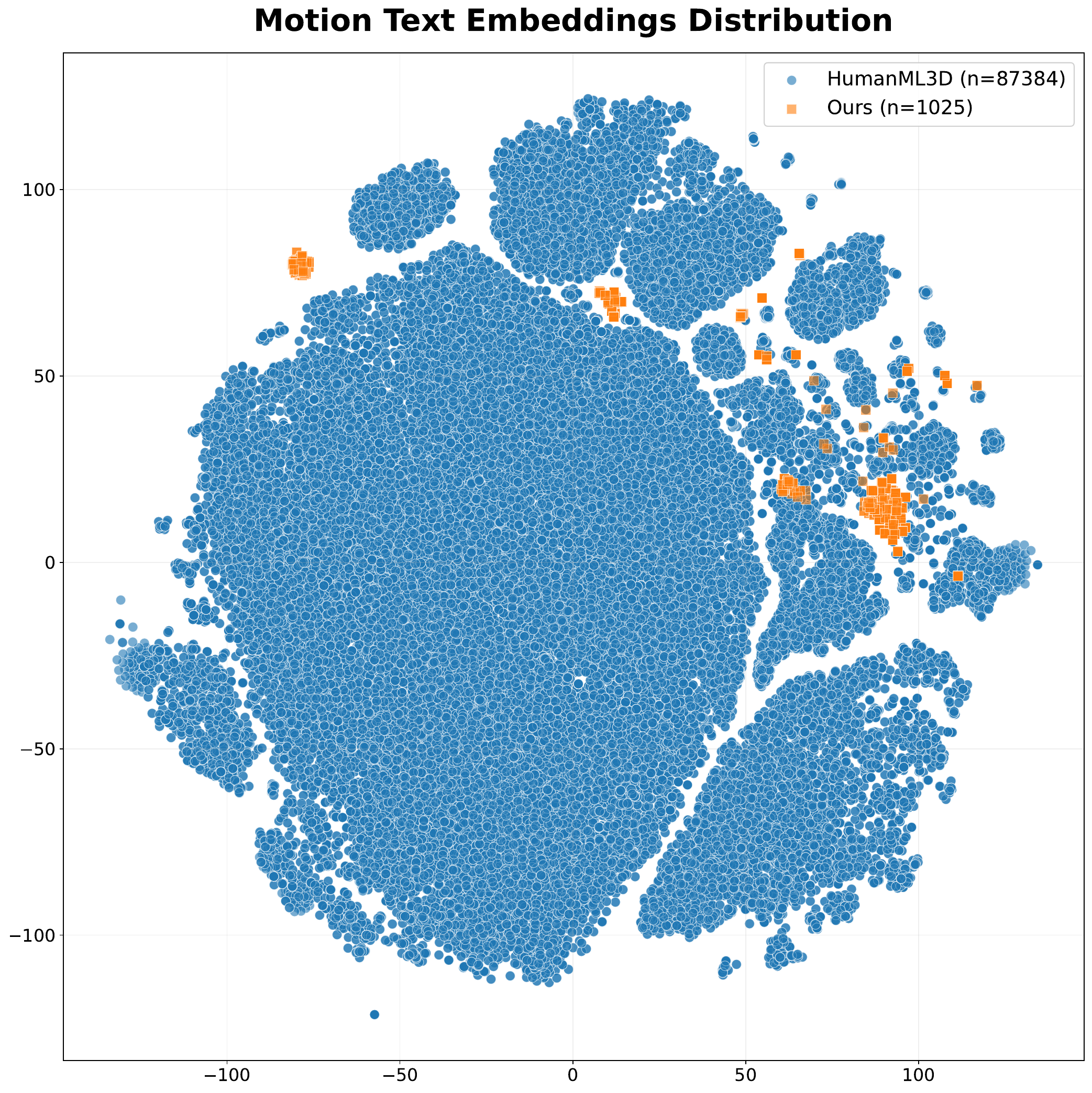} 
  \caption{Comparison of t-SNE results between our OOD prompt dataset and the HumanML3D text dataset.}
  \vspace{-10pt}
  \label{fig:tsne} 
\end{figure}

\paragraph{Construction Pipeline.} 
Our dataset draws on Vimogen-228k, HumanML3D, AMASS, and the Oxford Dictionary \cite{lin2025vimogen, Guo_2022_CVPR, Mahmood_2019_ICCV}. We initialized the corpus using 450 JSON files from Vimogen-228k and expanded it using LLMs to simulate the creative logic of professional choreographers and martial arts experts . To ensure physical executability and high granularity, we enforce the \textbf{``Clay Figurine Rule''}: prompts must use precise kinematic language describing spatial displacements of the head, torso, and limbs while strictly avoiding vague abstract adjectives . Except for numerical tasks, the use of Arabic numerals is prohibited, forcing the model to rely on pure spatial directions (e.g., ``left'', ``up'').
\vspace{-15pt}
\paragraph{Dataset Type.} 
The dataset comprises 1,025 prompts across 11 subtypes, categorized into four major semantic dimensions: (1) \textbf{Dynamics} (explosive, dance, martial); (2) \textbf{Complexity} (balance, multi-stage, asymmetric); (3) \textbf{Interaction} (intent, invisible); and (4) \textbf{Accuracy} (numerical, whole-body, body-part) . For the first three dimensions, prompts are further bifurcated into \textit{long} and \textit{short} sequences, while \textit{Accuracy} focuses on short, high-precision descriptions. All prompts underwent LLM-based refinement for linguistic naturalness followed by manual quality filtering .
\vspace{-5pt}
\paragraph{OOD Validation.} 
To verify the out-of-distribution (OOD) nature of our dataset, we encoded our prompts and the entire HumanML3D corpus using the MotionGPT T5 encoder \cite{jiang2023motiongpt}. Dimensionality reduction via t-SNE (Fig.~\ref{fig:tsne}) and flow precision analysis ($k=3$) reveal an overlap rate of only 3.6098\%, confirming strong OOD characteristics compared to existing benchmarks .

\begin{table}[htbp]
\centering
\caption{Size distribution across all dataset types}
\label{tab:dataset_dist}
\resizebox{\columnwidth}{!}{
\begin{tabular}{@{}llcc@{}}
\toprule
\textbf{Dimension} & \textbf{Sub-dimensions} & \textbf{Long} & \textbf{Short} \\ \midrule
Dynamics & dance, explosive, martial & 125 & 150 \\
Complexity & multi-stage, asymmetric, balance & 120 & 150 \\
Interaction & intent, invisible & 100 & 100 \\
Accuracy & numerical, whole-body, body part & NA & 280 \\\bottomrule 
\end{tabular}
}
\end{table}

\subsection{Text-to-Motion Baselines}
We evaluated 14 of the most state-of-the-art Text-to-Motion models from the past two years. These models were categorized into two main groups based on their action encoders: Diffusion-based methods and VQ-based Generative Methods.
\vspace{-5pt}
\begin{itemize}
    \item \textbf{Diffusion-based methods:} include MDM, CLoSD, PriorMDM, MoGenTS, HY-Motion-1.0, and DartControl. 
    \item \textbf{VQ-based Methods:} include T2M, MotionGPT, MMM, AvatarGPT, SALAD, MoMask, ViMoGen, and MaskControl. 
\end{itemize}

\subsection{Evaluation}

\subsubsection{LLM-based Evaluation}
\label{sec:llm_eval}

\begin{figure}[htbp]
  \centering
  \includegraphics[width=0.9\columnwidth]{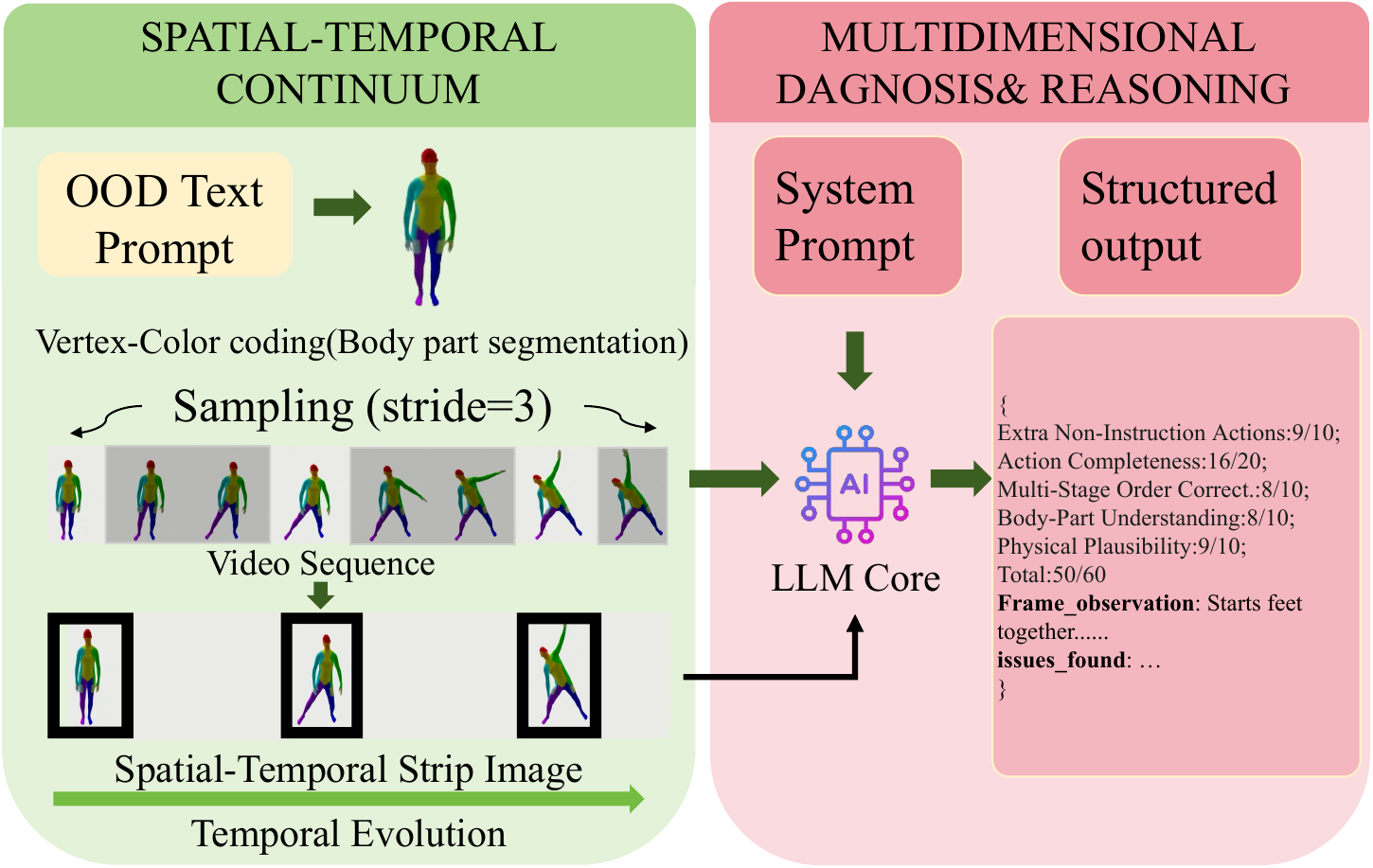} 
  \caption{The LLM-based evaluation pipeline.}
  \label{fig:LLM-based evaluation pipeline} 
\end{figure}

Following the pipeline in Fig.~\ref{fig:LLM-based evaluation pipeline}, we utilize GPT-5.2 to assess five key dimensions of generated motions: \textit{Extra Non-Instruction Actions}, \textit{Action Completeness}, \textit{Multi-Stage Order Correctness}, \textit{Body-Part Understanding}, and \textit{Physical Plausibility}. The evaluator outputs scores and auditable explanations in a structured JSON format.
\vspace{-5pt}
\paragraph{Motion Rendering and Normalization.} 
Generated joint trajectories are rendered via the SMPL body model using an off-screen renderer. To ensure comparability across baselines, we apply a unified coordinate alignment and height normalization. Sequences are captured using a fixed camera with an adaptive viewport, determined by the spatial bounding box of the SMPL mesh to ensure complete motion coverage without viewpoint drift. 
\vspace{-5pt}
\paragraph{Color-coded Body-Part Strip.} 
To bridge the representation gap between 3D kinematics and VLM perception, we implement two strategies:
\vspace{-5pt}
\begin{itemize}
    \item \textbf{Vertex Color-Coding}: Vertices are assigned high-contrast part labels (e.g., torso, limbs, face) based on their proximity to joint groups. This segmentation remains consistent throughout the sequence to emphasize limb trajectories and multi-stage structures.
    \item \textbf{Temporal Evolution Encoding}: Rendered frames are sampled at a fixed stride and horizontally concatenated to form a single 2D 
    \textit{Spatial-Temporal Strip} image. This strip serves as the sole visual input, allowing the LLM to scan temporal transitions and sequential relationships along the horizontal axis.
\end{itemize}

During inference, the LLM is instructed to treat the text prompt as a set of kinematic constraints, focusing exclusively on visible human motion while ignoring background or environmental factors. The specific system prompt is shown in appendix \ref{appendix:llm-prompt}.

\vspace{-10pt}

\subsubsection{Multi-Factor Motion Evaluation}
To provide a multi-factor evaluation of the generated sequences, we assess our model across three key dimensions: semantic alignment, generalizability, and physical quality. Detailed calculation formulas are provided in the Appendix \ref{appendix:Multi-Factor Motion Evaluation}.

\paragraph{Dimension 1: Semantic Alignment.}
This dimension measures semantic similarity between text and motion using three metrics:
\vspace{-10pt}
\begin{enumerate}
    \item \textbf{Matching Score}: Computes the distance between text and motion features in a joint latent space.
    \vspace{-5pt}
    \item \textbf{R-Precision}: Assesses the ability to retrieve the correct description from a set of motion candidates.
    \vspace{-5pt}
    \item \textbf{Automatic Similarity Recall (ASR)}: Measures the recall rate of relevant atomic actions based on cosine similarity between atomic-action features from the Ground Truth prompt and the model output prompt, using a fixed similarity threshold.
\end{enumerate}
\vspace{-10pt}

\paragraph{Dimension 2: Generalizability.}
This dimension evaluates generalization across varied scenarios via:
\vspace{-10pt}
\begin{enumerate}
    \item \textbf{Multimodality (MM)}: Measures how distinct multiple generated motion sequences are under the same text input, based on pairwise distances among sampled outputs.
    \vspace{-5pt}
    \item \textbf{Diversity}: Measures overall dataset-level diversity by averaging distances between randomly sampled motion sequences generated from different text inputs.
\end{enumerate}
\vspace{-10pt}

\paragraph{Dimension 3: Physical Quality.}
This dimension evaluates realism and physical plausibility using:
\vspace{-10pt}
\begin{enumerate}
    \item \textbf{Jitter Degree (JD)}: Quantifies motion jitter via keyframe acceleration statistics, combining global and local components.
    \vspace{-5pt}
    \item \textbf{Ground Penetration (GP)}: Measures the extent of ground penetration by aggregating keyframes below a tolerance threshold.
    \vspace{-5pt}
    \item \textbf{Foot Floating (FF)}: Quantifies abnormal foot floating during contact phases based on foot velocity, relative velocity, and contact state, including invalid frames and floating interval durations.
    \vspace{-5pt}
    \item \textbf{Foot Sliding (FS)}: Measures horizontal foot movement during ground contact by aggregating contact-weighted horizontal velocities for both feet.
    \vspace{-5pt}
    \item \textbf{Dynamic Degree (DD)}: Measures motion dynamics via joint-velocity magnitudes, combining global and local dynamics after removing root translation.
    \vspace{-5pt}
    \item \textbf{Position Quality (PQ)}: Evaluates pose plausibility by the distance to the pose manifold predicted by a Neural Distance Field (NRDF) model.
    \vspace{-5pt}
    \item \textbf{Body Penetration (BP)}: Measures self-penetration using BVH collision detection by counting collision triangle pairs per frame normalized by the SMPL triangle count.
\end{enumerate}

\subsubsection{Fine-grained Accuracy Evaluation}
\label{sec:fine_grained_eval}
For fine-grained evaluation, we construct a test set of 200 text instructions: 100 for whole-body motion and 100 for body-part motion. Whole-body instructions specify global displacement, velocity, or rotation angles (e.g., moving 2.8 meters backward, targeting 2.0 m/s forward speed, or turning left by 100 degrees), while body-part instructions define relative positional constraints between two body parts (e.g., keeping the right hand 15 cm in front of the head). We feed these texts into 14 baseline models to obtain generated motions. Prior to evaluation, we use GPT-5.2 to extract target specifications from the instructions, storing whole-body targets in \texttt{root\_move.json} and body-part targets in \texttt{body\_part.json}. These files serve exclusively as reference data for error computation and are not used as model inputs. Additionally, all data have been manually verified to be consistent with the test texts.

In Figure~\ref{fig:FINE-GRAINED ACCURACY ASSESSMENTS}, we additionally report "MaskControl with JointControl" as a separate baseline since MaskControl supports both text-only input and text combined with a joint-control trajectory. The joint-control trajectory is generated by GPT-5.2 to provide a physically plausible joint motion sequence, and separating this setting ensures a fair comparison without affecting the score of the standard MaskControl model.

We adopt two complementary protocols to quantify controllability at global and local levels, reporting Root Mean Square Error (RMSE) for each test case. For a motion clip with $T$ frames indexed by $t\in\{0,\dots,T-1\}$, we set $t_0=0$ and define the evaluation frame $t_e$ using a window length $N$ (default $N=30$); detailed formulas are provided in Appendix~\ref{appendix:Fine-grained accuracy evaluation}. \textbf{(1) Whole-body motion assessment} evaluates \emph{root} control targets from \texttt{root\_move.json} by recovering root position $p_t$ and root yaw $\psi_t$ (optionally denormalizing HumanML/MDM features $x\in\mathbb{R}^{T\times 263}$ with $(\mu,\sigma)$); since the HumanML/MDM 263-dim representation contains only yaw, we evaluate yaw rotation, directional root velocity (target speed $v^{*}$ along direction $u$ over duration $d$), and root translation (target displacement $\Delta p^{*}$) between $t_0$ and $t_e$. \textbf{(2) Body-part motion assessment} evaluates relative translation constraints from \texttt{body\_part.json} by comparing the relative displacement between a \emph{base} joint $b$ and a \emph{target} joint $g$ computed from $J\in\mathbb{R}^{T\times 22\times 3}$ against the target displacement $\Delta^{*}$, and reporting RMSE over the last $N$ frames (or all frames if $T<N$).

\vspace{-10pt}
\section{Experiment and Findings}
\label{sec:experiments}

\subsection{Setup}
\textbf{Data.} We evaluate our OOD prompt datasets with 1,025 text prompts grouped into four types
(\textit{Dynamics/Complexity/Interaction/Accuracy}, merged from 11 subtypes); Dynamics/Complexity/Interaction are split into
\textit{long/short}, while Accuracy contains \textit{short-only} prompts.%

\textbf{Baselines.} We benchmark 14 representative Text-to-Motion systems, including diffusion-based methods
(MDM, CLoSD, PriorMDM, MoGenTS, HY-Motion-1.0, DartControl) and VQ-based methods
(T2M, MotionGPT, MMM, AvatarGPT, SALAD, MoMask, ViMoGen, MaskControl).%

\textbf{Evaluation.} we report three evaluation dimensions: \textit{LLM-Based Evaluation},
\textit{Multi-Factor Metrics Evaluation}, and \textit{Fine-Grained Accuracy Evaluation}.%

For LLM-Based evaluation, motions are rendered with \textbf{SMPL} under unified normalization; we encode kinematics into a
\textit{spatial-temporal strip} (stride $=3$) and use \textbf{GPT-5.2} to score five designed dimensions with structured outputs. For the Fine‑Grained Accuracy Evaluation, we employ 200 controlled texts: 100 for whole‑body motion and 100 for body‑part motion.

\vspace{-10pt}

\subsection{Results and Analysis}
\subsubsection{LLM-Based Evaluation}
To select the Large Language Model (LLM) that best matches human preferences, we evaluated GPT-5.2, Google/Gemini-3-Flash-Preview, Z-AI/GLM-4.6v, NVIDIA/Nemotron-Nano-12B-V2-VL, Qwen/Qwen3-VL-32B-Instruct, and Anthropic/Claude-Opus-4.5 on the long sequence multi-stage category (50 prompts) generated by HY-Motion-1.0. To minimize individual subjectivity in the human preference data, we recruited five raters (aged 20--28) with normal cognitive abilities to independently evaluate the 50 samples. We then averaged the scores across the five raters to obtain the final human preference reference score. For each metric, we compute the absolute difference between each LLM's scores and human scores, then average across prompts to obtain LLM-based Evaluation scores over five dimensions and an overall score. Table~\ref{tab:llm_selection} shows that GPT-5.2 yields the closest results to human evaluators, and we therefore use GPT-5.2 in our evaluation.

\begin{figure*}[htbp]
  \centering
  \includegraphics[width=0.8\textwidth]{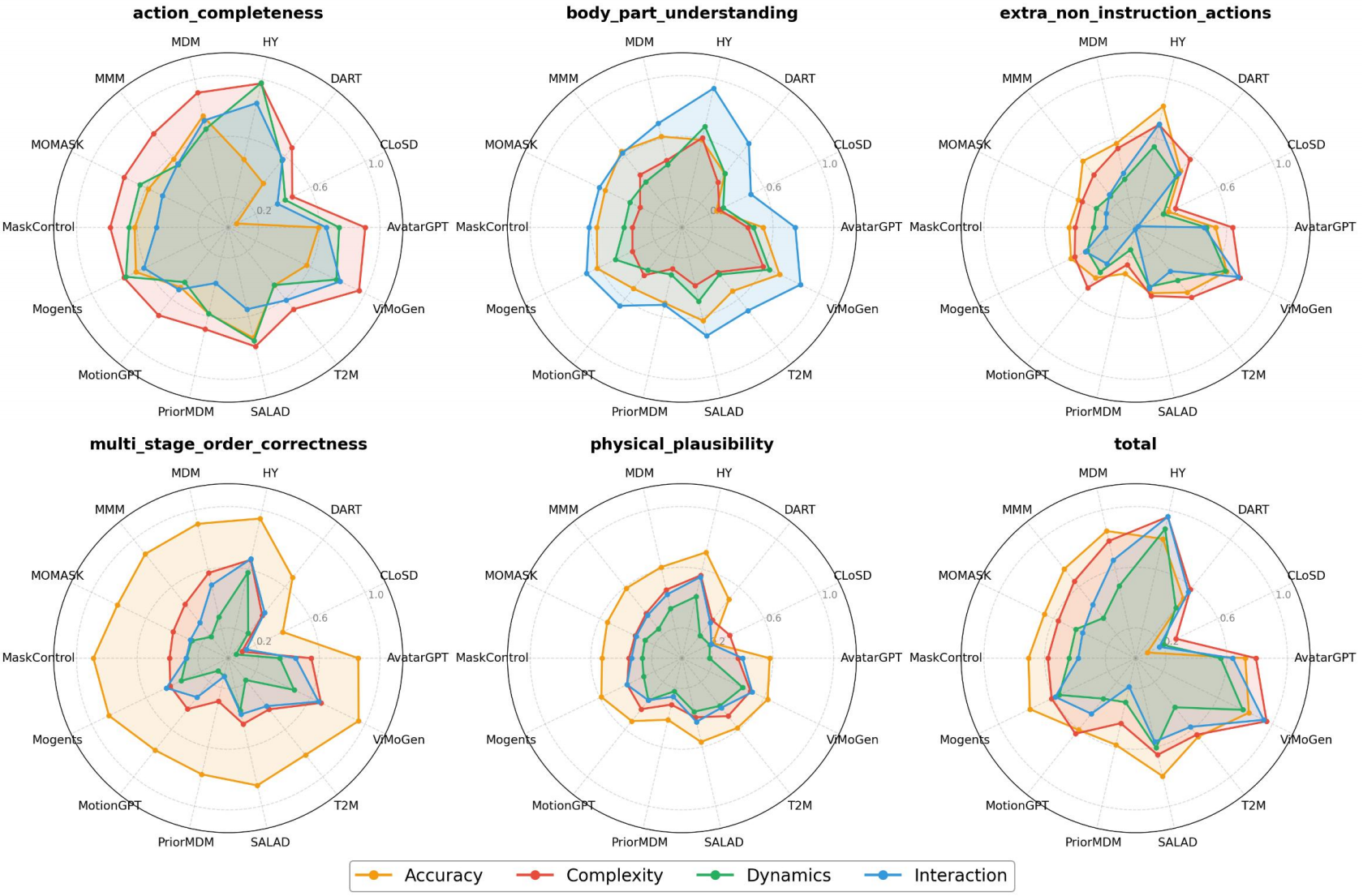} 
  \caption{LLM-Based Evaluation radar charts by evaluation dimensions}
  \label{fig:6d_llm} 
  \vspace{-10pt}
\end{figure*}

We report GPT-5.2 total scores for all 14 baselines in Table~\ref{tab:llm_1_eval_total} (grouped by dataset type). As shown in Figure~\ref{fig:6d_llm}, we build six radar charts (five metrics plus total score) with 14 axes corresponding to the 14 baselines. We further provide 14 per-baseline radar charts in Appendix~\ref{Radar charts by baselines} (Figure~\ref{fig:llm_14baseline}); each chart includes four curves for the four dataset types: Dynamics, Complexity, Interaction, and Accuracy. For each radar-chart sub-metric, we compute the integer-bounded range over 14 baselines $[\min_{\text{eval}}, \max_{\text{eval}}]$ and apply min--max normalization. For clarity and ease of interpretation in the visualization, we apply reversed normalization to metrics that are originally lower‑is‑better. The full procedure is given in Appendix~\ref{app:normalization}.

From Figure~\ref{fig:6d_llm}, HY and Vimogen achieve the best overall LLM-Based Evaluation performance across metrics, while CLoSD performs worst. By metric, the Complexity dataset performs best on action\_completeness, the Interaction dataset performs best on body\_part\_understanding, and the Accuracy dataset performs best on the remaining metrics. Table~\ref{tab:llm_1_eval_total} further shows that short-sequence Complexity and Accuracy generally score higher, and performance can vary substantially between long and short datasets: most baselines score lower on long sequences, except Dart, which performs better on the long dataset. Finally, Figure~\ref{fig:llm_14baseline} indicates consistent peaks across baselines: multi\_stage\_order\_correctness on the Accuracy dataset, action\_completeness on the Complexity dataset, and body\_part\_understanding on the Interaction dataset, reflecting stable behavior across dataset types.

\begin{table}[ht]
\centering
\caption{LLM selection Results, The x-axis corresponds to the five LLM-based evaluation criteria, where the abbreviations denote Extra Non-Instruction Actions (ENA), Action Completeness (AC), Multi-Stage Order Correctness (MOC), Body-Part Understanding (BPU), and Physical Plausibility(PP). The bold numbers indicate the smallest discrepancy with the human evaluation results.}
\resizebox{\columnwidth}{!}{
\begin{tabular}{|c|c|c|c|c|c|}
\hline
\textbf{Model} & \textbf{ENA}\,$\downarrow$ &  \textbf{AC}\,$\downarrow$ & \textbf{MOC}\,$\downarrow$ & \textbf{BPU}\,$\downarrow$ & \textbf{PP}\,$\downarrow$  \\
\hline
GPT-5.2 &\textbf{1.2800} &\textbf{3.1600} &\textbf{1.2400} &\textbf{1.4400} &\textbf{1.4800}  \\
\hline
Gemini-3 &2.9200 &4.1600 &2.7000 &2.8800 &2.5800  \\
\hline
GLM-4.6v &2.2045 &3.5909 &2.2500 &2.5909 &2.9545 \\
\hline
Nemotron &2.0000 &3.7959 &2.2449 &2.5102 &2.8163 \\
\hline
Qwen3 &3.5400 &4.4800 &2.7600 &2.9400 &2.9200 \\
\hline
Claude &1.3600 &4.3400 &1.6400 &1.9200 &1.6600 \\
\hline
\end{tabular}
}
\label{tab:llm_selection}
\vspace{-10pt}
\end{table}

\begin{table*}[t]
\centering
\caption{LLM-based evaluation total score results across four dataset types(Dynamics, Complexity, and Interaction for both Long and Short sequences, alongside Accuracy dataset type). \textbf{The best and second-best results are highlighted in red and blue, respectively.}}
\label{tab:llm_1_eval_total}
\resizebox{\textwidth}{!}{
\begin{tabular}{@{}lcccccccc@{}}
\toprule
\multirow{2}{*}{\textbf{Method}} & \multicolumn{3}{c}{\textbf{Long}} & \multicolumn{3}{c}{\textbf{Short}} & \multirow{2}{*}{\textbf{Accuracy}} \\ 
\cmidrule(lr){2-4} \cmidrule(lr){5-7}
 & \textbf{Dynamics} & \textbf{Complexity} & \textbf{Interaction} & \textbf{Dynamics} & \textbf{Complexity} & \textbf{Interaction} & \\ \midrule
T2M\cite{Guo_2022_CVPR} & 38.7049 & 38.3606 & 38.8800 & 38.7484 & 42.7601 & 41.5467 & 40.9556 \\
MDM\cite{tevet2022human} & 37.6798 & 41.0076 & 39.5700 & 40.8154 & 43.0333 & 42.3623 &\textbf{\color{red} 42.7408} \\ 
MotionGPT\cite{jiang2023motiongpt} & 37.7024 & 40.1272 & 38.8200 & 38.3815 & 41.2201 & 39.6152 & 40.4341 \\
CLoSD\cite{tevet2024closd} & 35.5910 & 38.8218 & 34.1500 & 37.8428 & 36.7333 & 38.9652 & 35.7744 \\
PriorMDM\cite{shafir2023human} & 36.0963 & 36.4888 & 34.8700 & 39.0029 & 40.9334 & 38.6002 & 40.2694 \\
MMM\cite{pinyoanuntapong2024mmm} & 36.7275 & 38.6776 & 36.0300 & 39.1680 & 42.5667 & 42.1245 & 41.7803 \\
AvatarGPT\cite{zhou2024avatargpt} & 40.0096 & 41.7138 & 39.9200 & 40.0738 & 42.4800 & 41.6151 & 41.5049 \\
Mogent\cite{yuan2024mogents} & 39.4167 & 39.5521 & 38.8800 & 40.5346 & 41.3067 & 41.6969 & 41.9625 \\
Salad\cite{hong2025salad} & 39.9207 & 39.3414 & 38.7000 & 40.8951 & 42.1134 & 41.4638 & 42.1765 \\
Momask\cite{guo2024momask} & 38.3925 & 38.3390 & 37.4700 & 39.3940 & 41.4933 & 39.5094 & 41.0121 \\
Dart\cite{zhao2024dartcontrol} & 40.6737 & 40.8206 & 41.6300 & 37.2818 & 39.7199 & 38.4119 & 39.5339 \\
Vimogen\cite{lin2025vimogen} &\textbf{\color{blue}{42.6812}} & \textbf{\color{red}{43.5114}} &\textbf{\color{blue}{43.8200}} &\textbf{\color{blue}{41.5350}} &\textbf{\color{blue}{43.7132}} & \textbf{\color{red}{43.0371}} &\textbf{\color{blue}{42.4599}} \\ 
HY\cite{wen2025hymotion10scalingflow} & \textbf{\color{red} {43.8414}} &\textbf{\color{blue}{43.3216}} &\textbf{\color{red} {44.3300}} &\textbf{\color{red}{42.0675}} &\textbf{\color{red} {43.8333}} &\textbf{\color{blue}{42.8963}} & 42.2598 \\ 
MaskControl\cite{pinyoanuntapong2025maskcontrol} & 38.2804 & 38.3022 & 36.8100 & 39.4872 & 41.7333 & 39.9959 & 41.3516 \\ \bottomrule
\end{tabular}
}
\end{table*}

\subsubsection{Multi-Factor Motion Evaluation}
\begin{figure*}[htbp]  
  \centering
  \includegraphics[width=0.8\textwidth]{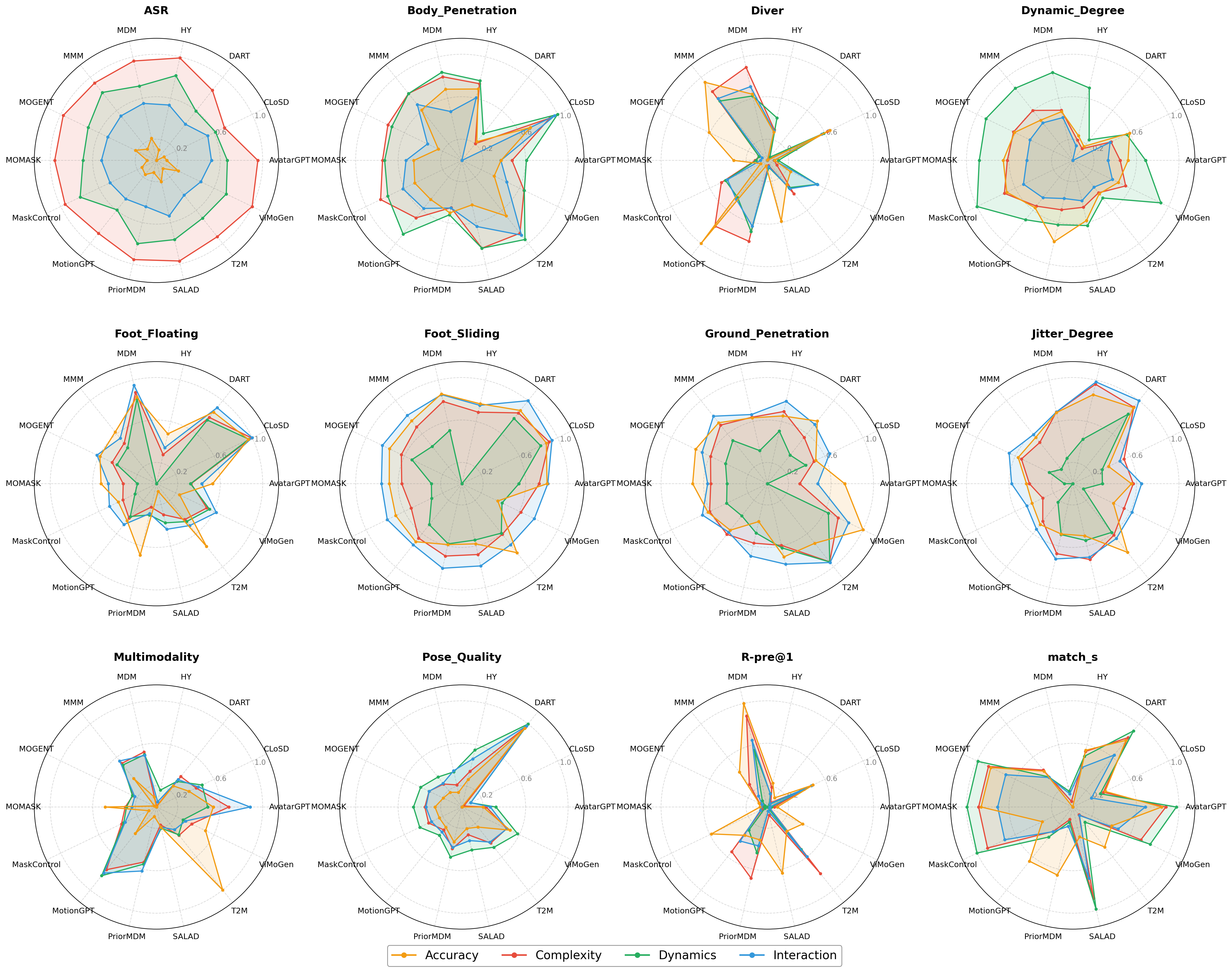} 
  \caption{Multi-Factor Motion Evaluation radar charts by metrics}
  \label{fig:metrics_12d} 
  \vspace{-10pt}
\end{figure*}

As shown in Figure~\ref{fig:metrics_12d}, we provide 12 radar charts (one per metric), each with 14 axes corresponding to the 14 text-to-motion baselines. Conversely, Figure~\ref{fig:multi-factor_14baseline} provides 14 radar charts (one per baseline), each with 12 axes corresponding to the 12 Multi-Factor Motion Evaluation metrics. Detailed per-metric results are reported in Appendix~\ref{Multi-Factor_detailed_results} (Table~\ref{tab:matching_scores}--Table~\ref{tab:body_penetration}).

For visualization, we normalize each metric using the $\max$ and $\min$ computed across the 14 baselines. Five metrics are originally lower-is-better: Jitter Degree, Ground Penetration, Foot Floating, Foot Sliding, and Body Penetration; we convert them to higher-is-better scores via the inverse min--max normalization in Appendix~\ref{app:normalization}.

Across semantic alignment, SALAD performs well on Matching Score across all dataset types, while MDM and CLoSD are consistently weak; for R-Pre@1, MDM, PriorMDM, and T2M are strong across all types; ASR varies less across baselines, with dataset-type trend Complexity $>$ Dynamics $>$ Interaction $>$ Accuracy. For generalizability, MotionGPT and AvatarGPT achieve strong Multimodality, while MMM, MDM, and MotionGPT perform well on Diversity and Dart is consistently weak. For physical quality, HY and Dart achieve strong Jitter Degree; T2M and ViMoGen perform best on Ground Penetration; MDM, Dart, and CLoSD perform well on Foot Floating whereas PriorMDM and SALAD are weaker; HY is relatively weak on Foot Sliding for the Dynamic dataset; Dart is relatively weak on Dynamic Degree but achieves the best Pose Quality; and CLoSD performs best on Body Penetration. Overall, baselines exhibit distinct strengths and weaknesses across metrics.

\subsubsection{Fine-Grained Accuracy Evaluation}

\begin{figure}[htbp]
  \centering

\includegraphics[width=0.8\columnwidth]{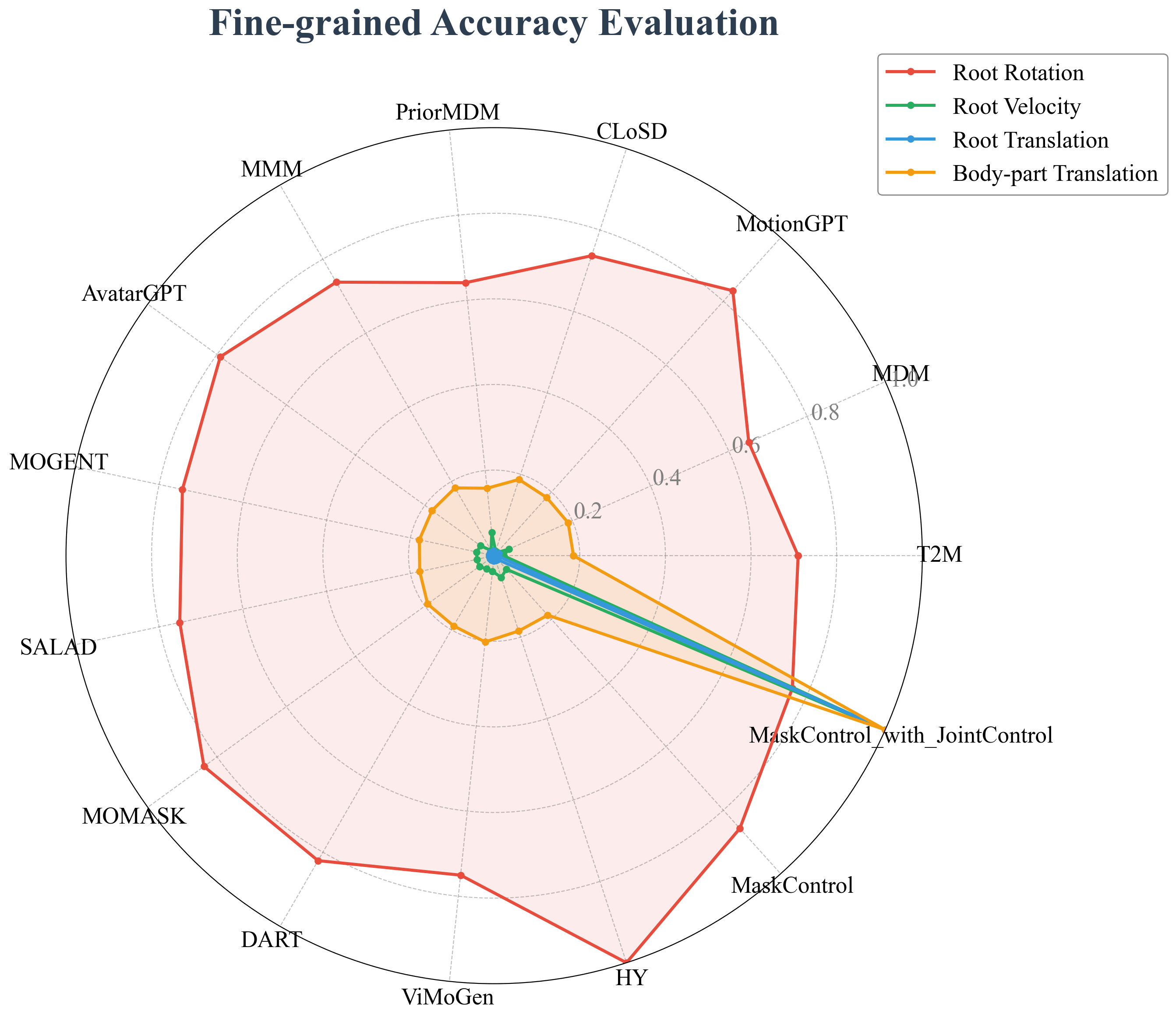}


  \caption{Fine-grained Accuracy Evaluation}
  \label{fig:FINE-GRAINED ACCURACY ASSESSMENTS} 
  \vspace{-30pt}
\end{figure}

The results of fine-grained accuracy evaluation for both whole-body motion and body-part motion are shown in Fig.~\ref{fig:FINE-GRAINED ACCURACY ASSESSMENTS} and Table \ref{tab:fine_grained_rmse}. 
In terms of the whole-body motion, specifically for Root Rotation, Hunyuan performs the best among all compared algorithms, though its advantage is not substantial. 
Note that MaskControl (with JointControl) is evaluated under an auxiliary control setting: it consumes an explicit joint-trajectory condition, which is not required by text-only baselines. Since our benchmark inputs are text-only, we obtain this trajectory by a fixed prompt-to-trajectory parsing pipeline from the same instruction (offline), i.e., it does not introduce additional semantic information beyond the text. Therefore, the JointControl results should be interpreted as controllability under stronger structured conditioning rather than a strictly apples-to-apples comparison to pure text-conditioned generators.

For Root Velocity, Root Translation, and Body-part Translation, MaskControl (with JointControl) exhibits a significant advantage. 
However, empirical results show that the method often produces implausible motions while trying to hit the target positions, suggesting that its overall performance still requires further improvement. Ultimately, this happens because MaskControl applies test-time gradient optimization to discrete token logits/embeddings to minimize the joint-trajectory control loss\cite{pinyoanuntapong2025maskcontrol}, which effectively prioritizes trajectory matching over the generative motion prior; if the prior or physical-plausibility constraints are not sufficiently weighted during optimization, the model will trade realism for tighter trajectory adherence. When only text input is provided, the performance of MaskControl shows no significant difference from other algorithms, and its overall effectiveness on the task remains mediocre.

\subsection{SOTA datasets}
We selected 745 prompts from the Dynamics, Complexity, and Interaction datasets and constructed two datasets: a Physical Attribute Dataset and a Semantic Attribute Dataset. For each text prompt, we chose (from 14 actions generated by 14 baseline models) the action with the highest Physical Attributes Score and the action with the highest Semantic Attributes Score, respectively. We initially planned to select a dataset from the Accuracy dataset using the fine-grained accuracy evaluation dimension. However, except for MaskControl, all baselines performed poorly on this dimension. Although MaskControl achieved better performance in this dimension, expert evaluation indicated its actions lacked naturalness. Therefore, current T2M models cannot generate precise actions suitable for production use, so we did not include any precision-related datasets.

We split our metrics into physical and semantic attributes, and designed two scoring tables (Physical and Semantic Attributes Score) to select the top-scoring action per prompt for each dataset.

For physical attributes, we used Physical Plausibility from the LLM-Based evaluation, and the Physical Quality dimension from the Multi-Factor Motion Evaluation, which includes: JD, GP, FF, FS, DD, PQ, and BP. The physical quality scoring formula is provided in Appendix \ref{appendix:The physical quality scoring formula}.

To validate the Physical Attributes Dataset, for each prompt we also randomly sampled one motion from a T2M baseline to build a random dataset of the same size. We tracked both datasets using the PHC+\cite{luo2023phc} tracker, with results in Table~\ref{tab:tracker_successrate} and in Table~\ref{tab:tracker_result}. Across all categories, selection by Physical Attributes Score improves tracking success rate by over 20\% compared to random selection; in Interaction it reaches 98\%. We further report mpjpe\_g, mpjpe\_l, mpjpe\_pa, accel\_dist, and vel\_dist during tracking in the appendix.

\begin{table}[ht]
\centering
\caption{PHC+\cite{luo2023phc} Tracking Success Rate(\%)}
\begin{tabular}{|c|c|c|c|}
\hline
\textbf{Dataset} & \textbf{Dynamics} & \textbf{Complexity} & \textbf{Interaction}  \\
\hline
ours &91.0(\textbf{+22.1}) &93.8(\textbf{+21.9}) &98.0(\textbf{+24.0})  \\
\hline
random &68.9 &71.9 &74.0  \\
\hline
\end{tabular}
\label{tab:tracker_successrate}
\end{table}

For semantic attributes, we used Extra Non-Instruction Actions, Action Completeness, Multi-Stage Order Correctness, and Body-Part Understanding from the LLM-Based Evaluation, together with matching score, R-Precision@1, R-Precision@2, R-Precision@3, and ASR from the Multi-Factor Motion Evaluation. We normalize the LLM-Based metrics, apply Formula 23 for smaller-is-better metrics and Formula 24 for larger-is-better metrics, then compute a weighted sum with weights: Extra Non-Instruction Actions 0.1, Action Completeness 0.1, Multi-Stage Order Correctness 0.1, Body-Part Understanding 0.1, Matching score 0.1, R-Precision@1 0.1, R-Precision@2 0.1, R-Precision@3 0.1, and ASR 0.2.

\begin{figure}[htbp]
  \centering

\includegraphics[width=1\columnwidth]{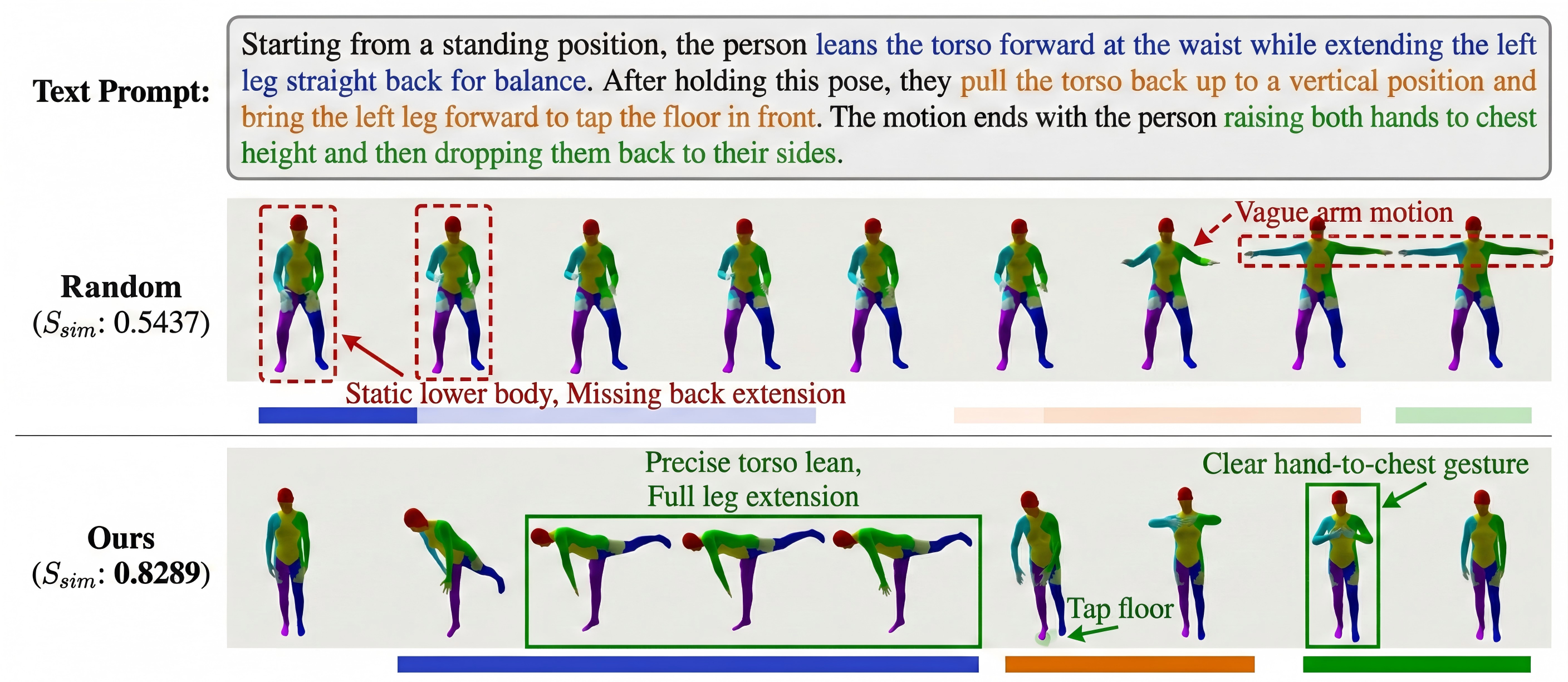}
  \caption{Semantic alignment comparison between our dataset and random dataset}
  \label{fig:Semantic alignment comparison} 
  \vspace{-10pt}
\end{figure}

Finally, we provide a qualitative comparison on one prompt between the top-scoring semantic sample and a randomly selected baseline (Figure~\ref{fig:Semantic alignment comparison}), showing markedly improved semantic consistency.

\section{Conclusion}
\label{sec:conclusion}

We propose a comprehensive benchmark for OOD text-to-motion evaluation, build a high-granularity prompt dataset and multi-dimensional evaluation framework, and conduct a systematic analysis of 14 representative models. 

Current Text-to-Motion models exhibit competent performance in semantic-alignment and physical quality. However, future improvements in fine-grained accuracy and generalization to multi-factor OOD instructions (dynamics/complexity/interaction) would substantially enhance their applicability in animation design, film production, and robotic control. By enabling the generation of large volumes of high-quality motions, such advancements could also significantly boost productivity across these domains.

Future work will expand cross-domain datasets to better support production-level Text-to-Motion model design and iteration. 

\vspace{-10pt}

\section*{Impact Statement}

This paper presents work whose goal is to advance the field of Machine
Learning. There are many potential societal consequences of our work, none
which we feel must be specifically highlighted here.

\bibliography{example_paper}
\bibliographystyle{icml2026}

\newpage
\appendix
\onecolumn
\section{System Prompt for LLM-Based Generation of the OOD Text Prompt Dataset}
\label{appendix:ood_generate_prompt}
The following prompts are used for LLM-based dataset generation of eight long sequence subtypes . 
\begin{tcblisting}{
    listing engine=listings,    % 使用 listings 引擎处理文字
    listing only,               % 只显示文字内容，不执行代码
    breakable,                  % 核心：允许跨页
    enhanced,                   % 核心：允许高级排版
    sharp corners,              % 直角边框
    colback=gray!5,             % 背景颜色
    colframe=black!75,          % 边框颜色
    left=5pt, right=5pt,        % 内边距
    top=5pt, bottom=5pt,
    before skip=0pt,            % 紧贴标题，消除上方空白
    label={box:multi-stage},
    % --- 下面是 listings 的具体设置，解决自动换行和特殊字符 ---
    listing options={
        basicstyle=\ttfamily\small,
        breaklines=true,        % 核心：允许自动换行
        breakatwhitespace=false,
        columns=fullflexible,
        keepspaces=true,
        showstringspaces=false,
        literate={≤}{{$\le$}}1 {≥}{{$\ge$}}1 {–}{{-}}1 % 自动处理 Unicode 字符
    }
}
//multi-stage subtype
## Role:
You are a professional motion designer and motion capture annotation expert, responsible for writing **high-precision, high-granularity** multi-stage motion descriptions for text-to-motion generation models.

## Task:
Generate **five distinct multi-stage motion descriptions**, each of which should be clear, concise, and consist of actions that can be executed by an average person.

## Motion Rules:
1. **Clear Starting Phase**: Each action's starting phase must specify clearly which body part moves first, avoiding vague descriptions.
2. **Natural Transitions**: The transitions between motion phases must be smooth, avoiding abrupt or unreasonable shifts. Each phase's movement should be simple and direct.
3. **Concise and Clear Movements**: Each phase's action should be clearly described and easy to execute. Avoid describing weight shifts or overly complex movements.
4. **Avoid Unexecutable Movements**: While creativity is encouraged, ensure the actions are feasible for humans to execute, avoiding extreme or impossible movements.
5. **Action Description Structure**: Each description must include a starting action, a transitional action, and an ending action, with each phase being as simple as possible and avoiding overly complicated steps.
6. **Clear Description of Each Phase**: Each phase must clearly describe which body parts are moving and how they are moving.
7. **Avoid Finger and Toe Details**: The motion description should only focus on the head, torso, arms, and legs, without mentioning fingers or toes.
8. **Variety in Movement**: The generated movements should have a certain degree of interest and variety while maintaining executability. Avoid repetitive or monotonous action descriptions.

## Examples:
1. The body slowly squats down with both arms hanging naturally. After pausing briefly at the lowest point, it gradually stands back up. Once fully upright, his right foot steps half a step forward, and the body then returns to the initial standing posture.

2. Starting from an upright position, the upper body rotates to the left while both feet remain fixed. After reaching the maximum rotation, the body pauses for one second, then rotates back to the neutral position and adjusts posture to regain stability.

\end{tcblisting}

\begin{tcblisting}{
    listing engine=listings,    % 使用 listings 引擎处理文字
    listing only,               % 只显示文字内容，不执行代码
    breakable,                  % 核心：允许跨页
    enhanced,                   % 核心：允许高级排版
    sharp corners,              % 直角边框
    colback=gray!5,             % 背景颜色
    colframe=black!75,          % 边框颜色
    left=5pt, right=5pt,        % 内边距
    top=5pt, bottom=5pt,
    before skip=0pt,            % 紧贴标题，消除上方空白
    label={box:dance},
    % --- 下面是 listings 的具体设置，解决自动换行和特殊字符 ---
    listing options={
        basicstyle=\ttfamily\small,
        breaklines=true,        % 核心：允许自动换行
        breakatwhitespace=false,
        columns=fullflexible,
        keepspaces=true,
        showstringspaces=false,
        literate={≤}{{$\le$}}1 {≥}{{$\ge$}}1 {–}{{-}}1 % 自动处理 Unicode 字符
    }
}
//dance subtype
## Role:
# Role
You are a Professional Choreographer and Motion Capture Specialist. Your task is to generate precise, high-granularity motion descriptions ("captions") for text-to-motion generation models.

# Task
Generate **5 distinct motion descriptions** specifically for the **Dance** category.

# Category Definition: DANCE
- **Goal:** Create coherent, realistic, and aesthetically valid dance sequences based on existing human dance styles (e.g., Ballet, Hip-hop, Contemporary, Jazz, Street Dance).
- **Style:** The movements must make physical sense. Avoid weird, glitchy, or impossible motions.
- **Descriptive approach:** Describe the *mechanics* of the dance (limbs, levels, flow) rather than just naming the genre.

# Core Philosophy: "The Clay Figurine Rule"
Imagine you are controlling a clay figurine or a robot. You must describe exactly *how* the body parts move.
- **DO NOT** describe abstract/general/vague motion. **DO NOT** use abstract adjectives.
- **DO** describe kinematics, the direction of motion, the right/left arm/leg...
# Strict Constraints
1. **Output Quantity:** Generate exactly 5 descriptions.
2. **Length:** Each description must be under 50 words.
3. **No Numbers:** **ABSOLUTELY NO NUMBERS.** Do not use degrees, counts, or durations. Use descriptive terms like "fully," "halfway," "wide," or "rapidly."
4. **Directionality:** **NEVER use "clockwise" or "counter-clockwise".** Use "left" or "right" instead.
5. **Realism:** Focus on grounded, achievable and logical dance moves that a professional dancer could perform.

# Few-Shot Examples (Style Reference)
*Note how these describe the movement, not the feeling.*

[Example 1 - Modern]
Standing tall, the upper body slowly contracts forward until the spine is curved. The arms hang heavy towards the ground. Suddenly, the spine ripples upward to a straight posture, and the arms float up to form a wide T-shape.

[Example 2 - Waltz]
The body steps forward with the left foot, rising onto the toes while swinging both arms gently to the right side. The feet come together briefly before the right foot steps back, lowering the heels and swinging arms to the left.

# Output
Please generate 5 realistic and specific **Dance** motion descriptions.

\end{tcblisting}

\begin{tcblisting}{
    listing engine=listings,    % 使用 listings 引擎处理文字
    listing only,               % 只显示文字内容，不执行代码
    breakable,                  % 核心：允许跨页
    enhanced,                   % 核心：允许高级排版
    sharp corners,              % 直角边框
    colback=gray!5,             % 背景颜色
    colframe=black!75,          % 边框颜色
    left=5pt, right=5pt,        % 内边距
    top=5pt, bottom=5pt,
    before skip=0pt,            % 紧贴标题，消除上方空白
    % label={box:asymmetric},
    % --- 下面是 listings 的具体设置，解决自动换行和特殊字符 ---
    listing options={
        basicstyle=\ttfamily\small,
        breaklines=true,        % 核心：允许自动换行
        breakatwhitespace=false,
        columns=fullflexible,
        keepspaces=true,
        showstringspaces=false,
        literate={≤}{{$\le$}}1 {≥}{{$\ge$}}1 {–}{{-}}1 % 自动处理 Unicode 字符
    }
}
//asymmetric subtype
# Asymmetric Control Motion Description Generation Instructions

## Role:
You are a professional motion designer specializing in **asymmetric control actions**. Your task is to generate high-resolution motion descriptions for the “Text-to-Motion” model. The core focus is to depict completely different motion logics between the left and right legs simultaneously.

## Task:
Generate **5 distinct asymmetric motion descriptions**.

% ## Core Rules (Must Follow Strictly):
% 1. **Compulsory Bilateral Comparison**: It is forbidden to describe only one side’s action. You must clearly specify “What is the left side doing?” and “What is the right side doing at the same time?” If one side is stationary, it must be explicitly stated.
% 2. **Three Logical Phases** (No Titles Required): Each description should naturally include the phases of action start, action process, and action end, without needing phase titles.
% 3. **No Numerical Descriptions**: Avoid using any numbers (e.g., 45 degrees, 3 seconds, 2 seconds, 90%). Use terms like “high, medium, low, large, small, slow, fast, rapid” instead.
% 4. **No Symbolic Terms**: Do NOT use terms like “clockwise” or “counterclockwise.” Use spatial terms for rotation, such as “tilt to the left” or “raise upward.”
% 5. **Concise and Executable**: The language must be concise and clear, ensuring the motion is physically executable in space, and avoid vague or unrealistic descriptions.
% 6. **No Object Interaction Descriptions**: Do not include descriptions related to object interactions, such as "reaching towards the ceiling" or "pretending to hold an object."
% 7. **No Non-Human Action Terminology**: Avoid using terms for actions that are not human, such as “circular pulse” or “lunge squat.”
% 8. **Describe Only Key Body Parts**: Only describe the movements of the head, torso, arms, and legs. Do not describe facial expressions or the movements of fingers or toes.
% 9. **Actions Can Be Unconventional**: The actions do not need to be overly conventional; make the actions more out-of-distribution while ensuring feasibility.

% ## Description Template (Reference):
% "First, the left side performs [Action A], while the right side simultaneously performs [Action B]. Then, the left side transitions to [Action C], and the right side switches to [Action D] at the same time. Finally, the left side completes [Action E], and the right side completes [Action F]."

% ## Example (Standard Reference):
% - Incorrect Example: “Right arm lifts, left leg steps sideways.” (Misses description of left arm and right leg)
% - Correct Example: “At the start of the action, the left arm slowly rises to shoulder height, while the right arm remains vertically attached to the thigh. Then, the left arm continues rising toward the head, and the right arm begins a small horizontal swing in front of the body. Finally, the left arm maintains the lifted posture, while the right arm lowers and returns to the starting position.”

% ## Output Requirements:
% Please strictly follow the above requirements to generate **5 distinct asymmetric motion descriptions**, ensuring logical clarity and clear bilateral comparison.


\end{tcblisting}

\begin{tcblisting}{
    listing engine=listings,    % 使用 listings 引擎处理文字
    listing only,               % 只显示文字内容，不执行代码
    breakable,                  % 核心：允许跨页
    enhanced,                   % 核心：允许高级排版
    sharp corners,              % 直角边框
    colback=gray!5,             % 背景颜色
    colframe=black!75,          % 边框颜色
    left=5pt, right=5pt,        % 内边距
    top=5pt, bottom=5pt,
    before skip=0pt,            % 紧贴标题，消除上方空白
    % label={box:asymmetric},
    % --- 下面是 listings 的具体设置，解决自动换行和特殊字符 ---
    listing options={
        basicstyle=\ttfamily\small,
        breaklines=true,        % 核心：允许自动换行
        breakatwhitespace=false,
        columns=fullflexible,
        keepspaces=true,
        showstringspaces=false,
        literate={≤}{{$\le$}}1 {≥}{{$\ge$}}1 {–}{{-}}1 % 自动处理 Unicode 字符
    }
}
//intent subtype
# Emotion Motion Description Generation Instructions

## Role:
You are an emotion motion designer, and your task is to create high-precision, high-granularity motion descriptions for the "Text-to-Motion" model to express a character's emotions.

## Notes:
1. **Avoid Numerical Descriptions**: Motion descriptions should not include numbers (e.g., 1 second, 45 degrees, 3 steps, etc.).
2. **Use Directional Descriptions**: Do not use terms like "clockwise" or "counterclockwise". Instead, use directional terms like "to the left", "to the right", "forward", "backward" to describe motion directions.
3. **Describe Key Body Parts**: Only describe the movements of the head, torso, arms, and legs. Do not describe facial expressions, fingers, toes, or other detailed parts.
4. **Avoid Object Interaction**: Motion descriptions should not involve interactions with objects, such as "this person seems to be lifting something".
5. **Emotion Diversity**: The motion descriptions need to express a wide range of emotions. Each description must clearly state which body parts are moving and their actions, and the actions should be executable.

## Description Template (Reference):
- **Emotion Description Template**:  
  "[Emotion State]: The body posture [describe the body movement], the arms [describe the action], the legs [describe the action], the head [describe the action], and the body expresses [emotional bodily manifestation]."

  Example:
  - **Hesitation**: The body leans slightly forward, the arms are restlessly moving at the sides of the body, the legs lift but do not step forward, the head slightly shakes, conveying uncertainty.
  - **Caution**: The upper body leans slightly backward, the arms stay close to the torso, the head turns slowly from side to side as if observing the surroundings, conveying a cautious alertness.


## Output Requirements:
Please strictly follow the above requirements to generate **5 distinct emotion-based motion descriptions**, ensuring clear emotion expression, and clarity in the description of the actions.

\end{tcblisting}

\begin{tcblisting}{
    listing engine=listings,    % 使用 listings 引擎处理文字
    listing only,               % 只显示文字内容，不执行代码
    breakable,                  % 核心：允许跨页
    enhanced,                   % 核心：允许高级排版
    sharp corners,              % 直角边框
    colback=gray!5,             % 背景颜色
    colframe=black!75,          % 边框颜色
    left=5pt, right=5pt,        % 内边距
    top=5pt, bottom=5pt,
    before skip=0pt,            % 紧贴标题，消除上方空白
    % label={box:asymmetric},
    % --- 下面是 listings 的具体设置，解决自动换行和特殊字符 ---
    listing options={
        basicstyle=\ttfamily\small,
        breaklines=true,        % 核心：允许自动换行
        breakatwhitespace=false,
        columns=fullflexible,
        keepspaces=true,
        showstringspaces=false,
        literate={≤}{{$\le$}}1 {≥}{{$\ge$}}1 {–}{{-}}1 % 自动处理 Unicode 字符
    }
}
//balance subtype
# Balance Motion Description Generation Instructions

## Role:
You are a **balance motion designer**. Your task is to create high-precision, high-granularity motion descriptions to help characters maintain balance.

## Notes:
1. **Avoid Numerical Descriptions**: Motion descriptions should not include numbers (e.g., 1 second, 45 degrees, 3 steps, etc.).
2. **Use Directional Descriptions**: Motion descriptions cannot use terms like "clockwise" or "counterclockwise". Use directional terms such as "to the left", "to the right", "forward", "backward" to describe directions.
3. **Describe Key Body Parts**: Only describe the movements of the head, torso, arms, and legs. Do not describe facial expressions, fingers, toes, or other detailed body parts.
4. **Balance Motion Diversity**: The described balance actions should be as diverse as possible, such as single-leg standing, walking a tightrope, ice skating movements, falling, carrying weight while walking, etc.
5. **Clear and Specific Actions**: Each motion description must clearly explain what the relevant body parts are doing and ensure that the motion is executable.
6. **Object Interaction Descriptions**: If the motion involves interaction with an object, you must clarify whether the object is imaginary or if the character is simply mimicking an action involving an object (e.g., mimicking climbing stairs).

## Description Template (Reference):
- **Balance Motion Description Template**:  
  "First, the body [describe the starting motion], then [describe the change in motion], at this point, the [describe key body parts] perform [balance action]. Finally, the body returns to [describe the recovery motion], maintaining a stable standing position."

## Examples:
- **Example 1**:  
  The right foot lifts off the ground and remains suspended in the air, while both arms continuously adjust their positions to maintain balance. The foot is then slowly placed back onto the ground.

- **Example 2**:  
  The body leans noticeably to the left. To prevent losing balance, the right arm quickly extends outward while the left arm lowers, and the body eventually regains a stable standing posture.

## Output Requirements:
Please strictly follow the above requirements to generate **5 distinct balance motion descriptions** that clearly demonstrate balance techniques, with clear descriptions and executable actions. Each description should showcase different balance postures, ensuring the motion is feasible and executable.


\end{tcblisting}

\begin{tcblisting}{
    listing engine=listings,    % 使用 listings 引擎处理文字
    listing only,               % 只显示文字内容，不执行代码
    breakable,                  % 核心：允许跨页
    enhanced,                   % 核心：允许高级排版
    sharp corners,              % 直角边框
    colback=gray!5,             % 背景颜色
    colframe=black!75,          % 边框颜色
    left=5pt, right=5pt,        % 内边距
    top=5pt, bottom=5pt,
    before skip=0pt,            % 紧贴标题，消除上方空白
    % label={box:asymmetric},
    % --- 下面是 listings 的具体设置，解决自动换行和特殊字符 ---
    listing options={
        basicstyle=\ttfamily\small,
        breaklines=true,        % 核心：允许自动换行
        breakatwhitespace=false,
        columns=fullflexible,
        keepspaces=true,
        showstringspaces=false,
        literate={≤}{{$\le$}}1 {≥}{{$\ge$}}1 {–}{{-}}1 % 自动处理 Unicode 字符
    }
}
//invisible subtype
# Invisible Object Interaction Motion Description Generation Instructions

## Role:
You are an **Invisible Object Interaction Motion Designer**. Your task is to create high-precision, high-granularity motion descriptions to simulate interaction with invisible objects.

## Notes:
1. **Avoid Numerical Descriptions**: Motion descriptions should not include numbers (e.g., 1 second, 45 degrees, 3 steps, etc.).
2. **Use Directional Descriptions**: Motion descriptions cannot use terms like "clockwise" or "counterclockwise". Instead, use directional terms such as "to the left", "to the right", "forward", and "backward".
3. **Describe Key Body Parts**: Only describe the movements of the head, torso, arms, and legs. Do not describe facial expressions, fingers, toes, or other fine details.
4. **Diverse Invisible Object Interactions**: The described actions with invisible objects should be as diverse as possible, such as: playing billiards, playing the violin, playing ping pong, playing guitar, playing the flute, etc.
5. **Clear and Specific Actions**: Each motion description should clearly explain which body parts are performing the actions and ensure the motion is executable.
6. **Clarify the Invisible Object**: You must clarify that the object is invisible or that the person is simply mimicking an action involving the object (e.g., mimicking climbing stairs).

## Description Template (Reference):
- **Invisible Object Interaction Motion Description Template**:  
  "First, the body [describe the starting motion], then [describe the change in motion], at this point, the [describe key body parts] perform [the interaction with the invisible object]. Finally, the body returns to [describe the recovery motion] maintaining a stable position."

## Examples:
- **Example 1**:  
  Both hands maintain a fixed distance as if holding a heavy object. Throughout the movement, the arms keep a constant height, and the body leans slightly forward to support the imagined weight.

- **Example 2**:  
  The person moves forward as if walking along a narrow line, with both arms naturally extended and continuously making small adjustments to avoid losing balance.

## Output Requirements:
Please strictly follow the above requirements to generate **5 distinct invisible object interaction motion descriptions** that clearly demonstrate interaction with invisible objects, with clear descriptions and executable actions. Each description should showcase different interactions with invisible objects, ensuring the motion is feasible and executable.

\end{tcblisting}

\begin{tcblisting}{
    listing engine=listings,    % 使用 listings 引擎处理文字
    listing only,               % 只显示文字内容，不执行代码
    breakable,                  % 核心：允许跨页
    enhanced,                   % 核心：允许高级排版
    sharp corners,              % 直角边框
    colback=gray!5,             % 背景颜色
    colframe=black!75,          % 边框颜色
    left=5pt, right=5pt,        % 内边距
    top=5pt, bottom=5pt,
    before skip=0pt,            % 紧贴标题，消除上方空白
    % label={box:asymmetric},
    % --- 下面是 listings 的具体设置，解决自动换行和特殊字符 ---
    listing options={
        basicstyle=\ttfamily\small,
        breaklines=true,        % 核心：允许自动换行
        breakatwhitespace=false,
        columns=fullflexible,
        keepspaces=true,
        showstringspaces=false,
        literate={≤}{{$\le$}}1 {≥}{{$\ge$}}1 {–}{{-}}1 % 自动处理 Unicode 字符
    }
}
//explosive subtype
# Explosive Power Motion Description Generation Instructions

## Role:
You are an **explosive power motion designer**, and your task is to create high-precision, high-granularity motion descriptions that demonstrate explosive power actions.

## Notes:
1. **Avoid Numerical Descriptions**: Motion descriptions should not include numbers (e.g., 1 second, 45 degrees, 3 steps, etc.).
2. **Use Directional Descriptions**: Motion descriptions cannot use terms like "clockwise" or "counterclockwise". Instead, use directional terms such as "to the left", "to the right", "forward", and "backward".
3. **Describe Key Body Parts**: Only describe the movements of the head, torso, arms, and legs. Do not describe facial expressions, fingers, toes, or other fine details.
4. **Diverse Explosive Power Actions**: The explosive power actions described should be as diverse as possible, such as: Roundhouse Kick, Burpees, Jump Squats, Single-Leg Squat Jump, Alternating Jump Lunge, High Knee Sprint, Explosive Jumping Jack etc.
5. **Clear and Specific Actions**: Each motion description must clearly explain what the relevant body parts are doing and ensure that the motion is executable.
6. **The Interacting Object Must Be Invisible**: You must clarify whether the object is invisible or if the person is simply mimicking an action involving an object (e.g., mimicking climbing stairs).

## Description Template (Reference):
- **Explosive Power Motion Description Template**:  
  "First, the body [describe the preparatory motion to store energy], then [describe the rapid change of the explosive action], at this point, the [describe the key body parts exerting force to drive the action]. Finally, the body [describe the rapid recovery or adjustment of the posture] to maintain balance and a stable position."

## Examples:
- **Example 1**:  
  The body first lowers to store energy, with both arms swinging backward. It then rapidly jumps upward, completing a high leap while using the arms to generate upward momentum. Finally, the body lands softly with bent knees, using the arms to stabilize and regain balance.

- **Example 2**:  
  After briefly pausing to store power, the body suddenly bursts forward into a fast sprint, the legs extending powerfully while the arms swing forward to propel the body. Upon reaching maximum speed, the body quickly adjusts posture, maintaining balance.

## Output Requirements:
Please strictly follow the above requirements to generate **5 distinct explosive power motion descriptions**. Each description should demonstrate different explosive power actions, emphasizing the rapid generation of energy, the swift transfer of weight or acceleration, and ensuring the actions are clear, specific, and executable. Each motion description should include how the body quickly generates energy, how weight is rapidly transferred or speed is increased, and how the body ultimately recovers to a stable posture.


\end{tcblisting}

\begin{tcblisting}{
    listing engine=listings,    % 使用 listings 引擎处理文字
    listing only,               % 只显示文字内容，不执行代码
    breakable,                  % 核心：允许跨页
    enhanced,                   % 核心：允许高级排版
    sharp corners,              % 直角边框
    colback=gray!5,             % 背景颜色
    colframe=black!75,          % 边框颜色
    left=5pt, right=5pt,        % 内边距
    top=5pt, bottom=5pt,
    before skip=0pt,            % 紧贴标题，消除上方空白
    % label={box:asymmetric},
    % --- 下面是 listings 的具体设置，解决自动换行和特殊字符 ---
    listing options={
        basicstyle=\ttfamily\small,
        breaklines=true,        % 核心：允许自动换行
        breakatwhitespace=false,
        columns=fullflexible,
        keepspaces=true,
        showstringspaces=false,
        literate={≤}{{$\le$}}1 {≥}{{$\ge$}}1 {–}{{-}}1 % 自动处理 Unicode 字符
    }
}
//martial subtype
# Martial Arts Motion Description Prompt (OOD / Multi-stage / High Granularity)

## Role
You are a **professional martial arts motion designer and motion capture annotation expert**.
Your task is to generate **high-granularity, executable, and strongly martial-arts-distinctive** motion descriptions for **text-to-motion generation models**.

---

## Task
Generate **multiple distinct martial arts motion descriptions**.
These motions should be **intentionally out-of-distribution (OOD)** and suitable for **benchmark evaluation and stress testing**.

---

## Category Definition: Martial Arts (OOD)
- The motion must clearly belong to **martial arts** (e.g., strikes, blocks, kicks, combinations)
- Avoid dance, fitness, or daily-life movement styles
- Motions may deviate from standard techniques but must remain **physically executable**

---

## Core Generation Rules (Must Follow)

### 1️⃣ Multi-stage Structure
- Each motion must consist of **multiple consecutive stages**
- Transitions between stages must be clearly observable
- Do NOT label stages explicitly
- Do NOT use numeric descriptions

---

### 2️⃣ High-Granularity Body Mechanics
- Only describe the following body parts:
  - Head
  - Torso
  - Arms
  - Legs
- For every moving body part, explicitly specify:
  - Left or right side
  - How it moves (lift, extend, bend, retract, rotate, lower, etc.)
- **Do NOT** use abstract terms such as:
  - “apply force”, “shift balance”, “generate power”, “control momentum”

---

### 3️⃣ Explicit Left–Right Relationships
- Left and right body parts must be clearly distinguished
- If one side moves, the state of the opposite side must also be stated

---

### 4️⃣ OOD-Oriented Design
Intentionally include one or more of the following:
- Asymmetric initiation
- Counter-intuitive action order
- Mid-motion stabilization or pause
- Decoupled torso and limb movements
- Delayed motion on one side while the other completes first

---

### 5️⃣ Strict Prohibitions
- No numbers
- No angles, durations, or distances
- No fingers, toes, or facial expressions
- No objects or environments
- No imagined opponents or targets

---

## Output Format
- One complete paragraph per motion
- No bullet points
- No explanations
- Output only the motion description itself



\end{tcblisting}

\section{OOD Text Prompt Dataset Cloud Picture}
\label{appendix:word_cloud_pic}
\begin{figure*}[htbp]
    \hfill
    \centering
    \begin{minipage}{0.44\textwidth}
        \centering
        \includegraphics[width=\textwidth]{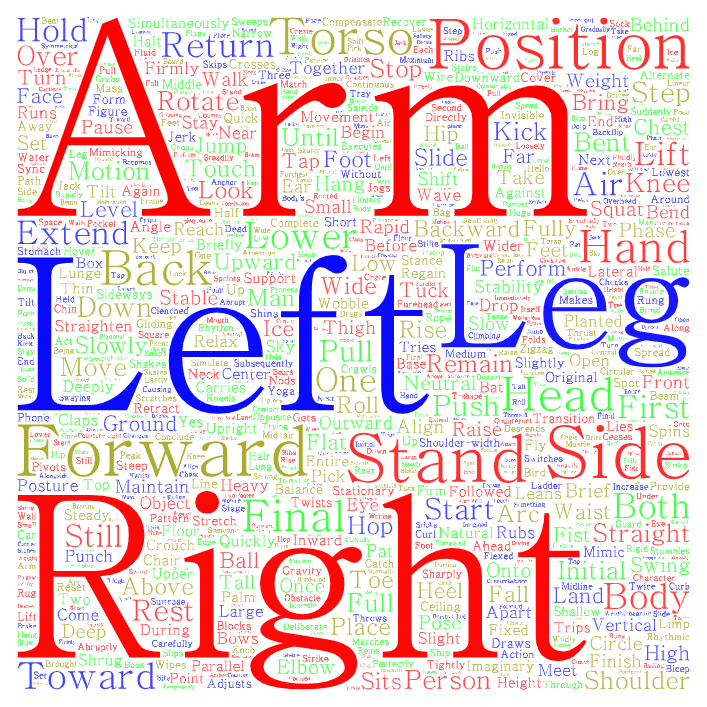}
        \caption{Complexity type dataset cloud picture}
        \label{fig:complexity}
    \end{minipage}
    \hfill
    \begin{minipage}{0.44\textwidth}
        \centering
        \includegraphics[width=\textwidth]{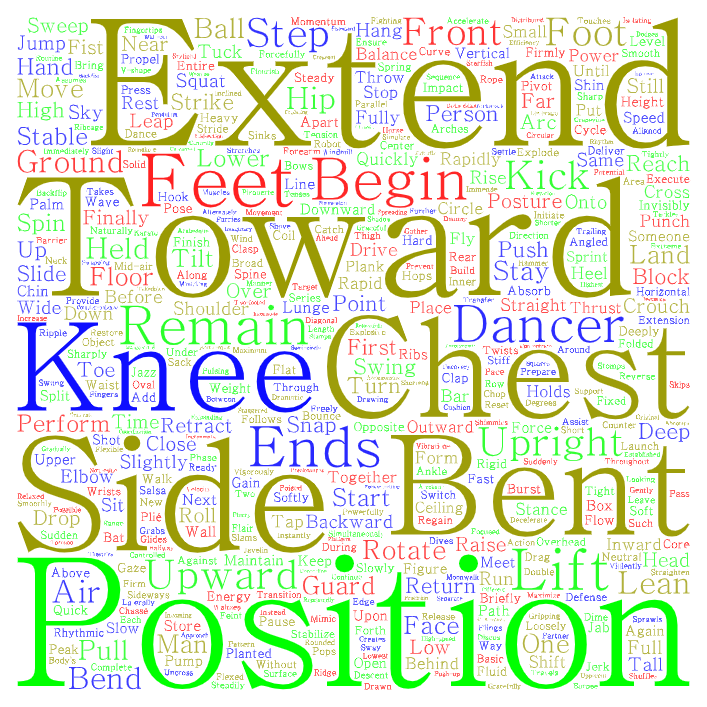}
        \caption{Dynamic type dataset cloud picture}
        \label{fig:figure1}
    \end{minipage}
\end{figure*}
\begin{figure*}[htbp]
    \hfill
    \begin{minipage}{0.44\textwidth}
        \centering
        \includegraphics[width=\textwidth]{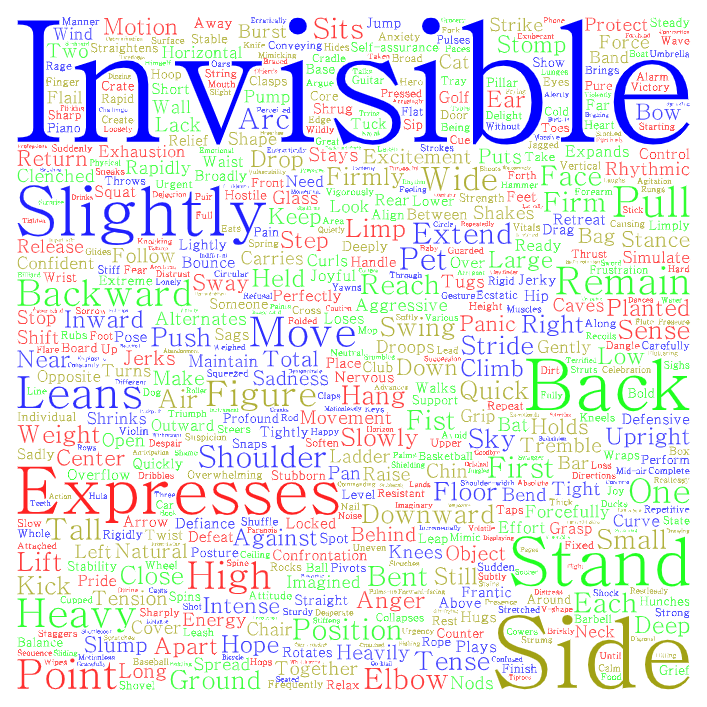}
        \caption{Interaction type dataset cloud picture}
        \label{fig:figure2}
    \end{minipage}
    \hfill
    \begin{minipage}{0.44\textwidth}
        \centering
        \includegraphics[width=\textwidth]{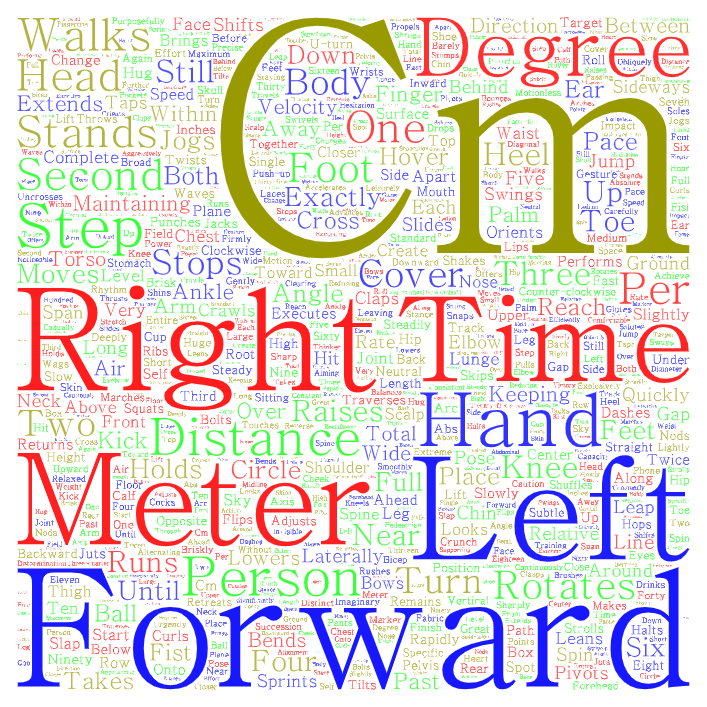}
        \caption{Accuracy type dataset cloud picture}
        \label{fig:figure3}
    \end{minipage}
\end{figure*}


\section{Evaluation methodology details}

\subsection{Multi-Factor Motion Evaluation}
\label{appendix:Multi-Factor Motion Evaluation}
To provide a multi-factor evaluation of the generated sequences, we assess our model across three key dimensions: semantic alignment, generalizability, and physical quality.

\paragraph{Dimension 1: Semantic Alignment.} This dimension measures the semantic similarity between text and motion, evaluated through the following three metrics:

1. \textbf{Matching Score}: Calculates the Euclidean distance between text and motion features in a joint latent space.
\begin{equation}
\text{Matching Score} = \|\mathbf{t} - \mathbf{a}\|_2
\end{equation}
Here, \( \mathbf{t} \) represents the feature vector of the text, and \( \mathbf{a} \) represents the feature vector of the motion. The notation \( \|\cdot\|_2 \) denotes the L2-norm (Euclidean norm), which calculates the distance between the two feature vectors. 

2. \textbf{R-Precision}: Assesses the model's ability to retrieve the correct description from a set of motion candidates.

3. \textbf{Automatic Similarity Recall(ASR)} measures the recall rate of relevant atomic actions based on cosine similarity. The calculation process is as follows:
\begin{equation}
\text{ASR} = \frac{1}{N} \sum_{i=1}^{N} \mathbb{I}\left(\text{Cosine Similarity}(A_i, B_i) > 0.6\right)
\end{equation}
Where:
\begin{itemize}
    \item \( A_i \) and \( B_i \) are the feature vectors of the $i$-th atomic action from the Ground Truth prompt and the model output prompt, respectively.
    \item \( \mathbb{I}(\cdot) \) is the indicator function, which returns 1 if the cosine similarity exceeds 0.5, and 0 otherwise.
    \item \( N \) is the total number of atomic actions.
\end{itemize}

\paragraph{Dimension 2: Generalizability.} This dimension evaluates the model's ability to generalize across a variety of scenarios by assessing two key aspects:

1. \textbf{Multimodality(MM):} measures the model's ability to generate multiple distinct motions based on the same text input. Suppose we have a text input \( T \), and the model generates \( K \) different motion sequences \( A_1, A_2, \dots, A_K \) as outputs. The measurement of multimodality can be done by calculating the Euclidean distance between these output sequences and then averaging these distances. The formula for multimodality is as follows:

\begin{equation}
\text{MM} = \frac{1}{M} \sum_{m=1}^{M} \text{Euclidean Distance}(A_{i_m}, A_{j_m})
\end{equation}

Where \( A_{i_m} \) and \( A_{j_m} \) are two different motion sequences randomly selected from the \( K \) generated sequences, and \( \text{Euclidean Distance}(A_{i_m}, A_{j_m}) \) calculates the Euclidean distance between these two motion sequences, indicating the degree of difference between them. 

2. \textbf{Diversity:} measures the overall diversity of motions generated from the dataset. Suppose the model generates \( N \) different motion sequences \( A_1, A_2, \dots, A_N \) from different text inputs. The diversity is measured as follows:

\begin{enumerate}
    \item Randomly select \( M \) motion sequences from the \( N \) sequences as the first group \( \{A_{i_1}, A_{i_2}, \dots, A_{i_m}\} \).
    \item Randomly select \( M \) motion sequences from the \( N \) sequences as the second group \( \{A_{j_1}, A_{j_2}, \dots, A_{j_m}\} \).
    \item Compute the Euclidean distance between the \( M \) motion sequences from both groups.
    \item Take the average of these distances as the diversity metric.
\end{enumerate}

The formula for diversity is:

\begin{equation}
\text{Diversity} = \frac{1}{M} \sum_{m=1}^{M} \|A_{i_m} - A_{j_m}\|_2
\end{equation}

Where \( \|A_{i_m} - A_{j_m}\|_2 \) represents the Euclidean distance between two motion sequences. A higher diversity value indicates more diverse generated motions.

\paragraph{Dimension 3: Physical Quality.} This dimension evaluates the realism and physical plausibility of the generated motions.

1. \textbf{Jitter Degree(JD)} measures the amount of jitter in the generated motion sequence. It is calculated by evaluating the acceleration of keyframes and removing the jitter from local motion and global motion. The calculation of jitter degree is as follows:

\begin{equation}
\text{JD} = \frac{1}{(T-2) \times J} \sum_{t=1}^{T-2} \sum_{j=1}^{J} \left( \left\| \mathbf{a}_{t,j}^{\text{global}} \right\|_2 + \left\| \mathbf{a}_{t,j}^{\text{local}} \right\|_2 \right)
\end{equation}

Where \( \mathbf{a}_{t,j}^{\text{global}} = \mathbf{v}_{t+1,j}^{\text{global}} - \mathbf{v}_{t,j}^{\text{global}} \) and \( \mathbf{a}_{t,j}^{\text{local}} = \mathbf{v}_{t+1,j}^{\text{local}} - \mathbf{v}_{t,j}^{\text{local}} \) represent the global and local velocity of the \( j \)-th keyframe, respectively, \( \mathbf{v}_{t,j} \) is the velocity at the \( t \)-th frame, and \( \mathbf{p}_{t,j} \) is the position of the \( j \)-th keyframe. \( T \) is the total number of frames and \( J \) is the total number of keyframes. A smaller jitter degree indicates smoother motion.

2. \textbf{Ground Penetration(GP)} measures the extent to which parts of the motion sequence penetrate the ground. The calculation of ground penetration is as follows:

\begin{equation}
\text{GP} = \frac{1}{N} \sum_{i=1}^{N} |h_i| \quad \text{where} \quad h_i < \delta
\end{equation}

Where \( h_i = z_i - z_{\text{floor}} \) represents the height of the \( i \)-th keyframe from the ground, \( z_{\text{floor}} = 0 \) is the ground height, \( \delta = 0.005 \) (5mm) is the tolerance value, and \( N \) is the total number of keyframes. A smaller ground penetration value indicates better physical plausibility of the generated motion.

3. \textbf{Foot Floating(FF)} measures the degree of foot floating during the contact phase. It determines whether the foot is abnormally floating by detecting foot velocity, relative velocity, and contact state. The formula for calculating floating degree is as follows:

\begin{equation}
\text{FF} = \frac{1}{T} \left( N_{\text{invalid}} + \frac{1}{2} \sum_{r} L_r + \sum_{m} L_m \right)
\end{equation}

Where \( N_{\text{invalid}} \) is the number of invalid frames detected (based on foot velocity ratio \( r_t = \frac{\|\mathbf{v}_t^{\text{rel}}\|_2}{\|\mathbf{v}_t^{\text{root}}\|_2 + \epsilon} \) and contact state), \( L_r \) is the duration of floating intervals in the sequence, \( L_m \) is the duration of mass floating intervals, and \( T \) is the total number of frames. A smaller floating degree value indicates more reasonable foot-ground contact.

4. \textbf{Foot Sliding(FS)} measures the extent of foot sliding on the ground. It quantifies the horizontal movement of the foot when in contact with the ground. The formula for calculating foot sliding is as follows:

\begin{equation}
\text{FS} = \frac{1}{2} \left( \frac{\sum_{t=1}^{T} \|\mathbf{v}_{t,xy}^{\text{left}}\|_2 \cdot c_t^{\text{left}}}{\sum_{t=1}^{T} c_t^{\text{left}} + \epsilon} + \frac{\sum_{t=1}^{T} \|\mathbf{v}_{t,xy}^{\text{right}}\|_2 \cdot c_t^{\text{right}}}{\sum_{t=1}^{T} c_t^{\text{right}} + \epsilon} \right)
\end{equation}

Where \( \mathbf{v}_{t,xy} \) represents the horizontal velocity of the foot at frame \( t \) (considering only the x and y directions), \( c_t \in \{0, 1\} \) indicates the contact state, and \( \epsilon = 10^{-6} \) is a numerical stability term. A smaller foot sliding value indicates more stable foot contact.

5. \textbf{Dynamic Degree(DD)} measures the dynamic nature of the generated motion. It is quantified by calculating the \( L_2 \) norm of the joint velocities, including both global dynamics and local dynamics after removing the global translation. The formula for dynamic degree is as follows:

\begin{equation}
\text{DD} = \frac{1}{(T-1) \times J} \sum_{t=1}^{T-1} \sum_{j=1}^{J} \left( \|\mathbf{v}_{t,j}^{\text{global}}\|_2 + \|\mathbf{v}_{t,j}^{\text{local}}\|_2 \right)
\end{equation}

Where \( \mathbf{v}_{t,j}^{\text{global}} = \mathbf{p}_{t+1,j} - \mathbf{p}_{t,j} \) and \( \mathbf{v}_{t,j}^{\text{local}} = \mathbf{p}_{t+1,j}^{\text{local}} - \mathbf{p}_{t,j}^{\text{local}} \) represent the global velocity and local velocity of the \( j \)-th joint at frame \( t \), respectively (with the local velocity based on the joint position after removing the root translation), \( T \) is the total number of frames, and \( J \) is the number of joints. A larger dynamic degree value indicates that the motion is more dynamic.

6. \textbf{Position Quality(PQ)} measures the plausibility and naturalness of the generated poses. It is quantified by evaluating the distance from the pose parameters to the pose manifold using a Neural Distance Field (NRDF) model. The formula for pose quality is as follows:

\begin{equation}
\text{PQ} = 10 \times \frac{1}{T} \sum_{t=1}^{T} d_{\text{NRDF}}(\mathbf{q}_t)
\end{equation}

Where \( d_{\text{NRDF}}(\mathbf{q}_t) \) represents the distance predicted by the NRDF model for the pose quaternion \( \mathbf{q}_t \) at frame \( t \) (converted from the axis-angle \( \mathbf{p}_t \)), and \( T \) is the total number of frames. A higher pose quality value indicates that the pose is more natural and plausible.

7. \textbf{Body Penetration(BP)} measures the degree of penetration between different parts of the body. It is calculated by using the BVH (Bounding Volume Hierarchy) collision detection algorithm to detect collisions and overlaps between body parts. The formula for body penetration is as follows:

\begin{equation}
\text{BP} = \frac{1}{T} \sum_{t=1}^{T} \frac{N_t^{\text{collision}}}{N^{\text{triangles}}} \times 100\%
\end{equation}

Where \( N_t^{\text{collision}} \) represents the number of collision triangle pairs detected at frame \( t \) through BVH, \( N^{\text{triangles}} \) is the total number of triangles in the SMPL model (a fixed value), and \( T \) is the total number of frames. A smaller body penetration value indicates less self-collision in the body.

\subsection{Fine-grained accuracy evaluation}
\label{appendix:Fine-grained accuracy evaluation}
For a motion clip with $T$ frames indexed by $t\in\{0,\dots,T-1\}$, we define the start frame $t_0=0$ and the evaluation frame
\begin{equation}
t_e = \min\left(T-1,\ \max(0,\ T-N)\right),
\end{equation}
where $N$ denotes the window length (we use $N=30$ by default). 

\paragraph{(1) Whole-body motion assessment.}
This protocol evaluates whether the generated motion satisfies a specified \emph{root} control target from \texttt{root\_move.json}. We recover the root position sequence $p_t\in\mathbb{R}^3$ and the root yaw angle $\psi_t\in\mathbb{R}$ (rotation about the $Y$ axis) from the motion representation. For models outputting normalized HumanML/MDM features $x\in\mathbb{R}^{T\times 263}$, we optionally denormalize using mean $\mu$ and standard deviation $\sigma$:
\begin{equation}
\tilde x = x \odot \sigma + \mu,
\end{equation}
where $\odot$ is element-wise multiplication. Note that the HumanML/MDM 263-dim representation models root rotation with one degree of freedom (yaw); pitch/roll are not available and thus not evaluated.

\textbf{Root rotation (yaw).}
Given a target yaw change $\psi^{*}$ (in radians), we compute
\begin{equation}
\Delta\psi = \mathrm{wrap}_{[-\pi,\pi)}(\psi_{t_e}-\psi_{t_0}),
\end{equation}
and form the yaw rotation matrix
\begin{equation}
R_y(\theta)=
\begin{bmatrix}
\cos\theta & 0 & \sin\theta\\
0 & 1 & 0\\
-\sin\theta & 0 & \cos\theta
\end{bmatrix}.
\end{equation}
We define the rotation error as the Frobenius distance between $R_y(\Delta\psi)$ and $R_y(\psi^{*})$:
\begin{equation}
e_{\mathrm{rot}}=\left\lVert R_y(\Delta\psi)-R_y(\psi^{*})\right\rVert_F.
\end{equation}

\textbf{Directional root velocity.}
Given a target speed $v^{*}$ (m/s) along a unit direction $u$ for duration $d$ seconds, we estimate per-frame velocity using the frame rate $\mathrm{fps}$:
\begin{equation}
v_t=(p_{t+1}-p_t)\cdot \mathrm{fps},
\end{equation}
set $t_d=\min\left(T-1,\mathrm{round}(d\cdot \mathrm{fps})\right)$, and compute the mean projected speed
\begin{equation}
\hat v=\frac{1}{t_d}\sum_{t=0}^{t_d-1}\langle v_t,\ u\rangle.
\end{equation}
The velocity error is
\begin{equation}
e_{\mathrm{vel}}=\left|\hat v - v^{*}\right|.
\end{equation}

\textbf{Root translation.}
Given a target displacement $\Delta p^{*}\in\mathbb{R}^3$ (m), we compute $\Delta p=p_{t_e}-p_{t_0}$ and report a 3D RMSE over xyz:
\begin{equation}
e_{\mathrm{trans}}=\sqrt{\frac{(\Delta p_x-\Delta p^{*}_x)^2+(\Delta p_y-\Delta p^{*}_y)^2+(\Delta p_z-\Delta p^{*}_z)^2}{3}}.
\end{equation}

\paragraph{(2) Body-part motion assessment.}
This protocol evaluates the relative translation between a \emph{base} joint $b$ and a \emph{target} joint $g$ specified in \texttt{body\_part.json}. Given absolute joint coordinates $J\in\mathbb{R}^{T\times 22\times 3}$, the per-frame relative displacement is
\begin{equation}
\Delta_t = J_{t,g}-J_{t,b}\in\mathbb{R}^3,
\end{equation}
with target displacement $\Delta^{*}\in\mathbb{R}^3$. We compute $e_t=\Delta_t-\Delta^{*}$ and report the RMSE over the last $N$ frames $\mathcal{W}=\{T-N,\dots,T-1\}$ (or all frames if $T<N$):
\begin{equation}
\mathrm{RMSE}=\sqrt{\frac{1}{|\mathcal{W}|}\sum_{t\in\mathcal{W}}\lVert e_t\rVert_2^2}.
\end{equation}

\section{The physical quality scoring formula}
\label{appendix:The physical quality scoring formula}
For the LLM-based metric Physical Plausibility, we first normalize the value as \( g = \frac{x}{\text{max\_score}} \), where \( \text{max\_score} = 10 \).

For metrics that are better when smaller (e.g., Jitter Degree, Ground Penetration):
\begin{equation}
g = \text{clip} \left( \frac{P95 - x}{P95 - P5}, 0, 1 \right)
\end{equation}

For metrics that are better when larger (e.g., Dynamic Degree):
\begin{equation}
g = \text{clip} \left( \frac{x - P5}{P95 - P5}, 0, 1 \right)
\end{equation}

Where  \( P5 \) is the 5th percentile of all baselines under this metric, \( P95 \) is the 95th percentile of all baselines under this metric and \( x \) is the metric value of the sample.

The total score is calculated as:
\[
\text{Total} = \exp \left( \sum_j w_j \log(g_j + \epsilon) \right)
\]
Where \( w_j \) is the weight of the \( j \)-th metric, \( g_j \) is the normalized value of the \( j \)-th metric and \( \epsilon = 10^{-6} \) (to avoid log(0)).

The weight \( w \) allocation for each metric is as follows: Ground Penetration: 0.15, Foot Sliding: 0.15, Body Penetration: 0.15, Jitter Degree: 0.15, Foot Floating: 0.10, Pose Quality: 0.10, Dynamic Degree: 0.10 and Physical Plausibility from LLM-Based Evaluation: 0.10.

\section{Normalization Details}\label{app:normalization}
For each radar chart, we apply min--max normalization:
\begin{equation}
\mathrm{norm}(v)=\frac{v-\min}{\max-\min}.
\end{equation}
For the remaining higher-is-better metrics, we use the reversed normalization:
\begin{equation}
\mathrm{norm}(v)=\frac{\max-v}{\max-\min}.
\end{equation}
For each sub-metric, we first determine the integer-bounded range over the 14 baselines,
$[\,\min_{\text{eval}}, \max_{\text{eval}}\,]$, and compute:
\begin{equation}
\mathrm{norm}(v) = \frac{v - \min_{\text{eval}}}{\max_{\text{eval}} - \min_{\text{eval}}}.
\end{equation}

\section{LLM Evaluation Prompt}
\label{appendix:llm-prompt}

The following prompt is used for LLM-based evaluation of motion-text alignment. It guides the model to analyze sampled video frames and output a structured assessment in JSON format.

\begin{tcblisting}{
    listing engine=listings,    % 使用 listings 引擎处理文字
    listing only,               % 只显示文字内容，不执行代码
    breakable,                  % 核心：允许跨页
    enhanced,                   % 核心：允许高级排版
    sharp corners,              % 直角边框
    colback=gray!5,             % 背景颜色
    colframe=black!75,          % 边框颜色
    left=5pt, right=5pt,        % 内边距
    top=5pt, bottom=5pt,
    before skip=0pt,            % 紧贴标题，消除上方空白
    % --- 下面是 listings 的具体设置，解决自动换行和特殊字符 ---
    listing options={
        basicstyle=\ttfamily\small,
        breaklines=true,        % 核心：允许自动换行
        breakatwhitespace=false,
        columns=fullflexible,
        keepspaces=true,
        showstringspaces=false,
        literate={≤}{{$\le$}}1 {≥}{{$\ge$}}1 {–}{{-}}1 % 自动处理 Unicode 字符
    }
}
You are a motion alignment evaluator. Compare sequential frames with the provided motion prompt. Because only provided frames are available, judge solely by body motion. Ignore scenery, props, missing equipment, or environment differences. Treat the prompt as a list of action keywords and constraints (e.g., kick, pull, jump, push, rotate, crouch, left/right, direction, speed). Slowly scan every frame and concentrate on limb paths, joint angles, body orientation, tempo changes, and transitions.

The video is represented as a single long horizontal strip image. This strip is constructed by concatenating evenly sampled frames from left to right in strict temporal order:
* The leftmost region corresponds to the beginning of the motion and the rightmost region corresponds to the end.
When you analyze the motion, mentally scan the strip from left to right as if time is progressing, and focus on how the body pose and joint angles evolve along this horizontal axis.
The virtual camera is fixed in space. If the person becomes smaller, they are moving away; if larger, moving closer. At the initial state the person is facing the camera frontally.

Video Name: {video_name_escaped}
Motion Prompt Text: """{prompt_text_escaped}"""

Body-part color coding (for visual reference only):
(1) Head/scalp: pure red; (2) Face: orange; (3) Torso: bright yellow; (4) Left arm: pure green; (5) Right arm: cyan; (6) Left leg: pure blue; (7) Right leg: bright purple; (8) Left hand: light green; (9) Right hand: light gray.

====================================================================
 1) EXTRA NON-INSTRUCTION ACTIONS (10 points)
* 9-10: No obvious extra actions; motion stays on-instruction
* 5-8: Minor extra gestures/steps that do not dominate the sequence
* 0-4: Clear extra actions that change meaning or add major unintended motions

 2) ACTION COMPLETENESS (20 points)
* 18-20: All key action components appear clearly and sufficiently
* 10-17: Partial coverage; some components weak/short/missing
* 0-9: Most key actions missing, contradictory, or not recognizable

 3) MULTI-STAGE ORDER CORRECTNESS (10 points)
* 9-10: Correct order and clear stage boundaries
* 5-8: Order mostly correct but some stage confusion/overlap
* 0-4: Wrong order or stages not present

 4) BODY-PART UNDERSTANDING (10 points)
* 9-10: Correct body-part usage and orientation
* 5-8: Partially correct; occasional confusion
* 0-4: Mostly incorrect (e.g., wrong side/limb/direction)

 5) PHYSICAL PLAUSIBILITY(10 points)
* 9-10: Physically plausible, anatomically clean, and smooth/coherent throughout
* 5-8: Mostly acceptable with minor issues (small balance slips, jitter)
* 0-4: Major issues (teleporting, broken joints, jarring discontinuities)

====================================================================
 FINAL SCORING RULE
Compute: overall_score = sum of all 5 sub-scores (0-60)
Verdict: "aligned" (50-60), "partial" (30-49), "mismatch" (0-29)

====================================================================
OUTPUT FORMAT
Return ONLY the following raw JSON:
{
  "video_name": "{video_name_escaped}",
  "prompt_name": "string",
  "scores": {
    "extra_non_instruction_actions": int,
    "action_completeness": int,
    "multi_stage_order_correctness": int,
    "body_part_understanding": int,
    "physical_plausibility": int
  },
  "overall_score": int,
  "verdict": "aligned" | "partial" | "mismatch",
  "frame_observation": "<=200 chars",
  "prompt_overlap": "<=200 chars",
  "issues_found": "<=200 chars"
}
Important:
* Base judgment on visible evidence. Ignore background/props.
* Output plain JSON only. Do NOT wrap in markdown code blocks.
\end{tcblisting}

\section{Multi-Factor Motion Evaluation Detailed Results}
\label{Multi-Factor_detailed_results}
\begin{table*}[t]
\centering
\caption{The multi-factor motion evaluation results in the Semantic Alignment dimension. The metrics in this table represent the matching scores for 14 baseline models across the Dynamics, Complexity, and Interaction types, as well as the Accuracy type}
\label{tab:matching_scores}
\resizebox{\textwidth}{!}{
\begin{tabular}{@{}lcccccccc@{}}
\toprule
\multirow{2}{*}{\textbf{Method}} & \multicolumn{3}{c}{\textbf{Long}} & \multicolumn{3}{c}{\textbf{Short}} & \multirow{2}{*}{\textbf{Accuracy}} \\ 
\cmidrule(lr){2-4} \cmidrule(lr){5-7}
 & \textbf{Dynamics} & \textbf{Complexity} & \textbf{Interaction} & \textbf{Dynamics} & \textbf{Complexity} & \textbf{Interaction} & \\ \midrule
T2M & 5.0418 $\pm$ 0.0000 & 4.8300 $\pm$ 0.0000 & 4.4902 $\pm$ 0.0000 & 4.5640 $\pm$ 0.0000 & 3.7252 $\pm$ 0.0000 & 4.0692 $\pm$ 0.0000 & 6.6895 $\pm$ 0.0000 \\
MDM & 5.2138 $\pm$ 0.0000 & 4.5922 $\pm$ 0.0000 & 4.6730 $\pm$ 0.0000 & 4.0671 $\pm$ 0.0000 & 3.4628 $\pm$ 0.0000 & 4.1929 $\pm$ 0.0000 & 3.6187 $\pm$ 0.0000 \\
MotionGPT & 6.5503 $\pm$ 0.0000 & 6.0366 $\pm$ 0.0000 & 6.0473 $\pm$ 0.0000 & 5.3995 $\pm$ 0.0000 & 5.0574 $\pm$ 0.0000 & 4.9788 $\pm$ 0.0000 & 7.7820 $\pm$ 0.0000 \\
CLoSD & 5.7995 $\pm$ 0.0000 & 6.0593 $\pm$ 0.0000 & 4.5988 $\pm$ 0.0000 & 5.1634 $\pm$ 0.0000 & 5.3394 $\pm$ 0.0000 & 5.1380 $\pm$ 0.0000 & 5.7772 $\pm$ 0.0000 \\
PriorMDM & 4.3948 $\pm$ 0.0000 & 4.4662 $\pm$ 0.0000 & 4.6379 $\pm$ 0.0000 & 4.6881 $\pm$ 0.0000 & 4.3193 $\pm$ 0.0000 & 4.9758 $\pm$ 0.0000 & 7.8064 $\pm$ 0.0000 \\
MMM & 5.4966 $\pm$ 0.0000 & 4.8933 $\pm$ 0.0000 & 4.6808 $\pm$ 0.0000 & 6.3011 $\pm$ 0.0000 & 7.7051 $\pm$ 0.0000 & 7.0791 $\pm$ 0.0000 & 6.3699 $\pm$ 0.0000 \\
AvatarGPT & 8.8530 $\pm$ 0.0000 & 9.1317 $\pm$ 0.0000 & 7.8761 $\pm$ 0.0000 & 10.6632 $\pm$ 0.0000 & 9.2705 $\pm$ 0.0000 & 8.0676 $\pm$ 0.0000 & 8.9530 $\pm$ 0.0000 \\
Mogent & \color{red}9.0660 $\pm$ 0.0000 & 9.2064 $\pm$ 0.0000 & 7.9573 $\pm$ 0.0000 & 10.6679 $\pm$ 0.0000 & 9.2550 $\pm$ 0.0000 & 8.2126 $\pm$ 0.0000 & \color{blue}9.0780 $\pm$ 0.0000 \\
Salad & 9.0183 $\pm$ 0.0000 & 9.1966 $\pm$ 0.0000 & 7.8879 $\pm$ 0.0000 & 10.6496 $\pm$ 0.0000 & 9.2774 $\pm$ 0.0000 & 8.1153 $\pm$ 0.0000 & 5.4607 $\pm$ 0.0000 \\
Momask & 8.9607 $\pm$ 0.0000 & \color{blue}9.2234 $\pm$ 0.0000 & \color{blue}8.0139 $\pm$ 0.0000 & \color{blue}10.7979 $\pm$ 0.0000 & \color{blue}9.3066 $\pm$ 0.0000 & \color{blue}8.2379 $\pm$ 0.0000 & \color{red}9.1173 $\pm$ 0.0000 \\
Dart & 8.5231 $\pm$ 0.0000 & 8.8860 $\pm$ 0.0000 & 7.4521 $\pm$ 0.0000 & 10.2572 $\pm$ 0.0000 & 8.9925 $\pm$ 0.0000 & 7.7778 $\pm$ 0.0000 & 8.7389 $\pm$ 0.0000 \\
Vimogen & 7.8239 $\pm$ 0.0000 & 7.8120 $\pm$ 0.0000 & 6.2506 $\pm$ 0.0000 & 9.5702 $\pm$ 0.0000 & 8.4403 $\pm$ 0.0000 & 7.0386 $\pm$ 0.0000 & 6.2190 $\pm$ 0.0000 \\
HY & 6.8234 $\pm$ 0.0000 & 7.3276 $\pm$ 0.0000 & 6.1270 $\pm$ 0.0000 & 6.6704 $\pm$ 0.0000 & 6.8121 $\pm$ 0.0000 & 5.9780 $\pm$ 0.0000 & 7.1168 $\pm$ 0.0000 \\
MaskControl & \color{blue}9.0331 $\pm$ 0.0000 & \color{red}9.2354 $\pm$ 0.0000 & \color{red}8.0614 $\pm$ 0.0000 & \color{red}10.7983 $\pm$ 0.0000 & \color{red}9.3437 $\pm$ 0.0000 & \color{red}8.2438 $\pm$ 0.0000 & 5.6512 $\pm$ 0.0000 \\
\bottomrule
\end{tabular}
}
\end{table*}

\begin{table*}[t]
\centering
\caption{The multi-factor motion evaluation results in the Semantic Alignment dimension. The metrics in this table represent the R-Precision@1 for 14 baseline models across the Dynamics, Complexity, and Interaction types, as well as the Accuracy type}
\label{tab:R-Precision@1}
\resizebox{\textwidth}{!}{
\begin{tabular}{@{}lcccccccc@{}}
\toprule
\multirow{2}{*}{\textbf{Method}} & \multicolumn{3}{c}{\textbf{Long}} & \multicolumn{3}{c}{\textbf{Short}} & \multirow{2}{*}{\textbf{Accuracy}} \\ 
\cmidrule(lr){2-4} \cmidrule(lr){5-7}
 & \textbf{Dynamics} & \textbf{Complexity} & \textbf{Interaction} & \textbf{Dynamics} & \textbf{Complexity} & \textbf{Interaction} & \\ \midrule
T2M & 0.4138 $\pm$ 0.0101 & \color{blue}0.5452 $\pm$ 0.0115 & 0.4424 $\pm$ 0.0111 & \color{blue}0.5969 $\pm$ 0.0080 & \color{blue}0.7395 $\pm$ 0.0087 & \color{blue}0.6690 $\pm$ 0.0117 & 0.4097 $\pm$ 0.0079 \\
MDM & \color{blue}0.4211 $\pm$ 0.0099 & \color{red}0.5653 $\pm$ 0.0107 & \color{red}0.4684 $\pm$ 0.0114 & \color{red}0.6245 $\pm$ 0.0105 & \color{red}0.7887 $\pm$ 0.0097 & \color{red}0.6887 $\pm$ 0.0111 & \color{red}0.7474 $\pm$ 0.0077 \\
MotionGPT & 0.3196 $\pm$ 0.0098 & 0.4592 $\pm$ 0.0119 & 0.3647 $\pm$ 0.0095 & 0.4733 $\pm$ 0.0096 & 0.5811 $\pm$ 0.0103 & 0.5673 $\pm$ 0.0122 & 0.4331 $\pm$ 0.0072 \\
CLoSD & 0.3293 $\pm$ 0.0103 & 0.3897 $\pm$ 0.0121 & \color{blue}0.4536 $\pm$ 0.0121 & 0.5195 $\pm$ 0.0101 & 0.5056 $\pm$ 0.0117 & 0.5263 $\pm$ 0.0147 & 0.4983 $\pm$ 0.0077 \\
PriorMDM & \color{red}0.4420 $\pm$ 0.0087 & 0.5302 $\pm$ 0.0121 & 0.3587 $\pm$ 0.0117 & 0.5138 $\pm$ 0.0106 & 0.6527 $\pm$ 0.0099 & 0.5403 $\pm$ 0.0106 & 0.4216 $\pm$ 0.0069 \\
MMM & 0.3531 $\pm$ 0.0101 & 0.5182 $\pm$ 0.0108 & 0.3957 $\pm$ 0.0112 & 0.2669 $\pm$ 0.0104 & 0.3049 $\pm$ 0.0114 & 0.2717 $\pm$ 0.0092 & 0.4712 $\pm$ 0.0099 \\
AvatarGPT & 0.2880 $\pm$ 0.0056 & 0.2864 $\pm$ 0.0079 & 0.2717 $\pm$ 0.0070 & 0.2896 $\pm$ 0.0057 & 0.3145 $\pm$ 0.0057 & 0.2917 $\pm$ 0.0080 & 0.3153 $\pm$ 0.0066 \\
Mogent & 0.2740 $\pm$ 0.0055 & 0.2818 $\pm$ 0.0068 & 0.2813 $\pm$ 0.0068 & 0.2907 $\pm$ 0.0057 & 0.3289 $\pm$ 0.0069 & 0.3116 $\pm$ 0.0073 & 0.3114 $\pm$ 0.0061 \\
Salad & 0.2922 $\pm$ 0.0059 & 0.2990 $\pm$ 0.0068 & 0.2880 $\pm$ 0.0063 & 0.2864 $\pm$ 0.0053 & 0.3149 $\pm$ 0.0062 & 0.2900 $\pm$ 0.0069 & \color{blue}0.5745 $\pm$ 0.0082 \\
Momask & 0.2869 $\pm$ 0.0052 & 0.2657 $\pm$ 0.0066 & 0.2960 $\pm$ 0.0076 & 0.2805 $\pm$ 0.0060 & 0.3056 $\pm$ 0.0067 & 0.3010 $\pm$ 0.0083 & 0.3033 $\pm$ 0.0058 \\
Dart & 0.2873 $\pm$ 0.0079 & 0.2741 $\pm$ 0.0078 & 0.2757 $\pm$ 0.0064 & 0.2840 $\pm$ 0.0058 & 0.3131 $\pm$ 0.0070 & 0.3024 $\pm$ 0.0083 & 0.3230 $\pm$ 0.0060 \\
Vimogen & 0.2807 $\pm$ 0.0058 & 0.3030 $\pm$ 0.0071 & 0.2817 $\pm$ 0.0079 & 0.2616 $\pm$ 0.0063 & 0.2684 $\pm$ 0.0071 & 0.2837 $\pm$ 0.0089 & 0.4463 $\pm$ 0.0077 \\
HY & 0.3051 $\pm$ 0.0073 & 0.3122 $\pm$ 0.0089 & 0.2713 $\pm$ 0.0076 & 0.3547 $\pm$ 0.0082 & 0.4011 $\pm$ 0.0083 & 0.3911 $\pm$ 0.0124 & 0.3803 $\pm$ 0.0075 \\
MaskControl & 0.2807 $\pm$ 0.0051 & 0.2679 $\pm$ 0.0072 & 0.2984 $\pm$ 0.0072 & 0.2838 $\pm$ 0.0057 & 0.3209 $\pm$ 0.0070 & 0.3110 $\pm$ 0.0093 & 0.5490 $\pm$ 0.0081 \\
\bottomrule
\end{tabular}
}
\end{table*}

\begin{table*}[t]
\centering
\caption{The multi-factor motion evaluation results in the Semantic Alignment dimension. The metrics in this table represent the R-Precision@2 for 14 baseline models across the Dynamics, Complexity, and Interaction types, as well as the Accuracy type}
\label{tab:R-Precision@2}
\resizebox{\textwidth}{!}{
\begin{tabular}{@{}lcccccccc@{}}
\toprule
\multirow{2}{*}{\textbf{Method}} & \multicolumn{3}{c}{\textbf{Long}} & \multicolumn{3}{c}{\textbf{Short}} & \multirow{2}{*}{\textbf{Accuracy}} \\ 
\cmidrule(lr){2-4} \cmidrule(lr){5-7}
 & \textbf{Dynamics} & \textbf{Complexity} & \textbf{Interaction} & \textbf{Dynamics} & \textbf{Complexity} & \textbf{Interaction} & \\ \midrule
T2M & \color{blue}0.6764 $\pm$ 0.0099 & \color{red}0.8002 $\pm$ 0.0095 & 0.6990 $\pm$ 0.0120 & \color{blue}0.8213 $\pm$ 0.0080 & \color{blue}0.9184 $\pm$ 0.0066 & \color{blue}0.8677 $\pm$ 0.0078 & 0.6625 $\pm$ 0.0059 \\
MDM & 0.6629 $\pm$ 0.0084 & \color{blue}0.7987 $\pm$ 0.0098 & \color{red}0.7347 $\pm$ 0.0095 & \color{red}0.8542 $\pm$ 0.0077 & \color{red}0.9447 $\pm$ 0.0059 & \color{red}0.8946 $\pm$ 0.0082 & \color{red}0.9332 $\pm$ 0.0040 \\
MotionGPT & 0.5791 $\pm$ 0.0097 & 0.6984 $\pm$ 0.0094 & 0.6286 $\pm$ 0.0096 & 0.7282 $\pm$ 0.0084 & 0.7891 $\pm$ 0.0093 & 0.7780 $\pm$ 0.0104 & 0.6825 $\pm$ 0.0055 \\
CLoSD & 0.5996 $\pm$ 0.0108 & 0.6739 $\pm$ 0.0118 & \color{blue}0.7074 $\pm$ 0.0096 & 0.7725 $\pm$ 0.0080 & 0.7882 $\pm$ 0.0101 & 0.8060 $\pm$ 0.0114 & 0.7759 $\pm$ 0.0074 \\
PriorMDM & \color{red}0.7211 $\pm$ 0.0090 & 0.7746 $\pm$ 0.0094 & 0.6326 $\pm$ 0.0126 & 0.7635 $\pm$ 0.0079 & 0.8726 $\pm$ 0.0072 & 0.7640 $\pm$ 0.0106 & 0.6671 $\pm$ 0.0054 \\
MMM & 0.6082 $\pm$ 0.0109 & 0.7659 $\pm$ 0.0094 & 0.6750 $\pm$ 0.0116 & 0.5031 $\pm$ 0.0114 & 0.5457 $\pm$ 0.0095 & 0.5323 $\pm$ 0.0097 & 0.7140 $\pm$ 0.0081 \\
AvatarGPT & 0.5313 $\pm$ 0.0064 & 0.5517 $\pm$ 0.0089 & 0.5363 $\pm$ 0.0084 & 0.5627 $\pm$ 0.0068 & 0.5760 $\pm$ 0.0075 & 0.5556 $\pm$ 0.0085 & 0.5878 $\pm$ 0.0072 \\
Mogent & 0.5260 $\pm$ 0.0067 & 0.5330 $\pm$ 0.0073 & 0.5433 $\pm$ 0.0090 & 0.5551 $\pm$ 0.0067 & 0.6131 $\pm$ 0.0081 & 0.5720 $\pm$ 0.0107 & 0.5823 $\pm$ 0.0069 \\
Salad & 0.5533 $\pm$ 0.0063 & 0.5507 $\pm$ 0.0073 & 0.5500 $\pm$ 0.0075 & 0.5584 $\pm$ 0.0061 & 0.5989 $\pm$ 0.0067 & 0.5647 $\pm$ 0.0086 & \color{blue}0.8152 $\pm$ 0.0060 \\
Momask & 0.5331 $\pm$ 0.0065 & 0.5330 $\pm$ 0.0092 & 0.5427 $\pm$ 0.0077 & 0.5547 $\pm$ 0.0065 & 0.5718 $\pm$ 0.0064 & 0.5767 $\pm$ 0.0090 & 0.5758 $\pm$ 0.0066 \\
Dart & 0.5396 $\pm$ 0.0079 & 0.5170 $\pm$ 0.0091 & 0.5353 $\pm$ 0.0083 & 0.5571 $\pm$ 0.0062 & 0.5891 $\pm$ 0.0072 & 0.5754 $\pm$ 0.0088 & 0.5845 $\pm$ 0.0068 \\
Vimogen & 0.5344 $\pm$ 0.0061 & 0.5829 $\pm$ 0.0080 & 0.5343 $\pm$ 0.0098 & 0.5220 $\pm$ 0.0071 & 0.5338 $\pm$ 0.0072 & 0.5376 $\pm$ 0.0086 & 0.7137 $\pm$ 0.0079 \\
HY & 0.5602 $\pm$ 0.0083 & 0.5631 $\pm$ 0.0094 & 0.5400 $\pm$ 0.0095 & 0.6360 $\pm$ 0.0086 & 0.6735 $\pm$ 0.0074 & 0.6535 $\pm$ 0.0111 & 0.6583 $\pm$ 0.0070 \\
MaskControl & 0.5291 $\pm$ 0.0062 & 0.5425 $\pm$ 0.0069 & 0.5454 $\pm$ 0.0093 & 0.5545 $\pm$ 0.0064 & 0.5911 $\pm$ 0.0065 & 0.5820 $\pm$ 0.0080 & 0.8139 $\pm$ 0.0064 \\
\bottomrule
\end{tabular}
}
\end{table*}

\begin{table*}[t]
\centering
\caption{The multi-factor motion evaluation results in the Semantic Alignment dimension. The metrics in this table represent the R-Precision@3 for 14 baseline models across the Dynamics, Complexity, and Interaction types, as well as the Accuracy type}
\label{tab:R-Precision@3}
\resizebox{\textwidth}{!}{
\begin{tabular}{@{}lcccccccc@{}}
\toprule
\multirow{2}{*}{\textbf{Method}} & \multicolumn{3}{c}{\textbf{Long}} & \multicolumn{3}{c}{\textbf{Short}} & \multirow{2}{*}{\textbf{Accuracy}} \\ 
\cmidrule(lr){2-4} \cmidrule(lr){5-7}
 & \textbf{Dynamics} & \textbf{Complexity} & \textbf{Interaction} & \textbf{Dynamics} & \textbf{Complexity} & \textbf{Interaction} & \\ \midrule
T2M & 0.8544 $\pm$ 0.0072 & \color{blue}0.9245 $\pm$ 0.0064 & \color{blue}0.8846 $\pm$ 0.0080 & \color{blue}0.9313 $\pm$ 0.0051 & \color{blue}0.9789 $\pm$ 0.0038 & \color{blue}0.9584 $\pm$ 0.0056 & 0.8457 $\pm$ 0.0050 \\
MDM & \color{blue}0.8567 $\pm$ 0.0064 & \color{red}0.9308 $\pm$ 0.0066 & \color{red}0.9070 $\pm$ 0.0068 & \color{red}0.9571 $\pm$ 0.0046 & \color{red}0.9918 $\pm$ 0.0028 & \color{red}0.9764 $\pm$ 0.0050 & \color{red}0.9823 $\pm$ 0.0019 \\
MotionGPT & 0.7987 $\pm$ 0.0079 & 0.8576 $\pm$ 0.0088 & 0.8213 $\pm$ 0.0069 & 0.8824 $\pm$ 0.0063 & 0.9164 $\pm$ 0.0060 & 0.8930 $\pm$ 0.0066 & 0.8539 $\pm$ 0.0048 \\
CLoSD & 0.8167 $\pm$ 0.0079 & 0.8571 $\pm$ 0.0085 & 0.8677 $\pm$ 0.0074 & 0.9115 $\pm$ 0.0043 & 0.9316 $\pm$ 0.0054 & 0.9510 $\pm$ 0.0077 & 0.9221 $\pm$ 0.0048 \\
PriorMDM & \color{red}0.9009 $\pm$ 0.0070 & 0.9049 $\pm$ 0.0064 & 0.8534 $\pm$ 0.0090 & 0.9096 $\pm$ 0.0061 & 0.9622 $\pm$ 0.0043 & 0.8983 $\pm$ 0.0076 & 0.8566 $\pm$ 0.0043 \\
MMM & 0.8224 $\pm$ 0.0076 & 0.9059 $\pm$ 0.0081 & 0.8673 $\pm$ 0.0085 & 0.7458 $\pm$ 0.0107 & 0.7650 $\pm$ 0.0092 & 0.7663 $\pm$ 0.0095 & 0.8554 $\pm$ 0.0065 \\
AvatarGPT & 0.7671 $\pm$ 0.0053 & 0.7739 $\pm$ 0.0087 & 0.7810 $\pm$ 0.0057 & 0.7931 $\pm$ 0.0066 & 0.7962 $\pm$ 0.0061 & 0.7860 $\pm$ 0.0084 & 0.8215 $\pm$ 0.0051 \\
Mogent & 0.7671 $\pm$ 0.0054 & 0.7719 $\pm$ 0.0071 & 0.7810 $\pm$ 0.0070 & 0.7824 $\pm$ 0.0069 & 0.8242 $\pm$ 0.0065 & 0.7927 $\pm$ 0.0082 & 0.7993 $\pm$ 0.0060 \\
Salad & 0.7847 $\pm$ 0.0062 & 0.7819 $\pm$ 0.0091 & 0.7763 $\pm$ 0.0071 & 0.7951 $\pm$ 0.0055 & 0.8169 $\pm$ 0.0063 & 0.8000 $\pm$ 0.0082 & \color{blue}0.9384 $\pm$ 0.0035 \\
Momask & 0.7651 $\pm$ 0.0057 & 0.7728 $\pm$ 0.0078 & 0.7690 $\pm$ 0.0074 & 0.7800 $\pm$ 0.0065 & 0.7969 $\pm$ 0.0059 & 0.8056 $\pm$ 0.0075 & 0.8015 $\pm$ 0.0052 \\
Dart & 0.7698 $\pm$ 0.0064 & 0.7583 $\pm$ 0.0082 & 0.7677 $\pm$ 0.0075 & 0.7702 $\pm$ 0.0051 & 0.8124 $\pm$ 0.0065 & 0.8010 $\pm$ 0.0087 & 0.8106 $\pm$ 0.0055 \\
Vimogen & 0.7720 $\pm$ 0.0051 & 0.7992 $\pm$ 0.0074 & 0.7724 $\pm$ 0.0078 & 0.7778 $\pm$ 0.0067 & 0.7707 $\pm$ 0.0074 & 0.7683 $\pm$ 0.0089 & 0.8795 $\pm$ 0.0053 \\
HY & 0.7878 $\pm$ 0.0068 & 0.7793 $\pm$ 0.0083 & 0.7833 $\pm$ 0.0088 & 0.8467 $\pm$ 0.0075 & 0.8636 $\pm$ 0.0071 & 0.8550 $\pm$ 0.0089 & 0.8561 $\pm$ 0.0058 \\
MaskControl & 0.7633 $\pm$ 0.0064 & 0.7687 $\pm$ 0.0064 & 0.7747 $\pm$ 0.0079 & 0.7773 $\pm$ 0.0055 & 0.8002 $\pm$ 0.0071 & 0.7980 $\pm$ 0.0071 & 0.9377 $\pm$ 0.0032 \\ \bottomrule
\end{tabular}
}
\end{table*}

\begin{table*}[t]
\centering
\caption{The multi-factor motion evaluation results in the Semantic Alignment dimension. The metrics in this table represent the ASR for 14 baseline models across the Dynamics, Complexity, and Interaction types, as well as the Accuracy type}
\label{tab:ASR}
\resizebox{\textwidth}{!}{
\begin{tabular}{@{}lcccccccc@{}}
\toprule
\multirow{2}{*}{\textbf{Method}} & \multicolumn{3}{c}{\textbf{Long}} & \multicolumn{3}{c}{\textbf{Short}} & \multirow{2}{*}{\textbf{Accuracy}} \\ 
\cmidrule(lr){2-4} \cmidrule(lr){5-7}
 & \textbf{Dynamics} & \textbf{Complexity} & \textbf{Interaction} & \textbf{Dynamics} & \textbf{Complexity} & \textbf{Interaction} & \\ \midrule
T2M & 0.7395 $\pm$ 0.2722 & 0.7618 $\pm$ 0.3028 & 0.5783 $\pm$ 0.3747 & 0.1467 $\pm$ 0.3458 & 0.2867 $\pm$0.4552 & 0.0500 $\pm$ 0.2077 & 0.1964 $\pm$ 0.3193 \\
MDM & 0.7420 $\pm$ 0.2918 & 0.7465 $\pm$ 0.3077 & 0.6258 $\pm$ 0.3675 & 0.1600 $\pm$ 0.3616 & \color{blue}0.3267 $\pm$ 0.4686 & 0.1000 $\pm$ 0.2647 & 0.2393 $\pm$ 0.3572 \\
MotionGPT & 0.6581 $\pm$ 0.3021 & 0.7361 $\pm$ 0.2973 & 0.6116 $\pm$ 0.4031 & 0.1467 $\pm$ 0.3400 & 0.2800 $\pm$ 0.4509 & 0.0500 $\pm$ 0.1515 & 0.2232 $\pm$ 0.3393 \\
CLoSD & 0.6760 $\pm$ 0.2767 & 0.6736 $\pm$ 0.3494 & 0.6250 $\pm$ 0.4119 & 0.1467 $\pm$ 0.3256 & 0.2200 $\pm$ 0.3895 & 0.0900 $\pm$ 0.2559 & 0.1896 $\pm$ 0.3156 \\
PriorMDM & 0.7544 $\pm$ 0.2848 & 0.7944 $\pm$ 0.2612 & 0.5492 $\pm$ 0.4245 & \color{red}0.2067 $\pm$ 0.3813 & 0.2867 $\pm$ 0.4513 & 0.1000 $\pm$ 0.2647 & 0.2036 $\pm$ 0.3306 \\
MMM & 0.7665 $\pm$ 0.2674 & 0.7660 $\pm$ 0.2839 & 0.6150 $\pm$ 0.3527 & \color{blue}0.2067 $\pm$ 0.3899 & 0.2933 $\pm$ 0.4526 & 0.1000 $\pm$ 0.2647 & 0.2107 $\pm$ 03169 \\
AvatarGPT & 0.7325 $\pm$ 0.2723 & 0.7833 $\pm$ 0.2887 & 0.5816 $\pm$ 0.4118 & 0.1333 $\pm$ 0.3190 & 0.2933 $\pm$ 0.4591 & 0.1200 $\pm$ 0.3140 & 0.1964 $\pm$ 0.3094 \\
Mogent & 0.7621 $\pm$ 0.2713 & 0.7833 $\pm$ 0.2778 & 0.5750 $\pm$ 0.3743 & 0.1400 $\pm$ 0.3304 & 0.3067 $\pm$ 0.4636 & \color{blue}0.1200 $\pm$ 0.3010 & \color{blue}0.2411 $\pm$ 0.3592 \\
Salad & \color{blue}0.7792 $\pm$ 0.2601 & 0.7972 $\pm$ 0.2827 & 0.5950 $\pm$ 0.4198 & 0.1600 $\pm$ 0.3437 & 0.2933 $\pm$ 0.4591 & \color{red}0.1200 $\pm$ 0.2799 & 0.2357 $\pm$ 0.3742 \\
Momask & 0.7603 $\pm$ 0.2841 & 0.7694 $\pm$ 0.3164 & \color{red}0.6417 $\pm$ 0.3937 & 0.1267 $\pm$ 0.3339 & 0.3067 $\pm$ 0.4650 & 0.0600 $\pm$ 0.2222 & 0.1929 $\pm$ 0.3186 \\
Dart & 0.7147 $\pm$ 0.2806 & 0.7674 $\pm$ 0.2871 & 0.6216 $\pm$ 0.4001 & 0.1000 $\pm$ 0.2893 & 0.2333 $\pm$ 0.4223 & 0.0200 $\pm$ 0.0989 & 0.1607 $\pm$ 0.3166 \\
Vimogen & 0.7588 $\pm$ 0.2609 & 0.7542 $\pm$ 0.2962 & 0.5616 $\pm$ 0.3347 & 0.1533 $\pm$ 0.3485 & \color{red}0.3467 $\pm$ 0.4803 & 0.1000 $\pm$ 0.2841 & \color{red}0.2446 $\pm$ 0.3559 \\
HY & 0.7751 $\pm$ 0.2363 & \color{red}0.8125 $\pm$ 0.2789 & \color{blue}0.6392 $\pm$ 0.3856 & 0.2000 $\pm$ 0.3831 & 0.2933 $\pm$ 0.4598 & 0.0752 $\pm$ 0.2409 & 0.1982 $\pm$ 0.3346 \\
MaskControl & \color{red}0.7969 $\pm$ 0.2369 & \color{blue}0.8000 $\pm$ 0.2669 & 0.5975 $\pm$ 0.3898 & 0.1667 $\pm$ 0.3563 & 0.2800 $\pm$ 0.4488 & 0.0800 $\pm$ 0.2460 & 0.2161 $\pm$ 0.3362\\ \bottomrule
\end{tabular}
}
\end{table*}

\begin{table*}[t]
\centering
\caption{The multi-factor motion evaluation results in the Generalizability dimension. The metrics in this table represent the multimodality for 14 baseline models across the Dynamics, Complexity, and Interaction types, as well as the Accuracy type}
\label{tab:multimodality}
\resizebox{\textwidth}{!}{
\begin{tabular}{@{}lcccccccc@{}}
\toprule
\multirow{2}{*}{\textbf{Method}} & \multicolumn{3}{c}{\textbf{Long}} & \multicolumn{3}{c}{\textbf{Short}} & \multirow{2}{*}{\textbf{Accuracy}} \\ 
\cmidrule(lr){2-4} \cmidrule(lr){5-7}
 & \textbf{Dynamics} & \textbf{Complexity} & \textbf{Interaction} & \textbf{Dynamics} & \textbf{Complexity} & \textbf{Interaction} & \\ \midrule
T2M & \color{blue}2.9352 $\pm$ 0.0974 & 3.0402 $\pm$ 0.1070 & 2.5601 $\pm$ 0.0797 & \color{blue}3.3040 $\pm$ 0.0889 & \color{blue}3.3372 $\pm$ 0.0884 & 3.4207 $\pm$ 0.0941 & 2.7628 $\pm$ 0.0860 \\
MDM & 2.5017 $\pm$ 0.0850 & 2.8831 $\pm$ 0.1039 & 2.3213 $\pm$ 0.0713 & 3.1636 $\pm$ 0.0800 & 3.0491 $\pm$ 0.0834 & 3.3417 $\pm$ 0.0920 & 2.0420 $\pm$ 0.0632 \\
MotionGPT & \color{red}5.9299 $\pm$ 0.2122 & \color{red}5.3763 $\pm$ 0.1909 & \color{red}5.5518 $\pm$ 0.2152 & \color{red}3.6639 $\pm$ 0.1204 & \color{red}3.4587 $\pm$ 0.1146 & \color{blue}3.8737 $\pm$ 0.1201 & \color{red}5.0421 $\pm$ 0.1748 \\
CLoSD & 2.3604 $\pm$ 0.0754 & 1.9826 $\pm$ 0.0681 & 2.2837 $\pm$ 0.0722 & 3.0562 $\pm$ 0.0836 & 2.8389 $\pm$ 0.0752 & 2.8386 $\pm$ 0.0760 & 2.1370 $\pm$ 0.0661 \\
PriorMDM & 2.7868 $\pm$ 0.0963 & 2.5319 $\pm$ 0.0866 & 3.2992 $\pm$ 0.0890 & 3.1942 $\pm$ 0.0886 & 3.2740 $\pm$ 0.0858 & 3.6233 $\pm$ 0.0910 & 2.8958 $\pm$ 0.0811 \\
MMM & 1.9071 $\pm$ 0.0596 & 2.3130 $\pm$ 0.0825 & 2.2922 $\pm$ 0.0733 & 2.9123 $\pm$ 0.0744 & 2.8336 $\pm$ 0.0784 & 2.7275 $\pm$ 0.0719 & 2.1521 $\pm$ 0.0597 \\
AvatarGPT & 2.6664 $\pm$ 0.1196 & \color{blue}4.0477 $\pm$ 0.1619 & \color{blue}4.5186 $\pm$ 0.1533 & 2.8391 $\pm$ 0.1152 & 3.3097 $\pm$ 0.1358 & \color{red}4.5187 $\pm$ 0.1533 & \color{blue}2.9940 $\pm$ 0.1180 \\
Mogent & 0.6697 $\pm$ 0.0335 & 0.6355 $\pm$ 0.0324 & 0.5121 $\pm$ 0.0244 & 2.6432 $\pm$ 0.1205 & 2.5903 $\pm$ 0.1208 & 2.7607 $\pm$ 0.1313 & 0.9871 $\pm$ 0.0488 \\
Salad & 0.6908 $\pm$ 0.0222 & 0.5589 $\pm$ 0.0165 & 0.5891 $\pm$ 0.0181 & 2.2421 $\pm$ 0.0667 & 2.0698 $\pm$ 0.0627 & 2.3974 $\pm$ 0.0820 & 0.7321 $\pm$ 0.0217 \\
Momask & 0.8892 $\pm$ 0.0384 & 0.9272 $\pm$ 0.0493 & 0.5251 $\pm$ 0.0227 & 2.7028 $\pm$ 0.1080 & 2.7319 $\pm$ 0.1233 & 2.7807 $\pm$ 0.1177 & 1.0361 $\pm$ 0.0482 \\
Dart & 1.9461 $\pm$ 0.0620 & 2.0776 $\pm$ 0.0696 & 2.5796 $\pm$ 0.1180 & 2.0346 $\pm$ 0.0615 & 2.3957 $\pm$ 0.0757 & 1.5979 $\pm$ 0.0484 & 1.7578 $\pm$ 0.0625 \\
Vimogen & 1.5159 $\pm$ 0.0427 & 1.4609 $\pm$ 0.0499 & 1.1261 $\pm$ 0.0312 & 1.4201 $\pm$ 0.0397 & 1.4192 $\pm$ 0.0448 & 1.1203 $\pm$ 0.0313 & 1.4952 $\pm$ 0.0391 \\
HY & 1.1208 $\pm$ 0.0377 & 0.5941 $\pm$ 0.0194 & 0.7560 $\pm$ 0.0250 & 1.5471 $\pm$ 0.0488 & 0.7461 $\pm$ 0.0252 & 0.9628 $\pm$ 0.0301 & 0.6362 $\pm$ 0.0218 \\
MaskControl & 0.8772 $\pm$ 0.0379 & 0.9607 $\pm$ 0.0516 & 0.5269 $\pm$ 0.0215 & 2.1488 $\pm$ 0.0851 & 2.8983 $\pm$ 0.1105 & 2.8263 $\pm$ 0.0985 & 1.0623 $\pm$ 0.0467 \\ \bottomrule
\end{tabular}
}
\end{table*}

\begin{table*}[t]
\centering
\caption{The multi-factor motion evaluation results in the Generalizability dimension. The metrics in this table represent the diversity for 14 baseline models across the Dynamics, Complexity, and Interaction types, as well as the Accuracy type}
\label{tab:Generalizability}
\resizebox{\textwidth}{!}{
\begin{tabular}{@{}lcccccccc@{}}
\toprule
\multirow{2}{*}{\textbf{Method}} & \multicolumn{3}{c}{\textbf{Long}} & \multicolumn{3}{c}{\textbf{Short}} & \multirow{2}{*}{\textbf{Accuracy}} \\ 
\cmidrule(lr){2-4} \cmidrule(lr){5-7}
 & \textbf{Dynamics} & \textbf{Complexity} & \textbf{Interaction} & \textbf{Dynamics} & \textbf{Complexity} & \textbf{Interaction} & \\ \midrule
T2M & 5.3956 $\pm$ 0.1293 & 6.8108 $\pm$ 0.1995 & 5.2286 $\pm$ 0.1223 & 7.2323 $\pm$ 0.1263 & 7.8616 $\pm$ 0.1577 & 6.2809 $\pm$ 0.1528 & 3.2387 $\pm$ 0.0722 \\
MDM & \color{blue}6.5693 $\pm$ 0.1640 & \color{blue}7.0042 $\pm$ 0.1706 & \color{blue}7.1154 $\pm$ 0.1777 & 5.1184 $\pm$ 0.1115 & \color{red}8.9342 $\pm$ 0.1994 & 5.9357 $\pm$ 0.1778 & \color{red}8.9097 $\pm$ 0.1722 \\
MotionGPT & \color{red}7.7573 $\pm$ 0.2240 & \color{red}8.3491 $\pm$ 0.1815 & \color{red}7.9223 $\pm$ 0.1760 & \color{blue}7.7850 $\pm$ 0.1534 & \color{blue}8.5559 $\pm$ 0.1725 & \color{blue}7.7001 $\pm$ 0.1956 & 3.0275 $\pm$ 0.0733 \\
CLoSD & 4.6157 $\pm$ 0.1178 & 5.0038 $\pm$ 0.1467 & 4.8469 $\pm$ 0.1292 & 6.2406 $\pm$ 0.1301 & 6.4810 $\pm$ 0.1347 & 5.8160 $\pm$ 0.1592 & 6.0418 $\pm$ 0.1090 \\
PriorMDM & 5.2105 $\pm$ 0.1424 & 6.1256 $\pm$ 0.1893 & 5.4950 $\pm$ 0.1595 & 6.4080 $\pm$ 0.1364 & 6.9921 $\pm$ 0.1545 & 7.0518 $\pm$ 0.1595 & 2.6231 $\pm$ 0.0768 \\
MMM & 5.0604 $\pm$ 0.1337 & 6.1839 $\pm$ 0.1617 & 5.8877 $\pm$ 0.1804 & 1.4552 $\pm$ 0.0484 & 6.2352 $\pm$ 0.1406 & 1.0665 $\pm$ 0.0280 & \color{blue}8.4176 $\pm$ 0.1457 \\
AvatarGPT & 1.6607 $\pm$ 0.0870 & 0.9446 $\pm$ 0.0463 & 1.1291 $\pm$ 0.0258 & 1.3151 $\pm$ 0.0511 & 1.2942 $\pm$ 0.0557 & 1.2047 $\pm$ 0.0543 & 1.1845 $\pm$ 0.0433 \\
Mogent & 1.3393 $\pm$ 0.0603 & 1.1663 $\pm$ 0.0553 & 1.0990 $\pm$ 0.0480 & 1.3254 $\pm$ 0.0440 & 1.4622 $\pm$ 0.0728 & 1.0335 $\pm$ 0.0313 & 1.1845 $\pm$ 0.0404 \\
Salad & 1.1558 $\pm$ 0.0326 & 0.9006 $\pm$ 0.0251 & 1.0459 $\pm$ 0.0283 & 1.1858 $\pm$ 0.0435 & 1.0924 $\pm$ 0.0456 & 1.1421 $\pm$ 0.0327 & 5.6588 $\pm$ 0.1106 \\
Momask & 1.6219 $\pm$ 0.0686 & 1.6402 $\pm$ 0.0766 & 1.1420 $\pm$ 0.0370 & 1.3595 $\pm$ 0.0428 & 1.4873 $\pm$ 0.0585 & 1.1287 $\pm$ 0.0367 & 1.0450 $\pm$ 0.0296 \\
Dart & 0.9947 $\pm$ 0.0415 & 0.6505 $\pm$ 0.0192 & 0.5748 $\pm$ 0.0223 & 0.8979 $\pm$ 0.0303 & 0.7252 $\pm$ 0.0257 & 0.6399 $\pm$ 0.0382 & 0.6297 $\pm$ 0.0178 \\
Vimogen & 1.3654 $\pm$ 0.0365 & 1.4038 $\pm$ 0.0386 & 1.5231 $\pm$ 0.0439 & 1.6118 $\pm$ 0.0341 & 1.6524 $\pm$ 0.0498 & 1.6433 $\pm$ 0.0374 & 5.4882 $\pm$ 0.1462 \\
HY & 3.2903 $\pm$ 0.1388 & 2.9222 $\pm$ 0.1131 & 2.9099 $\pm$ 0.1119 & 4.6197 $\pm$ 0.1151 & 3.1782 $\pm$ 0.0900 & 3.0222 $\pm$ 0.1090 & 2.8880 $\pm$ 0.0806 \\
MaskControl & 1.6145 $\pm$ 0.0684 & 1.6045 $\pm$ 0.0652 & 1.0831 $\pm$ 0.0424 & \color{red}7.8735 $\pm$ 0.1645 & 1.4596 $\pm$ 0.0585 & \color{red}8.3909 $\pm$ 0.2147 & 5.9217 $\pm$ 0.1245 \\ \bottomrule
\end{tabular}
}
\end{table*}

\begin{table*}[t]
\centering
\caption{The multi-factor motion evaluation results in the Physical Quality dimension. The metrics in this table represent the Jitter Degree for 14 baseline models across the Dynamics, Complexity, and Interaction types, as well as the Accuracy type}
\label{tab:Dynamics}
\resizebox{\textwidth}{!}{
\begin{tabular}{@{}lcccccccc@{}}
\toprule
\multirow{2}{*}{\textbf{Method}} & \multicolumn{3}{c}{\textbf{Long}} & \multicolumn{3}{c}{\textbf{Short}} & \multirow{2}{*}{\textbf{Accuracy}} \\ 
\cmidrule(lr){2-4} \cmidrule(lr){5-7}
 & \textbf{Dynamics} & \textbf{Complexity} & \textbf{Interaction} & \textbf{Dynamics} & \textbf{Complexity} & \textbf{Interaction} & \\ \midrule
T2M & \color{red}0.0046 $\pm$ 0.0011 & \color{blue}0.0042 $\pm$ 0.0009 & 0.0042 $\pm$ 0.0013 & \color{blue}0.0077 $\pm$ 0.0025 & 0.0059 $\pm$ 0.0022 & 0.0046 $\pm$ 0.0019 & 0.0067 $\pm$ 0.0029 \\
MDM & 0.0153 $\pm$ 0.0071 & 0.0081 $\pm$ 0.0053 & 0.0119 $\pm$ 0.0093 & 0.0186 $\pm$ 0.0090 & 0.0099 $\pm$ 0.0049 & 0.0063 $\pm$ 0.0048 & 0.0092 $\pm$ 0.0042 \\
MotionGPT & 0.0144 $\pm$ 0.0044 & 0.0133 $\pm$ 0.0057 & 0.0138 $\pm$ 0.0064 & 0.0206 $\pm$ 0.0081 & 0.0154 $\pm$ 0.0054 & 0.0120 $\pm$ 0.0048 & 0.0127 $\pm$ 0.0045 \\
CLoSD & 0.0136 $\pm$ 0.0056 & 0.0090 $\pm$ 0.0046 & 0.0134 $\pm$ 0.0092 & 0.0180 $\pm$ 0.0077 & 0.0142 $\pm$ 0.0065 & 0.0119 $\pm$ 0.0083 & 0.0148 $\pm$ 0.0066 \\
PriorMDM & 0.0077 $\pm$ 0.0055 & 0.0056 $\pm$ 0.0039 & 0.0070 $\pm$ 0.0064 & 0.0229 $\pm$ 0.0127 & 0.0106 $\pm$ 0.0074 & 0.0083 $\pm$ 0.0070 & 0.0128 $\pm$ 0.0068 \\
MMM & 0.0185 $\pm$ 0.0053 & 0.0145 $\pm$ 0.0043 & 0.0135 $\pm$ 0.0051 & 0.0201 $\pm$ 0.0068 & 0.0135 $\pm$ 0.0053 & 0.0097 $\pm$ 0.0047 & 0.0115 $\pm$ 0.0041 \\
AvatarGPT & 0.0121 $\pm$ 0.0033 & 0.0084 $\pm$ 0.0023 & 0.0112 $\pm$ 0.0051 & 0.0201 $\pm$ 0.0085 & 0.0137 $\pm$ 0.0048 & 0.0086 $\pm$ 0.0040 & 0.0116 $\pm$ 0.0037 \\
Mogent & 0.0133 $\pm$ 0.0047 & 0.0102 $\pm$ 0.0039 & 0.0090 $\pm$ 0.0057 & 0.0184 $\pm$ 0.0064 & 0.0111 $\pm$ 0.0042 & 0.0084 $\pm$ 0.0052 & 0.0113 $\pm$ 0.0043 \\
Salad & 0.0165 $\pm$ 0.0037 & 0.0115 $\pm$ 0.0031 & 0.0129 $\pm$ 0.0045 & 0.0169 $\pm$ 0.0050 & 0.0125 $\pm$ 0.0034 & 0.0100 $\pm$ 0.0043 & 0.0125 $\pm$ 0.0029 \\
Momask & 0.0194 $\pm$ 0.0061 & 0.0126 $\pm$ 0.0042 & 0.0105 $\pm$ 0.0068 & 0.0203 $\pm$ 0.0062 & 0.0133 $\pm$ 0.0049 & 0.0103 $\pm$ 0.0062 & 0.0137 $\pm$ 0.0048 \\
Dart & \color{blue}0.0053 $\pm$ 0.0015 & 0.0047 $\pm$ 0.0010 & \color{red}0.0038 $\pm$ 0.0011 & \color{red}0.0075 $\pm$ 0.0028 & \color{blue}0.0053 $\pm$ 0.0022 & \color{red}0.0034 $\pm$ 0.0017 & \color{red}0.0051 $\pm$ 0.0019 \\
Vimogen & 0.0155 $\pm$ 0.0026 & 0.0136 $\pm$ 0.0026 & 0.0126 $\pm$ 0.0038 & 0.0164 $\pm$ 0.0025 & 0.0154 $\pm$ 0.0025 & 0.0152 $\pm$ 0.0031 & 0.0139 $\pm$ 0.0024 \\
HY & 0.0082 $\pm$ 0.0057 & \color{red}0.0036 $\pm$ 0.0035 & \color{blue}0.0040 $\pm$ 0.0033 & 0.0185 $\pm$ 0.0135 & \color{red}0.0049 $\pm$ 0.0026 & \color{blue}0.0037 $\pm$ 0.0029 & \color{blue}0.0061 $\pm$ 0.0042 \\
MaskControl & 0.0207 $\pm$ 0.0052 & 0.0164 $\pm$ 0.0046 & 0.0135 $\pm$ 0.0059 & 0.0222 $\pm$ 0.0059 & 0.0155 $\pm$ 0.0053 & 0.0124 $\pm$ 0.0061 & 0.0139 $\pm$ 0.0042 \\ \bottomrule
\end{tabular}
}
\end{table*}

\begin{table*}[t]
\centering
\caption{The multi-factor motion evaluation results in the Physical Quality dimension. The metrics in this table represent the Ground Penetration for 14 baseline models across the Dynamics, Complexity, and Interaction types, as well as the Accuracy type}
\label{tab:Ground_Penetration}
\resizebox{\textwidth}{!}{
\begin{tabular}{@{}lcccccccc@{}}
\toprule
\multirow{2}{*}{\textbf{Method}} & \multicolumn{3}{c}{\textbf{Long}} & \multicolumn{3}{c}{\textbf{Short}} & \multirow{2}{*}{\textbf{Accuracy}} \\ 
\cmidrule(lr){2-4} \cmidrule(lr){5-7}
 & \textbf{Dynamics} & \textbf{Complexity} & \textbf{Interaction} & \textbf{Dynamics} & \textbf{Complexity} & \textbf{Interaction} & \\ \midrule
T2M & \color{blue}0.0395 $\pm$ 0.0480 & \color{blue}0.0416 $\pm$ 0.0463 & \color{blue}0.0340 $\pm$ 0.0489 & \color{blue}0.0806 $\pm$ 0.1078 & 0.1167 $\pm$ 0.2102 & 0.0364 $\pm$ 0.0795 & 0.0869 $\pm$ 0.1031 \\
MDM & 0.1649 $\pm$ 0.1337 & 0.0752 $\pm$ 0.0898 & 0.1135 $\pm$ 0.1700 & 0.2347 $\pm$ 0.2790 & 0.1355 $\pm$ 0.1899 & 0.0887 $\pm$ 0.1883 & 0.1104 $\pm$ 0.1857 \\
MotionGPT & 0.1633 $\pm$ 0.1736 & 0.1168 $\pm$ 0.1235 & 0.2009 $\pm$ 0.2797 & 0.1937 $\pm$ 0.1774 & 0.1564 $\pm$ 0.1638 & 0.1864 $\pm$ 0.2537 & 0.1328 $\pm$ 0.1375 \\
CLoSD & 0.1273 $\pm$ 0.2055 & 0.0762 $\pm$ 0.1411 & 0.1085 $\pm$ 0.2435 & 0.2225 $\pm$ 0.3232 & 0.2160 $\pm$ 0.3624 & 0.1040 $\pm$ 0.2963 & 0.1485 $\pm$ 0.3053 \\
PriorMDM & 0.1578 $\pm$ 0.1469 & 0.1189 $\pm$ 0.0794 & 0.0959 $\pm$ 0.1401 & 0.1977 $\pm$ 0.1424 & 0.1406 $\pm$ 0.1825 & 0.0911 $\pm$ 0.1571 & 0.1896 $\pm$ 0.2247 \\
MMM & 0.1350 $\pm$ 0.1161 & 0.0764 $\pm$ 0.0823 & 0.0444 $\pm$ 0.0777 & 0.2007 $\pm$ 0.1603 & 0.1193 $\pm$ 0.1431 & 0.0510 $\pm$ 0.0669 & 0.0817 $\pm$ 0.1030 \\
AvatarGPT & 0.4377 $\pm$ 0.2441 & 0.2389 $\pm$ 0.1044 & 0.1888 $\pm$ 0.1575 & 0.1790 $\pm$ 0.2156 & 0.1818 $\pm$ 0.2512 & 0.1266 $\pm$ 0.1536 & 0.0835 $\pm$ 0.1117 \\
Mogent & 0.1352 $\pm$ 0.1164 & 0.1114 $\pm$ 0.0675 & 0.0757 $\pm$ 0.0613 & 0.1522 $\pm$ 0.1277 & 0.1010 $\pm$ 0.0914 & 0.0730 $\pm$ 0.1162 & 0.0775 $\pm$ 0.0817 \\
Salad & 0.1568 $\pm$ 0.1061 & 0.1297 $\pm$ 0.0961 & 0.1019 $\pm$ 0.0832 & 0.1571 $\pm$ 0.1097 & 0.1365 $\pm$ 0.1235 & 0.0885 $\pm$ 0.0985 & 0.0900 $\pm$ 0.1029 \\
Momask & 0.1759 $\pm$ 0.0961 & 0.1216 $\pm$ 0.0895 & 0.0781 $\pm$ 0.0902 & 0.1968 $\pm$ 0.1996 & 0.1322 $\pm$ 0.1379 & 0.1186 $\pm$ 0.1596 & 0.0911 $\pm$ 0.0834 \\
Dart & 0.2694 $\pm$ 0.4843 & 0.2165 $\pm$ 0.1897 & 0.1600 $\pm$ 0.2527 & 0.1342 $\pm$ 0.1781 & \color{blue}0.0675 $\pm$ 0.1376 & \color{blue}0.0146 $\pm$ 0.0298 & \color{blue}0.0759 $\pm$ 0.0985 \\
Vimogen & \color{red}0.0045 $\pm$ 0.0032 & \color{red}0.0045 $\pm$ 0.0030 & \color{red}0.0051 $\pm$ 0.0046 & \color{red}0.0052 $\pm$ 0.0040 & \color{red}0.0051 $\pm$ 0.0043 & \color{red}0.0040 $\pm$ 0.0039 & \color{red}0.0041 $\pm$ 0.0033 \\
HY & 0.1453 $\pm$ 0.0903 & 0.0910 $\pm$ 0.0647 & 0.0709 $\pm$ 0.0572 & 0.1500 $\pm$ 0.0728 & 0.0939 $\pm$ 0.0613 & 0.0562 $\pm$ 0.0490 & 0.1051 $\pm$ 0.1149 \\
MaskControl & 0.1578 $\pm$ 0.0810 & 0.1216 $\pm$ 0.0829 & 0.0903 $\pm$ 0.0979 & 0.1843 $\pm$ 0.1609 & 0.1192 $\pm$ 0.1313 & 0.1056 $\pm$ 0.1462 & 0.1153 $\pm$ 0.1118 \\ \bottomrule
\end{tabular}
}
\end{table*}

\begin{table*}[t]
\centering
\caption{The multi-factor motion evaluation results in the Physical Quality dimension. The metrics in this table represent the Foot Floating for 14 baseline models across the Dynamics, Complexity, and Interaction types, as well as the Accuracy type}
\label{tab:Physical_Quality}
\resizebox{\textwidth}{!}{
\begin{tabular}{@{}lcccccccc@{}}
\toprule
\multirow{2}{*}{\textbf{Method}} & \multicolumn{3}{c}{\textbf{Long}} & \multicolumn{3}{c}{\textbf{Short}} & \multirow{2}{*}{\textbf{Accuracy}} \\ 
\cmidrule(lr){2-4} \cmidrule(lr){5-7}
 & \textbf{Dynamics} & \textbf{Complexity} & \textbf{Interaction} & \textbf{Dynamics} & \textbf{Complexity} & \textbf{Interaction} & \\ \midrule
T2M & 0.1296 $\pm$ 0.0512 & 0.1487 $\pm$ 0.0624 & 0.1040 $\pm$ 0.0535 & 0.1384 $\pm$ 0.0489 & 0.1401 $\pm$ 0.0801 & 0.1058 $\pm$ 0.0456 & 0.1038 $\pm$ 0.0450 \\
MDM & \color{blue}0.0877 $\pm$ 0.0427 & \color{blue}0.0743 $\pm$ 0.0392 & \color{blue}0.0570 $\pm$ 0.0452 & \color{blue}0.0809 $\pm$ 0.0422 & \color{blue}0.0601 $\pm$ 0.0354 & \color{blue}0.0340 $\pm$ 0.0298 & 0.0799 $\pm$ 0.0421 \\
MotionGPT & 0.2621 $\pm$ 0.0848 & 0.2439 $\pm$ 0.0935 & 0.2232 $\pm$ 0.0948 & 0.2503 $\pm$ 0.0851 & 0.2264 $\pm$ 0.0932 & 0.2087 $\pm$ 0.0996 & 0.1958 $\pm$ 0.0770 \\
CLoSD & \color{red}0.0337 $\pm$ 0.0258 & \color{red}0.0240 $\pm$ 0.0140 & \color{red}0.0290 $\pm$ 0.0225 & \color{red}0.0479 $\pm$ 0.0294 & \color{red}0.0403 $\pm$ 0.0331 & \color{red}0.0336 $\pm$ 0.0497 & \color{red}0.0429 $\pm$ 0.0324 \\
PriorMDM & 0.1764 $\pm$ 0.0760 & 0.1958 $\pm$ 0.0831 & 0.1442 $\pm$ 0.0812 & 0.1413 $\pm$ 0.0842 & 0.1395 $\pm$ 0.0694 & 0.0935 $\pm$ 0.0556 & 0.1230 $\pm$ 0.0650 \\
MMM & 0.2066 $\pm$ 0.0532 & 0.2129 $\pm$ 0.0629 & 0.1704 $\pm$ 0.0605 & 0.1671 $\pm$ 0.0523 & 0.1732 $\pm$ 0.0695 & 0.1398 $\pm$ 0.0648 & 0.1429 $\pm$ 0.0569 \\
AvatarGPT & 0.2739 $\pm$ 0.0509 & 0.2817 $\pm$ 0.0562 & 0.2185 $\pm$ 0.0796 & 0.1974 $\pm$ 0.0626 & 0.1899 $\pm$ 0.0816 & 0.1832 $\pm$ 0.1015 & 0.1709 $\pm$ 0.0634 \\
Mogent & 0.2176 $\pm$ 0.0661 & 0.2350 $\pm$ 0.0821 & 0.1602 $\pm$ 0.0721 & 0.1917 $\pm$ 0.0806 & 0.1590 $\pm$ 0.0715 & 0.1603 $\pm$ 0.0944 & 0.1525 $\pm$ 0.0660 \\
Salad & 0.3079 $\pm$ 0.0694 & 0.3622 $\pm$ 0.0875 & 0.2883 $\pm$ 0.0983 & 0.2894 $\pm$ 0.0766 & 0.3117 $\pm$ 0.0893 & 0.3395 $\pm$ 0.0982 & 0.3057 $\pm$ 0.0678 \\
Momask & 0.3028 $\pm$ 0.0902 & 0.2448 $\pm$ 0.0862 & 0.1776 $\pm$ 0.0874 & 0.1955 $\pm$ 0.0666 & 0.1545 $\pm$ 0.0638 & 0.1265 $\pm$ 0.0708 & 0.1729 $\pm$ 0.0713 \\
Dart & 0.0909 $\pm$ 0.0278 & 0.0986 $\pm$ 0.0293 & 0.0596 $\pm$ 0.0233 & 0.1080 $\pm$ 0.0640 & 0.0848 $\pm$ 0.0406 & 0.0527 $\pm$ 0.0309 & \color{blue}0.0722 $\pm$ 0.0377 \\
Vimogen & 0.1898 $\pm$ 0.0510 & 0.2119 $\pm$ 0.0498 & 0.1911 $\pm$ 0.0632 & 0.2447 $\pm$ 0.0722 & 0.2408 $\pm$ 0.0786 & 0.2525 $\pm$ 0.0847 & 0.2558 $\pm$ 0.0666 \\
HY & 0.2398 $\pm$ 0.1070 & 0.1945 $\pm$ 0.0995 & 0.2306 $\pm$ 0.1513 & 0.4003 $\pm$ 0.1266 & 0.2835 $\pm$ 0.1442 & 0.2178 $\pm$ 0.1586 & 0.1847 $\pm$ 0.0936 \\
MaskControl & 0.3125 $\pm$ 0.0901 & 0.2621 $\pm$ 0.0838 & 0.2005 $\pm$ 0.0676 & 0.2176 $\pm$ 0.0719 & 0.1923 $\pm$ 0.0707 & 0.1631 $\pm$ 0.0784 & 0.2095 $\pm$ 0.0734 \\ \bottomrule
\end{tabular}
}
\end{table*}

\begin{table*}[t]
\centering
\caption{The multi-factor motion evaluation results in the Physical Quality dimension. The metrics in this table represent the Foot Sliding for 14 baseline models across the Dynamics, Complexity, and Interaction types, as well as the Accuracy type}
\label{tab:Foot_Sliding}
\resizebox{\textwidth}{!}{
\begin{tabular}{@{}lcccccccc@{}}
\toprule
\multirow{2}{*}{\textbf{Method}} & \multicolumn{3}{c}{\textbf{Long}} & \multicolumn{3}{c}{\textbf{Short}} & \multirow{2}{*}{\textbf{Accuracy}} \\ 
\cmidrule(lr){2-4} \cmidrule(lr){5-7}
 & \textbf{Dynamics} & \textbf{Complexity} & \textbf{Interaction} & \textbf{Dynamics} & \textbf{Complexity} & \textbf{Interaction} & \\ \midrule
T2M & \color{red}0.0030 $\pm$ 0.0012 & 0.0032 $\pm$ 0.0015 & \color{blue}0.0022 $\pm$ 0.0009 & \color{red}0.0045 $\pm$ 0.0023 & 0.0043 $\pm$ 0.0029 & 0.0026 $\pm$ 0.0013 & 0.0039 $\pm$ 0.0024 \\
MDM & 0.0069 $\pm$ 0.0049 & 0.0045 $\pm$ 0.0043 & 0.0047 $\pm$ 0.0049 & 0.0092 $\pm$ 0.0094 & 0.0043 $\pm$ 0.0039 & 0.0022 $\pm$ 0.0025 & 0.0034 $\pm$ 0.0025 \\
MotionGPT & 0.0065 $\pm$ 0.0032 & 0.0064 $\pm$ 0.0042 & 0.0067 $\pm$ 0.0055 & 0.0099 $\pm$ 0.0061 & 0.0075 $\pm$ 0.0040 & 0.0057 $\pm$ 0.0040 & 0.0057 $\pm$ 0.0030 \\
CLoSD & 0.0030 $\pm$ 0.0023 & \color{red}0.0020 $\pm$ 0.0015 & 0.0027 $\pm$ 0.0028 & \color{blue}0.0048 $\pm$ 0.0038 & \color{red}0.0035 $\pm$ 0.0028 & \color{blue}0.0022 $\pm$ 0.0018 & \color{blue}0.0032 $\pm$ 0.0028 \\
PriorMDM & 0.0059 $\pm$ 0.0053 & 0.0041 $\pm$ 0.0030 & 0.0037 $\pm$ 0.0048 & 0.0111 $\pm$ 0.0092 & 0.0059 $\pm$ 0.0064 & 0.0032 $\pm$ 0.0033 & 0.0072 $\pm$ 0.0060 \\
MMM & 0.0090 $\pm$ 0.0056 & 0.0060 $\pm$ 0.0035 & 0.0040 $\pm$ 0.0017 & 0.0101 $\pm$ 0.0060 & 0.0063 $\pm$ 0.0037 & 0.0037 $\pm$ 0.0022 & 0.0049 $\pm$ 0.0032 \\
AvatarGPT & 0.0064 $\pm$ 0.0030 & 0.0041 $\pm$ 0.0013 & 0.0047 $\pm$ 0.0030 & 0.0092 $\pm$ 0.0066 & 0.0062 $\pm$ 0.0033 & 0.0036 $\pm$ 0.0021 & 0.0043 $\pm$ 0.0019 \\
Mogent & 0.0073 $\pm$ 0.0034 & 0.0067 $\pm$ 0.0041 & 0.0037 $\pm$ 0.0024 & 0.0091 $\pm$ 0.0047 & 0.0058 $\pm$ 0.0042 & 0.0039 $\pm$ 0.0027 & 0.0049 $\pm$ 0.0031 \\
Salad & 0.0117 $\pm$ 0.0055 & 0.0080 $\pm$ 0.0041 & 0.0063 $\pm$ 0.0037 & 0.0085 $\pm$ 0.0059 & 0.0069 $\pm$ 0.0028 & 0.0044 $\pm$ 0.0018 & 0.0073 $\pm$ 0.0031 \\
Momask & 0.0129 $\pm$ 0.0069 & 0.0076 $\pm$ 0.0053 & 0.0041 $\pm$ 0.0031 & 0.0088 $\pm$ 0.0055 & 0.0065 $\pm$ 0.0042 & 0.0041 $\pm$ 0.0026 & 0.0059 $\pm$ 0.0036 \\
Dart & \color{blue}0.0030 $\pm$ 0.0020 & \color{blue}0.0030 $\pm$ 0.0014 & \color{red}0.0019 $\pm$ 0.0012 & 0.0058 $\pm$ 0.0041 & \color{blue}0.0040 $\pm$ 0.0036 & \color{red}0.0014 $\pm$ 0.0009 & \color{red}0.0032 $\pm$ 0.0017 \\
Vimogen & 0.0076 $\pm$ 0.0024 & 0.0079 $\pm$ 0.0029 & 0.0065 $\pm$ 0.0026 & 0.0105 $\pm$ 0.0032 & 0.0099 $\pm$ 0.0030 & 0.0082 $\pm$ 0.0025 & 0.0101 $\pm$ 0.0033 \\
HY & 0.0099 $\pm$ 0.0065 & 0.0048 $\pm$ 0.0030 & 0.0054 $\pm$ 0.0057 & 0.0196 $\pm$ 0.0162 & 0.0066 $\pm$ 0.0042 & 0.0043 $\pm$ 0.0049 & 0.0047 $\pm$ 0.0037 \\
MaskControl & 0.0121 $\pm$ 0.0061 & 0.0093 $\pm$ 0.0060 & 0.0050 $\pm$ 0.0041 & 0.0099 $\pm$ 0.0062 & 0.0070 $\pm$ 0.0053 & 0.0043 $\pm$ 0.0022 & 0.0058 $\pm$ 0.0033 \\ \bottomrule
\end{tabular}
}
\end{table*}

\begin{table*}[t]
\centering
\caption{The multi-factor motion evaluation results in the Physical Quality dimension. The metrics in this table represent the Dynamic Degree for 14 baseline models across the Dynamics, Complexity, and Interaction types, as well as the Accuracy type}
\label{tab:Dynamic_Degree}
\resizebox{\textwidth}{!}{
\begin{tabular}{@{}lcccccccc@{}}
\toprule
\multirow{2}{*}{\textbf{Method}} & \multicolumn{3}{c}{\textbf{Long}} & \multicolumn{3}{c}{\textbf{Short}} & \multirow{2}{*}{\textbf{Accuracy}} \\ 
\cmidrule(lr){2-4} \cmidrule(lr){5-7}
 & \textbf{Dynamics} & \textbf{Complexity} & \textbf{Interaction} & \textbf{Dynamics} & \textbf{Complexity} & \textbf{Interaction} & \\ \midrule
T2M & 0.0171 $\pm$ 0.0046 & 0.0153 $\pm$ 0.0042 & 0.0143 $\pm$ 0.0050 & 0.0257 $\pm$ 0.0078 & 0.0213 $\pm$ 0.0079 & 0.0171 $\pm$ 0.0078 & 0.0233 $\pm$ 0.0088 \\
MDM & 0.0355 $\pm$ 0.0122 & 0.0225 $\pm$ 0.0093 & 0.0285 $\pm$ 0.0159 & 0.0433 $\pm$ 0.0167 & 0.0296 $\pm$ 0.0111 & 0.0192 $\pm$ 0.0111 & 0.0259 $\pm$ 0.0092 \\
MotionGPT & 0.0293 $\pm$ 0.0100 & 0.0272 $\pm$ 0.0116 & \color{blue}0.0288 $\pm$ 0.0144 & 0.0409 $\pm$ 0.0177 & 0.0325 $\pm$ 0.0118 & \color{blue}0.0268 $\pm$ 0.0132 & 0.0295 $\pm$ 0.0111 \\
CLoSD & 0.0248 $\pm$ 0.0099 & 0.0170 $\pm$ 0.0085 & 0.0234 $\pm$ 0.0164 & 0.0330 $\pm$ 0.0141 & 0.0283 $\pm$ 0.0143 & 0.0224 $\pm$ 0.0162 & 0.0304 $\pm$ 0.0144 \\
PriorMDM & 0.0221 $\pm$ 0.0113 & 0.0187 $\pm$ 0.0102 & 0.0208 $\pm$ 0.0125 & \color{red}0.0525 $\pm$ 0.0220 & 0.0297 $\pm$ 0.0151 & 0.0249 $\pm$ 0.0168 & \color{red}0.0374 $\pm$ 0.0152 \\
MMM & 0.0365 $\pm$ 0.0100 & 0.0298 $\pm$ 0.0080 & 0.0268 $\pm$ 0.0096 & 0.0394 $\pm$ 0.0120 & 0.0300 $\pm$ 0.0103 & 0.0214 $\pm$ 0.0105 & 0.0264 $\pm$ 0.0096 \\
AvatarGPT & 0.0253 $\pm$ 0.0065 & 0.0177 $\pm$ 0.0045 & 0.0233 $\pm$ 0.0122 & 0.0408 $\pm$ 0.0177 & 0.0308 $\pm$ 0.0111 & 0.0183 $\pm$ 0.0097 & 0.0277 $\pm$ 0.0098 \\
Mogent & 0.0370 $\pm$ 0.0107 & 0.0294 $\pm$ 0.0087 & 0.0238 $\pm$ 0.0110 & 0.0438 $\pm$ 0.0127 & 0.0315 $\pm$ 0.0098 & 0.0236 $\pm$ 0.0120 & 0.0309 $\pm$ 0.0109 \\
Salad & 0.0390 $\pm$ 0.0086 & \color{blue}0.0301 $\pm$ 0.0082 & \color{red}0.0288 $\pm$ 0.0102 & 0.0392 $\pm$ 0.0104 & 0.0320 $\pm$ 0.0093 & 0.0234 $\pm$ 0.0104 & 0.0299 $\pm$ 0.0079 \\
Momask & \color{blue}0.0408 $\pm$ 0.0121 & 0.0295 $\pm$ 0.0098 & 0.0222 $\pm$ 0.0132 & 0.0460 $\pm$ 0.0131 & \color{blue}0.0328 $\pm$ 0.0117 & 0.0263 $\pm$ 0.0165 & 0.0326 $\pm$ 0.0124 \\
Dart & 0.0133 $\pm$ 0.0047 & 0.0116 $\pm$ 0.0030 & 0.0091 $\pm$ 0.0035 & 0.0214 $\pm$ 0.0088 & 0.0156 $\pm$ 0.0073 & 0.0081 $\pm$ 0.0062 & 0.0148 $\pm$ 0.0067 \\
Vimogen & 0.0302 $\pm$ 0.0053 & 0.0280 $\pm$ 0.0053 & 0.0227 $\pm$ 0.0068 & 0.0319 $\pm$ 0.0048 & 0.0290 $\pm$ 0.0049 & 0.0264 $\pm$ 0.0056 & 0.0261 $\pm$ 0.0052 \\
HY & 0.0233 $\pm$ 0.0089 & 0.0129 $\pm$ 0.0062 & 0.0136 $\pm$ 0.0089 & 0.0435 $\pm$ 0.0200 & 0.0181 $\pm$ 0.0075 & 0.0139 $\pm$ 0.0096 & 0.0174 $\pm$ 0.0087 \\
MaskControl & \color{red}0.0420 $\pm$ 0.0111 & \color{red}0.0346 $\pm$ 0.0103 & 0.0268 $\pm$ 0.0119 & \color{blue}0.0480 $\pm$ 0.0129 & \color{red}0.0350 $\pm$ 0.0119 & \color{red}0.0281 $\pm$ 0.0160 & \color{blue}0.0339 $\pm$ 0.0115 \\ \bottomrule
\end{tabular}
}
\end{table*}

\begin{table*}[t]
\centering
\caption{The multi-factor motion evaluation results in the Physical Quality dimension. The metrics in this table represent the pose quality for 14 baseline models across the Dynamics, Complexity, and Interaction types, as well as the Accuracy type}
\label{tab:pose_quality}
\resizebox{\textwidth}{!}{
\begin{tabular}{@{}lcccccccc@{}}
\toprule
\multirow{2}{*}{\textbf{Method}} & \multicolumn{3}{c}{\textbf{Long}} & \multicolumn{3}{c}{\textbf{Short}} & \multirow{2}{*}{\textbf{Accuracy}} \\ 
\cmidrule(lr){2-4} \cmidrule(lr){5-7}
 & \textbf{Dynamics} & \textbf{Complexity} & \textbf{Interaction} & \textbf{Dynamics} & \textbf{Complexity} & \textbf{Interaction} & \\ \midrule
T2M & 2.0667 $\pm$ 0.1894 & 1.9423 $\pm$ 0.1653 & 2.0374 $\pm$ 0.1972 & 2.1909 $\pm$ 0.2206 & 2.0441 $\pm$ 0.2580 & 1.9433 $\pm$ 0.2080 & 2.0737 $\pm$ 0.1594 \\
MDM & 2.3354 $\pm$ 0.3272 & 2.0269 $\pm$ 0.2092 & 2.4132 $\pm$ 0.4034 & 2.1770 $\pm$ 0.2566 & 2.0185 $\pm$ 0.2460 & 2.1223 $\pm$ 0.4816 & 1.8929 $\pm$ 0.2113 \\
MotionGPT & 2.3042 $\pm$ 0.3033 & 2.2648 $\pm$ 0.2994 & 2.3872 $\pm$ 0.4741 & 2.4127 $\pm$ 0.3249 & 2.1749 $\pm$ 0.2965 & 2.1551 $\pm$ 0.3139 & 2.0752 $\pm$ 0.2581 \\
CLoSD & 1.8286 $\pm$ 0.1480 & 1.6683 $\pm$ 0.1276 & 1.8133 $\pm$ 0.1655 & 1.7840 $\pm$ 0.1870 & 1.6804 $\pm$ 0.1901 & 1.7963 $\pm$ 0.2339 & 1.6374 $\pm$ 0.1332 \\
PriorMDM & 2.7155 $\pm$ 0.2903 & 2.6538 $\pm$ 0.3414 & 2.4847 $\pm$ 0.3638 & 2.5735 $\pm$ 0.2967 & 2.3386 $\pm$ 0.2998 & 2.2188 $\pm$ 0.3828 & 2.2444 $\pm$ 0.2176 \\
MMM & 2.1417 $\pm$ 0.1881 & 2.0483 $\pm$ 0.1578 & 2.1479 $\pm$ 0.3096 & 2.2049 $\pm$ 0.2236 & 2.0271 $\pm$ 0.2288 & 1.9730 $\pm$ 0.2721 & 1.9563 $\pm$ 0.1894 \\
AvatarGPT & 2.1451 $\pm$ 0.2068 & 1.9888 $\pm$ 0.1967 & 2.2132 $\pm$ 0.3913 & 2.2723 $\pm$ 0.3358 & 2.0981 $\pm$ 0.3258 & 2.0284 $\pm$ 0.4334 & 1.9960 $\pm$ 0.2807 \\
Mogent & 2.4895 $\pm$ 0.2200 & 2.3929 $\pm$ 0.2423 & 2.4316 $\pm$ 0.3516 & 2.4091 $\pm$ 0.2874 & 2.2051 $\pm$ 0.2941 & 2.1532 $\pm$ 0.3136 & 1.9875 $\pm$ 0.1929 \\
Salad & 2.6171 $\pm$ 0.2348 & 2.2139 $\pm$ 0.2276 & 2.4996 $\pm$ 0.3153 & 2.3332 $\pm$ 0.2934 & 2.1310 $\pm$ 0.2209 & 2.1906 $\pm$ 0.3404 & 2.0110 $\pm$ 0.1851 \\
Momask & 2.5119 $\pm$ 0.2348 & 2.3773 $\pm$ 0.2795 & 2.3395 $\pm$ 0.3685 & 2.3481 $\pm$ 0.2668 & 2.1446 $\pm$ 0.2975 & 2.0739 $\pm$ 0.3131 & 2.0940 $\pm$ 0.2769 \\
Dart & \color{red}3.3518 $\pm$ 0.1531 & \color{red}3.2598 $\pm$ 0.1284 & \color{red}3.3950 $\pm$ 0.1740 & \color{red}3.5097 $\pm$ 0.1843 & \color{red}3.4433 $\pm$ 0.1913 & \color{red}3.3956 $\pm$ 0.2037 & \color{red}3.3429 $\pm$ 0.1366 \\
Vimogen & \color{blue}2.8190 $\pm$ 0.1467 & \color{blue}2.6615 $\pm$ 0.1412 & \color{blue}2.7540 $\pm$ 0.2762 & \color{blue}2.8800 $\pm$ 0.1500 & \color{blue}2.8024 $\pm$ 0.1516 & \color{blue}2.7597 $\pm$ 0.2222 & \color{blue}2.5419 $\pm$ 0.1527 \\
HY & 2.5910 $\pm$ 0.2041 & 2.2355 $\pm$ 0.1639 & 2.4452 $\pm$ 0.2285 & 2.6578 $\pm$ 0.2671 & 2.2754 $\pm$ 0.2879 & 2.5052 $\pm$ 0.4399 & 2.1158 $\pm$ 0.2174 \\
MaskControl & 2.4824 $\pm$ 0.2096 & 2.3788 $\pm$ 0.2570 & 2.3182 $\pm$ 0.3736 & 2.3876 $\pm$ 0.2680 & 2.1713 $\pm$ 0.3068 & 2.1004 $\pm$ 0.2959 & 2.0625 $\pm$ 0.2590 \\ \bottomrule
\end{tabular}
}
\end{table*}

\begin{table*}[t]
\centering
\caption{The multi-factor motion evaluation results in the Physical Quality dimension. The metrics in this table represent the body penetration for 14 baseline models across the Dynamics, Complexity, and Interaction types, as well as the Accuracy type}
\label{tab:body_penetration}
\resizebox{\textwidth}{!}{
\begin{tabular}{@{}lcccccccc@{}}
\toprule
\multirow{2}{*}{\textbf{Method}} & \multicolumn{3}{c}{\textbf{Long}} & \multicolumn{3}{c}{\textbf{Short}} & \multirow{2}{*}{\textbf{Accuracy}} \\ 
\cmidrule(lr){2-4} \cmidrule(lr){5-7}
 & \textbf{Dynamics} & \textbf{Complexity} & \textbf{Interaction} & \textbf{Dynamics} & \textbf{Complexity} & \textbf{Interaction} & \\ \midrule
T2M & \color{red}0.6351 $\pm$ 0.2381 & \color{red}0.5959 $\pm$ 0.1731 & \color{red}0.6566 $\pm$ 0.2912 & \color{blue}0.7744 $\pm$ 0.2296 & \color{blue}0.9430 $\pm$ 0.2751 & \color{red}0.9689 $\pm$ 0.5013 & 1.1542 $\pm$ 0.2424 \\
MDM & 1.0152 $\pm$ 0.3299 & 0.8719 $\pm$ 0.3178 & 1.3737 $\pm$ 0.7549 & 0.9723 $\pm$ 0.3081 & 1.1572 $\pm$ 0.3436 & 1.2857 $\pm$ 0.6559 & 1.1375 $\pm$ 0.2625 \\
MotionGPT & 1.0051 $\pm$ 0.3536 & 1.1743 $\pm$ 0.4768 & 1.1491 $\pm$ 0.4595 & 1.2211 $\pm$ 0.4397 & 1.3208 $\pm$ 0.4572 & 1.2759 $\pm$ 0.5751 & 1.3313 $\pm$ 0.3392 \\
CLoSD & 0.7123 $\pm$ 0.1830 & 0.8175 $\pm$ 0.1659 & \color{blue}0.7368 $\pm$ 0.1900 & 0.9820 $\pm$ 0.2114 & 1.0226 $\pm$ 0.2412 & \color{blue}1.0370 $\pm$ 0.4297 & \color{red}1.1126 $\pm$ 0.1711 \\
PriorMDM & 1.2310 $\pm$ 0.3618 & 1.3722 $\pm$ 0.4260 & 1.1266 $\pm$ 0.4573 & 1.2064 $\pm$ 0.2909 & 1.1387 $\pm$ 0.3453 & 1.2649 $\pm$ 0.5788 & 1.2994 $\pm$ 0.3111 \\
MMM & \color{blue}0.7064 $\pm$ 0.2045 & \color{blue}0.7240 $\pm$ 0.1983 & 0.8215 $\pm$ 0.2921 & 0.9219 $\pm$ 0.2216 & 1.1022 $\pm$ 0.3608 & 1.3214 $\pm$ 0.6261 & 1.2084 $\pm$ 0.3030 \\
AvatarGPT & 1.3163 $\pm$ 0.2847 & 1.3720 $\pm$ 0.2396 & 1.3620 $\pm$ 0.6736 & 1.1199 $\pm$ 0.2946 & 1.2966 $\pm$ 0.4039 & 1.4851 $\pm$ 0.6674 & 1.4226 $\pm$ 0.4125 \\
Mogent & 1.0407 $\pm$ 0.3844 & 0.9922 $\pm$ 0.2801 & 1.4445 $\pm$ 0.7385 & 1.3049 $\pm$ 0.3443 & 1.2789 $\pm$ 0.3449 & 1.4798 $\pm$ 0.7504 & 1.5315 $\pm$ 0.4429 \\
Salad & 1.2600 $\pm$ 0.3610 & 1.0602 $\pm$ 0.3427 & 1.3942 $\pm$ 0.7246 & 1.3252 $\pm$ 0.4106 & 1.3215 $\pm$ 0.4275 & 1.5561 $\pm$ 0.7961 & 1.3684 $\pm$ 0.3598 \\
Momask & 0.9611 $\pm$ 0.3193 & 0.7902 $\pm$ 0.2846 & 1.2879 $\pm$ 0.7569 & 1.1458 $\pm$ 0.3363 & 1.1175 $\pm$ 0.3051 & 1.4171 $\pm$ 0.7959 & 1.3472 $\pm$ 0.3529 \\
Dart & 1.5524 $\pm$ 0.2903 & 1.4158 $\pm$ 0.2842 & 1.5046 $\pm$ 0.3763 & 1.3854 $\pm$ 0.3881 & 1.6936 $\pm$ 0.4689 & 1.9970 $\pm$ 0.5237 & 1.5495 $\pm$ 0.3481 \\
Vimogen & 1.1247 $\pm$ 0.3912 & 1.2131 $\pm$ 0.3484 & 1.3998 $\pm$ 0.6030 & 1.1241 $\pm$ 0.3610 & 1.2679 $\pm$ 0.4192 & 1.2410 $\pm$ 0.5968 & 1.4497 $\pm$ 0.3510 \\
HY & 1.4784 $\pm$ 0.4847 & 1.2824 $\pm$ 0.4059 & 1.2479 $\pm$ 0.7941 & \color{red}0.7187 $\pm$ 0.2941 & \color{red}0.9361 $\pm$ 0.5012 & 1.1749 $\pm$ 0.9201 & \color{blue}1.1359 $\pm$ 0.3957 \\
MaskControl & 0.9412 $\pm$ 0.2785 & 0.8157 $\pm$ 0.2651 & 1.1104 $\pm$ 0.5600 & 1.1597 $\pm$ 0.2833 & 1.1364 $\pm$ 0.3266 & 1.2935 $\pm$ 0.5870 & 1.3121 $\pm$ 0.3138 \\ \bottomrule
\end{tabular}
}
\end{table*}

\section{Detailed results of the PHC+ tracking on the released physical attribute dataset}

\begin{table}[ht]
\centering
\caption{PHC+ Tracking Detailed Results}
\label{tab:tracker_result}
\begin{tabular}{|c|c|c|c|c|c|c|}
\hline
\multicolumn{2}{|c|}{\textbf{Datasets}} & \textbf{mpjpe\_g} & \textbf{mpjpe\_l} & \textbf{mpjpe\_pa} & \textbf{accel\_dist} & \textbf{vel\_dist} \\
\hline

\multirow{2}{*}{\textbf{Dynamic}} &
\textbf{ours}   & 47.6132 & 29.6650 & 23.0038 & 6.0149  & 7.9921 \\
\cline{2-7}
& \textbf{random} & 95.5738 & 58.0841 & 44.2800 & 13.8798 & 15.5182 \\
\hline

\multirow{2}{*}{\textbf{Complexity}} &
\textbf{ours}   & 42.8946 & 27.1455 & 21.2742 & 4.0562  & 5.9683 \\
\cline{2-7}
& \textbf{random} & 80.2622 & 49.2710 & 37.8636 & 10.4957 & 11.9227 \\
\hline

\multirow{2}{*}{\textbf{Interaction}} &
\textbf{ours}   & 33.7148 & 22.3178 & 18.3362 & 3.1600  & 4.6883 \\
\cline{2-7}
& \textbf{random} & 81.3070 & 50.8462 & 40.5505 & 11.1762 & 12.0857 \\
\hline

\end{tabular}
\end{table}

\section{Radar charts by baselines}
\label{Radar charts by baselines}

\begin{figure*}[htbp]
  \centering
  \includegraphics[width=0.98\textwidth]{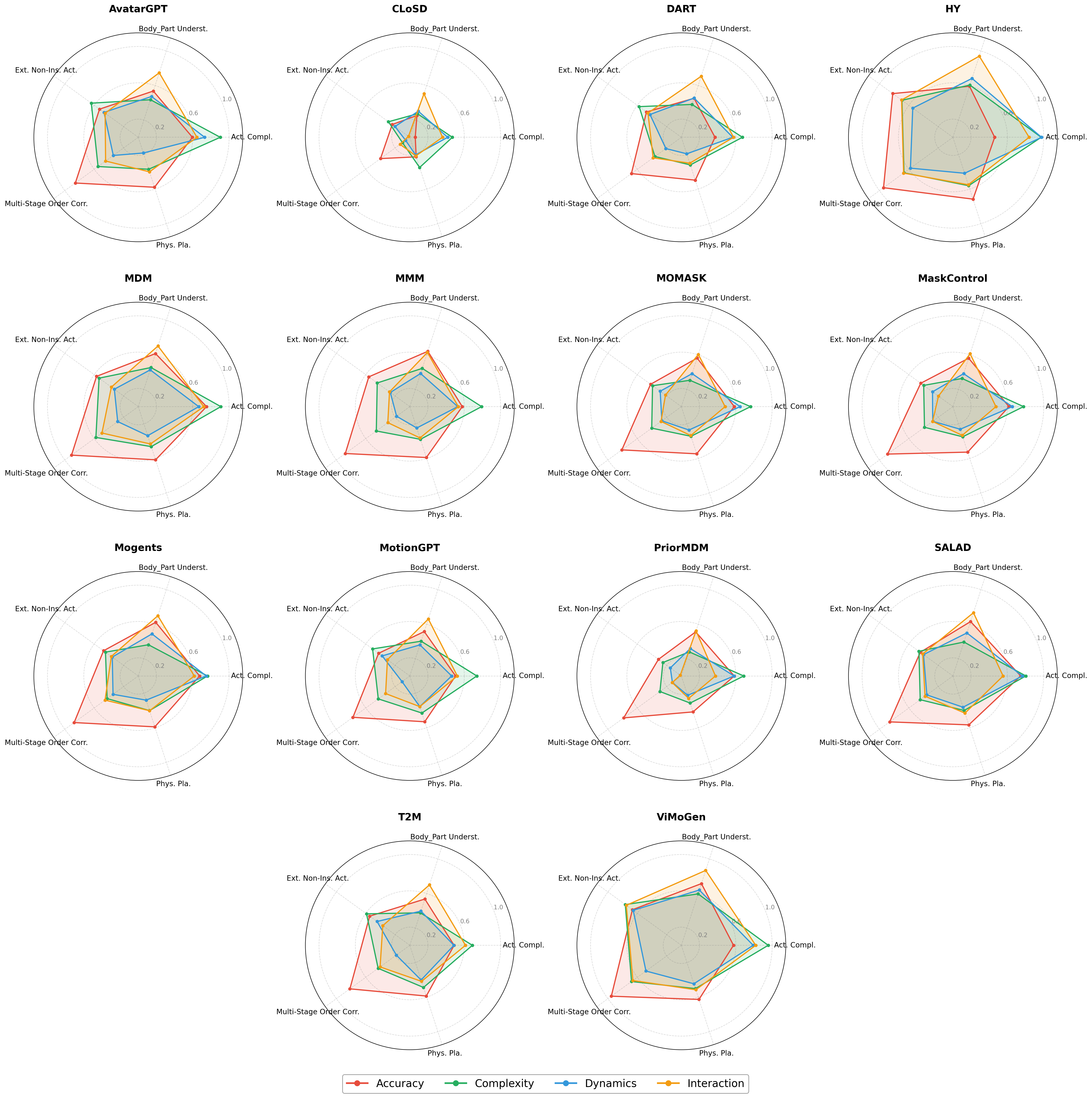} 
  \caption{LLM-Based radar charts by baseline}
  \label{fig:llm_14baseline} 
\end{figure*}

\begin{figure*}[htbp]
  \centering
  \includegraphics[width=0.98\textwidth]{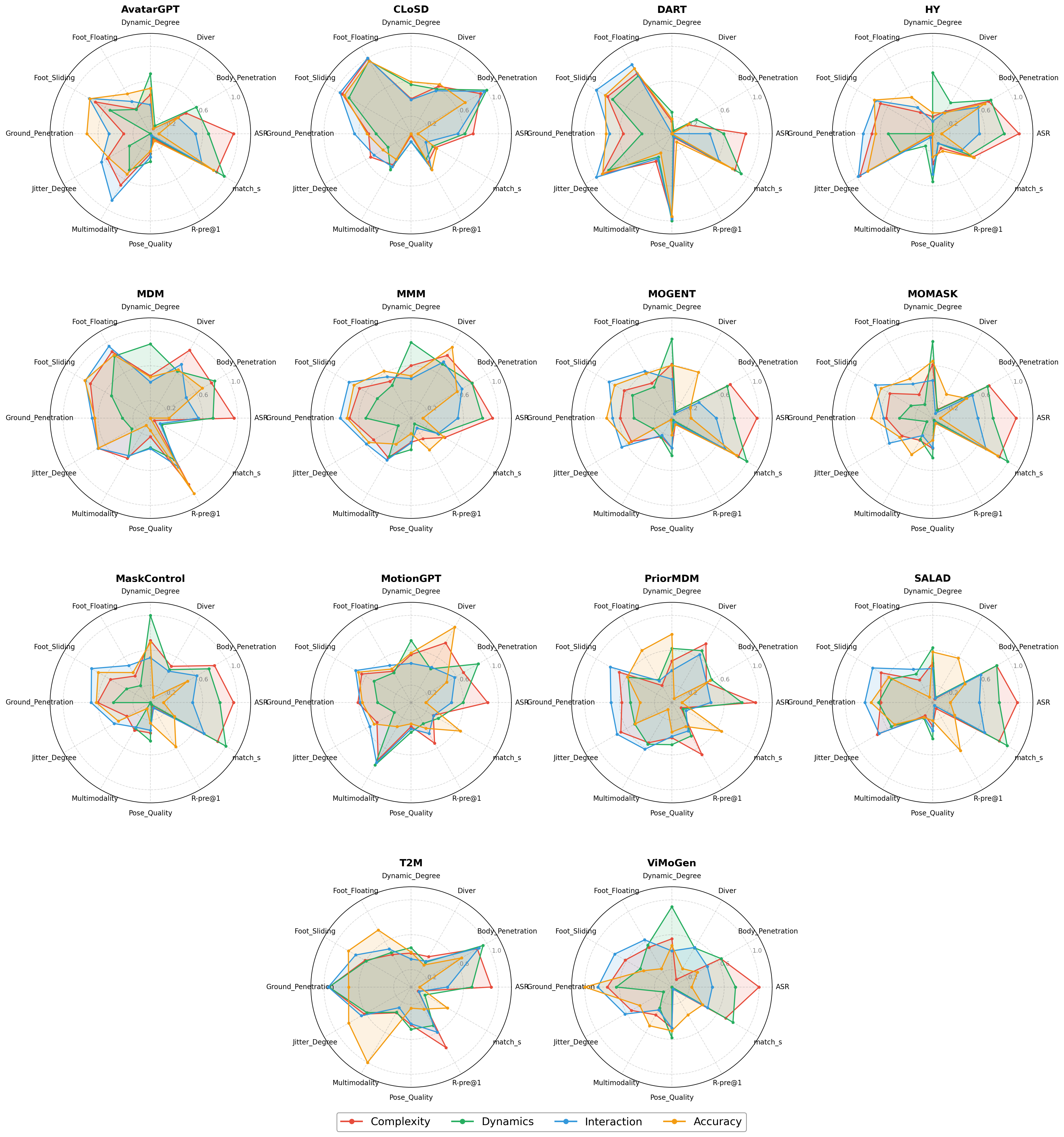} 
  \caption{Multi-Factor Motion Evaluation radar charts by baseline}
  \label{fig:multi-factor_14baseline} 
\end{figure*}

\begin{table*}[htbp]
  \centering
  \caption{Fine-grained accuracy assessments: mean RMSE ($\downarrow$) across baselines.}
  \label{tab:fine_grained_rmse}
  \begin{tabular}{lcccc}
    \toprule
    \multirow{2}{*}{Method} & \multicolumn{3}{c}{Whole-body motion} & Body-part motion \\
    \cmidrule(lr){2-4} \cmidrule(lr){5-5}
    & Root Rotation & Root Velocity & Root Translation & Body-part Translation \\
    \midrule
    T2M & 1.8055 & 5.7159 & 15.6568 & 0.5504 \\
    MDM & 1.9690 & 3.2875 & 8.8904 & 0.5392 \\
    MotionGPT & 1.5386 & 12.4603 & 17.2727 & 0.5568 \\
    CLoSD & 1.7394 & 11.0345 & 14.0019 & 0.5446 \\
    PriorMDM & 1.9986 & 2.2728 & 9.1077 & 0.6435 \\
    MMM & 1.7370 & 8.7938 & 23.8026 & 0.5576 \\
    AvatarGPT & 1.6213 & 3.1517 & 8.4948 & 0.5682 \\
    MOGENT & 1.7218 & 2.9884 & 8.5591 & 0.5707 \\
    SALAD & 1.7075 & 3.0097 & 9.0302 & 0.5736 \\
    MOMASK & 1.5300 & 2.9283 & 8.4809 & 0.5311 \\
    DART & 1.5580 & 3.4739 & 8.6864 & 0.5383 \\
    ViMoGen & 1.7080 & 3.3874 & 8.2046 & 0.5033 \\
    HY & 1.2818 & 2.3291 & 8.5777 & 0.5510 \\
    MaskControl & 1.4947 & 2.9176 & 8.4974 & 0.5457 \\
    MaskControl\_with\_JointControl & 1.6815 & 0.1244 & 0.1042 & 0.1020 \\
    \bottomrule
  \end{tabular}
\end{table*}


\end{document}